\documentclass[a4paper,fleqn]{cas-dc}

\usepackage[numbers]{natbib}
\usepackage{amsfonts}
\usepackage{balance}
\usepackage{booktabs}
\usepackage{amsmath}
\usepackage{tabularx}
\usepackage{subfigure}
\usepackage{graphicx}
\usepackage{epstopdf}
\usepackage{multirow}
\usepackage{rotating}
\usepackage{algorithmic}
\usepackage{footnote}
\usepackage{threeparttable}
\usepackage{color}
\usepackage[ruled,vlined,linesnumbered]{algorithm2e}

\DeclareMathOperator*{\argmin}{arg\,min}
\def\tsc#1{\csdef{#1}{\textsc{\lowercase{#1}}\xspace}}
\tsc{WGM}
\tsc{QE}
\tsc{EP}
\tsc{PMS}
\tsc{BEC}
\tsc{DE}
\begin{document}
\let\WriteBookmarks\relax
\def\floatpagepagefraction{1}
\def\textpagefraction{.001}
\shorttitle{Divide-and-conquer based Large-Scale Spectral Clustering}
\shortauthors{Hongmin Li et~al.}

\title [mode = title]{Divide-and-conquer based Large-Scale Spectral Clustering}
% \tnotemark[1,2]

\author{Hongmin Li}[type=editor,
                        auid=000,bioid=1,
                        % prefix=Sir,
                        % role=Researcher,
                        orcid=0000-0003-0228-0600]
\cormark[1]
%\cormark[2]
% \fnmark[1]
\ead{li.hongmin.xa@alumni.tsukuba.ac.jp}

\address{Department of Computer Science, University of Tsukuba, Tsukuba, Ibaraki 305-8577, Japan}

\author{Xiucai Ye}[auid=002,bioid=2,
                        % prefix=Sir,
                        % role=Researcher,
                        orcid=0000-0002-5547-3919]
\ead{yexiucai@cs.tsukuba.ac.jp}
\author{Akira Imakura}[auid=003,bioid=3,
                        % prefix=Sir,
                        % role=Researcher,
                        orcid=0000-0003-4994-2499]
\ead{imakura@cs.tsukuba.ac.jp}
\author{Tetsuya Sakurai}[auid=004,bioid=4,
                        % prefix=Sir,
                        % role=Researcher,
                        orcid=0000-0001-9128-9396]
\ead{sakurai@cs.tsukuba.ac.jp}
\cortext[cor1]{Corresponding author}
%\cortext[cor2]{Principal corresponding author}

\begin{abstract}
        Spectral clustering is one of the most popular clustering methods. 
        However, how to balance the efficiency and effectiveness of the large-scale spectral clustering with limited computing resources has not been properly solved for a long time.
        In this paper, we propose a divide-and-conquer based large-scale spectral clustering method to strike a good balance between efficiency and effectiveness.
        In the proposed method, a divide-and-conquer based landmark selection algorithm and a novel approximate similarity matrix approach are designed to construct a sparse similarity matrix within low computational complexities. Then clustering results can be computed quickly through a 
        bipartite graph partition process. 
        The proposed method achieves the lower computational complexity than most existing large-scale spectral clustering methods.
        Experimental results on ten large-scale datasets have demonstrated the efficiency and effectiveness of the proposed method.
        The MATLAB code of the proposed method and experimental datasets are available at \url{https://github.com/Li-Hongmin/MyPaperWithCode}.

\end{abstract}


\maketitle

\section{Introduction}
Clustering is one of the most fundamental problems in data mining and machine learning, aiming to categorize data points into clusters such that the data points in the same cluster are more similar while data points in different clusters are more different from each other \cite{rokach2005clustering,xu2008clustering,liu2013understanding}.
Spectral clustering has attracted increasing attention due to the promising ability to deal with nonlinearly separable datasets \cite{filippone2008survey,ng2002spectral}.
It has been successfully applied to various problem domains such as biology \cite{pentney2005spectral}, image segmentation \cite{zhang2008spectral}, and recommend systems \cite{zhang2014detection,ye2018spectral}.
Although spectral clustering algorithm often provides better performances than traditional clustering algorithm likes $K$-means especially for complex datasets, it is significantly limited to be applied to large-scale datasets due to its high computational complexity and space complexity \cite{li2015large,ye2018large}.

The conventional spectral clustering algorithm mainly consists of two high-cost steps, i.e., similarity matrix construction and eigen-decomposition.
For a dataset with $N$ objects, the two steps take computational complexities of $O(N^2d)$ and $O(N^3)$, respectively.
The computational consumption of these two steps is the main reason that hinders the application of spectral clustering algorithms on large-scale data.

In recent years, there has been an increasing amount of literature on alleviating the computational complexity of spectral clustering \cite{fowlkes2004spectral,cai2014large,li2015superpixel,bouneffouf2015sampling,zhang2016sampling,ye2018large,huang2019ultra,li2020hubness}.
Previous research \cite{chen2010parallel} has established that the sparse similarity matrix construed by only remaining $k$-nearest neighbors or $\epsilon$-nearest neighbors can efficiently reduce the space complexity. As a result, some sparse eigensolvers can solve the eigen-decomposition problems within the lower computational complexity.
The matrix specification strategy can avoid storing the dense similarity matrix to reduces the space complexity, but it still needs to compute the dense similarity matrix at first, which costs $O(N^2d)$ computational complexity.
Besides the matrix specification, another commonly used strategy is based on a cross-similarity matrix construction \cite{fowlkes2004spectral, cai2014large, li2015superpixel, ye2018large,li2020hubness,huang2019ultra}.
Fowlkes et al. \cite{fowlkes2004spectral} apply the Nystr{\"{o}}m method to reduce the high complexity of spectral clustering algorithm, which first randomly selects a small subset of samples as landmarks, then construct a similarity sub-matrix between these landmarks and remaining samples.
Although the random landmark selection is very efficient, it is often unstable concerning the quality of the landmark set.
Moreover, it has been shown that a larger $p$ is often favorable for better approximation. 
To address the potential instability of random selection, Cai et al. \cite{cai2014large} extend the Nystr{\"{o}}m method and propose a landmark based large-scale spectral clustering (LSC) method, which uses $k$-means to obtain $p$ cluster centers as $p$ landmark points to construct the similarity sub-matrix.
With the constructed $N\times p$ sub-matrix, they then convert it into sparse by preserving the $k$-nearest landmarks of the data points and filling with zeros to others.
By the $k$-means based landmarks selection, the LSC algorithm shows better performance than Nystr\"{o}m.
On this basis, some studies \cite{zhang2008improved,bouneffouf2015sampling,zhang2016sampling,rafailidis2017landmark,ye2018large, li2020hubness} on the landmark selection are further proposed to improve the instability of sub-matrix based large-scale spectral clustering.
However, the computational complexity of the sub-matrix construction can still be a critical bottleneck when dealing with large-scale clustering tasks.
Huang et al. \cite{huang2019ultra} propose a hybrid representative (landmark) selection method that initializes candidate samples randomly from the dataset and performs $k$-means to obtain $p$ cluster centers as the representative points then computes the approximation of $K$-nearest representatives. It does not compute the dense similarity sub-matrix but approximates a sparse sub-matrix, further reducing similarity construction costs.
However, those sub-matrix based spectral clustering algorithms are typically restricted by an $O(Np)$ or $O(Np^{\frac{1}{2}})$ complexity bottleneck, which is still a critical hurdle for them to deal with large-scale datasets where a larger $p$ is often desired for achieving better approximation.
Although some considerable studies have been proposed in recent years, it remains a highly challenging problem, i.e., how to make spectral clustering handle large-scale datasets efficiently and effectively within limited computing resources.

In this paper, to achieve a better balance between the effectiveness and efficiency of the spectral clustering for large-scale datasets, we propose the \textbf{d}ivide-a\textbf{n}d-\textbf{c}onquer \textbf{s}pectral \textbf{c}lustering (DnC-SC) method.
In DnC-SC, a novel divide-and-conquer based landmark selection method is proposed to generate high-quality $p$ landmarks, which reduces the computational complexity of $k$-means based selection from $O(Npdt)$ to $O(N\alpha d)$, where $\alpha$ is the \textit{selection rate parameter} that determines the upper bound of computational complexity.
Besides, a fast approximation method for $K$-nearest landmarks is designed to efficiently build a sparse sub-matrix with $O(NKd)$ computational complexity and $O(NKd)$ space complexity.
A cross similarity matrix is constructed between the data points and the $p$ landmarks, which can be interpreted as the edges matrix of a bipartite graph.
The bipartite graph partitioning is then conducted to solve the spectrum with $O(NK(K+k)+p^3)$, where $k$ is the number of clusters.
Finally, the $k$-means method is used to obtain the clustering result on the spectrum with $O(Nk^2t)$, where $t$ is the number of iterations during $k$-means.
As it generally holds that $k, K, \alpha \ll p\ll N$, the computational and space complexity of our DnC-SC algorithm are respectively dominated by $O(N\alpha d)$ and $O(NK)$.
The experimental results on ten large-scale datasets (consisting of five real-word datasets and five synthetic datasets) show the priority performance of proposed methods on both efficiency and effectiveness.

The main contributions of the proposed method are summaries as follows: 

\begin{itemize}
  \item A divide-and-conquer-based landmark selection method is proposed to efficiently find $p$ centralized subset centers as landmarks in a recursive manner.
  \item A fast $K$-nearest landmarks search method is designed, which uses centers' nature of landmarks to identify the most possible $K$-nearest landmarks candidates.
  \item A large-scale spectral clustering algorithm termed DnC-SC is proposed, which efficiently constructs the similarity matrix and uses bipartite graph partitioning to obtain final clustering results. Its computational and space complexity is dominated by $O(N\alpha d)$ and $O(NK)$, which achieves a lower computational complexity than most existing large-scale spectral clustering methods.

\end{itemize}

\section{Preliminaries}
\label{sec:related_work}

This section reviews the literature related to spectral clustering and large-scale spectral clustering extensions.

\subsection{Spectral Clustering}

Spectral clustering aims to partition the data points into $k$ clusters using the spectrum of the graph Laplacians \cite{von2007tutorial}.
Given a dataset ${X}=\left\{x_{1}, \ldots, x_{N}\right\}$ with $N$ data points, spectral clustering algorithm first constructs similarity matrix ${W}$, where ${w_{ij}}$ indicates the similarity between data points $x_i$ and $x_j$ via a similarity measure metric.

Let $L=D-W$, where $L$ is called graph Laplacian and ${D}$ is a diagonal matrix with $d_{ii} = \sum_ {j=1}^n w_{ij}$.
The objective function of spectral clustering can be formulated based on the graph Laplacian as follow:
\begin{equation}
  \label{eq:SC_obj}
  {\min_{{U}}  \operatorname{tr}\left({U}^{T} {L} {U}\right)}, \\ {\text { s.t. } \quad {U}^{T} {{U}={I}}},
\end{equation}
where $\operatorname{tr(\cdot)}$ denotes the trace norm of a matrix.
The rows of matrix ${U}$ are the low dimensional embedding of the original data points.
Generally, spectral clustering computes ${U}$ as the bottom $k$ eigenvectors of ${L}$, and finally applies $k$-means on ${U}$ to obtain the clustering results.

\subsection{Large-scale Spectral Clustering}

\subsubsection{Similarity Sub-matrix construction}
Instead of an $N\times N$ similarity matrix, many large-scale spectral clustering methods \cite{fowlkes2004spectral, cai2014large, li2015superpixel, ye2018large,li2020hubness,huang2019ultra} are using a similarity sub-matrix to represent each data points.
The similarity sub-matrix consists of the cross-similarities between data points and a set of representative data points (i.e., landmarks) via some similarity measures, as
\begin{equation}
    \label{eq: cross-similarity}
    B = \Phi(X,R),
\end{equation}
where $R = \{r_1,r_2,\dots, r_p \}$ ($p \ll N$) is a set of landmarks with the same dimension to $X$, $\Phi(\cdot)$ indicate a similarity measure metric, and $B\in \mathbb{R}^{N\times p}$ is the similarity sub-matrix to represent the $X \in \mathbb{R}^{N\times d}$ with respect to the $R\in \mathbb{R}^{p\times d}$.

Ideally, the landmark points $r_1, r_2,\cdots, r_p$ would roughly represent the distribution of $X$.
Some previous studies \cite{cai2014large} show the effectiveness of $k$-means based selection.
The objective function of $k$-means based landmark selection can represented as follows: 
\begin{equation}
  \label{eq: opt}
  {R}= \argmin_{r_1, \dots, r_p} \sum_{i = 1}^p\sum_{x_j\in S_i}\left\|x_j-r_i\right\|^{2},
\end{equation}
where $S_1,S_2,\dots,S_p$ indicate the subsets that are nearest to $r_{1}, r_{2}, \cdots, r_{p}$, respectively.
However, directly conducting $k$-means on large-scale datasets faces a high time cost of $O(Npdt)$.
Moreover, $k$-means often needs more iterations to converges on large-scale datasets.

\subsubsection{Efficient Bipartite Graph Partitioning}

The similarity sub-matrix $B$ reflects the relationship between $X$ and $R$, which can be naturally treated as a bipartite graph $G=\{X,R,B\}$.
The goal of bipartite graph partitioning is to partition the graph $G$ into $k$ groups.
The full similarity matrix of $G$ is then designed as \cite{zha2001bipartite}
\begin{equation}
  \label{eq: WusedB }
  W=\left[\begin{array}{ll}
      \mathbf{0} & B          \\
      B^{T}      & \mathbf{0}
    \end{array}\right].
\end{equation}
The size of matrix $W$ is $(N+p)\times (N+p)$. 
The conventional spectral clustering finds a low dimensional embedding via the spectrum of graph Laplacian, which solves the generalized eigen-problem \cite{shi2000normalized}:
\begin{align}
  \label{eq:general_eigen_problem}
  Lf=\gamma Df,
\end{align}
where $L=D-W$ is the graph Laplacian and $D$ is a diagonal matrix with $d_{ii} = \sum_ {j=1}^n w_{ij}$.
Note that the eigenvector $f$ can be interpreted as two parts $u \in \mathbb{R}^{N}$ and $v \in \mathbb{R}^{p}$.
\begin{equation}
  \label{eq:eigenvectors_f}
  f=\left[\begin{array}{l}
      u   \\
      v
    \end{array}\right],
\end{equation}
where $u$ is the eigenvector on $X$ side while $v$ is the eigenvector on $R$ side.

The problem is how to efficiently compute eigenvector $u$ and construct a low dimensional embedding on original data $X$.
An efficient computation method called \textit{transfer cut} is often used to compute the spectrum of graph Laplacian for bipartite graph partitioning problems.
Instead of directly computing $f$ by partial SVDs or dual property of SVD \cite{cai2014large}, the transfer cuts process first computes the $v$ by solving a much smaller eigen-problem as follows:
\begin{align}
  L_{{R}}v =\lambda D_{R}v,\label{eq:general_eigen_problem_reduced}
\end{align}
where $L_{{R}}=D_{R}-B^TD_{X}^{-1}B$, $D_{X} \in \mathbb{R}^{n \times n}$ and $D_{R}\in \mathbb{R}^{p \times p}$ are the diagonal matrices whose entries are $d_{X}(i,i) = \sum_{j=1}^n B_{ij}$ and $D_{R}(j,j) = \sum_{i=1}^n B_{ij}$, respectively.
It has been pointed out that the eigen-problems \eqref{eq:general_eigen_problem} on original bipartite graph $G$ and the much smaller one \eqref{eq:general_eigen_problem_reduced} are essential equivalence \cite{li2012segmentation}.
Let $\left\{\left(\lambda_{i}, {v}_{i}\right)\right\}_{i=1}^{k}$ be the bottom $k$ eigenpairs of the eigen-problem \eqref{eq:general_eigen_problem_reduced} and $0=\lambda_{1} \leq \cdots \leq \lambda_{k}<1$.
Then $\left\{\left(\gamma_{i}, {f}_{i}\right)\right\}_{i=1}^{k}$ are the bottom $k$ eigenpairs of the eigen-problem \eqref{eq:general_eigen_problem} and $0=\gamma_{1} \leq \cdots \leq \gamma_{k}<1$.
It have been proved that \cite{li2015superpixel}
\begin{equation}
  u_i = \frac{1}{1-\gamma_i}Tv_i, \label{eq:u}
\end{equation}
where $1 \leq \gamma_i <1$, $\lambda_i=\gamma_i(2-\gamma_i)$ and $T:=D_{X}^{-1}B$ is called the associated transition probability matrix.
Therefore bottom $k$ eigenvectors $u_1,\dots,u_k$ are calculated according to \eqref{eq:general_eigen_problem_reduced} and \eqref{eq:u}.
Let $U\in \mathbb{R}^{N\times k}$ be the matrix containing the vectors $u_1,\dots,u_k$ as columns.
Then $U$ is the spectral embedding of the large-scale spectral clustering algorithm.
Finally, $k$-means is conducted on the embedding to obtain final clustering results.

\section{Proposed Framework}
\label{sec:framework}

\begin{figure*}
    \centering
    \includegraphics[width=1.0\textwidth]{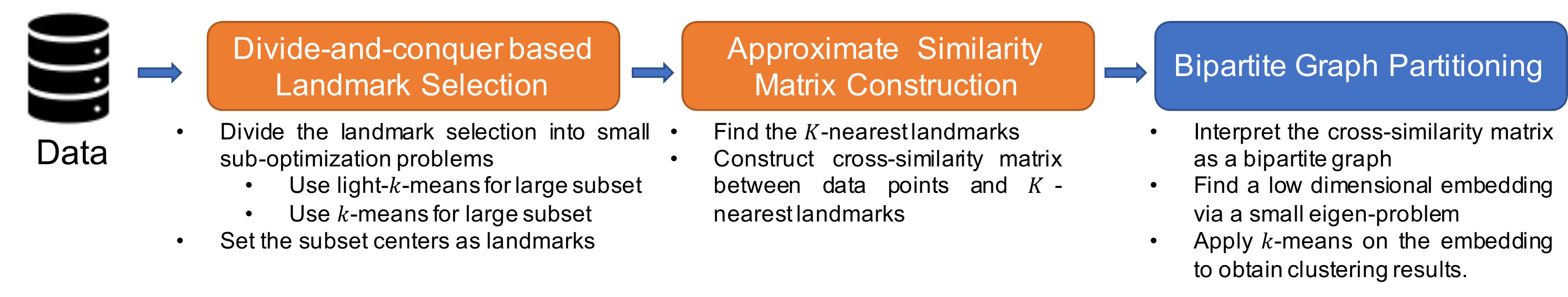}
    \caption{An overview of proposed DnC-SC method. Given a dataset, the DnC-SC method first finds the landmarks via divide-and-conquer based landmark selection, then approximately constructs the similarity matrix, finally conducts the bipartite graph partitioning to obtain final clustering results.
    Our main contributions focus on the first two phases (colored as orange), i.e., the landmark selection and similarity construction phases.}
    \label{fig:overview}
\end{figure*}

To further reduce the complexity of spectral clustering, we propose the DnC-SC method that complies with the sub-matrix based formulation \cite{fowlkes2004spectral,cai2014large} and aims to break through the efficiency bottleneck of previous algorithms.
DnC-SC consists of three phases:
(1) Divide-and-conquer based landmark selection:
we consider landmark selection as an optimization problem and present a divide-and-conquer based landmark selection method to find the landmarks via solving the sub-optimization problems recursively.
(2) Approximate similarity matrix construction: we design a novel strategy to efficiently approximate the $K$-nearest landmarks for each data point and construct a sparse cross-similarity matrix between the $N$ data points and the $p$ landmarks.
(3) Bipartite graph partitioning: we interpret the cross-similarity as a bipartite graph and conduct the bipartite graph partitioning to obtain the clustering result.
We summarize the proposed method in Figure \ref{fig:overview}.

\subsection{Divide-and-conquer based Landmark Selection}
\label{sec:landmark_selection}

\begin{figure}
    \centering
    \includegraphics{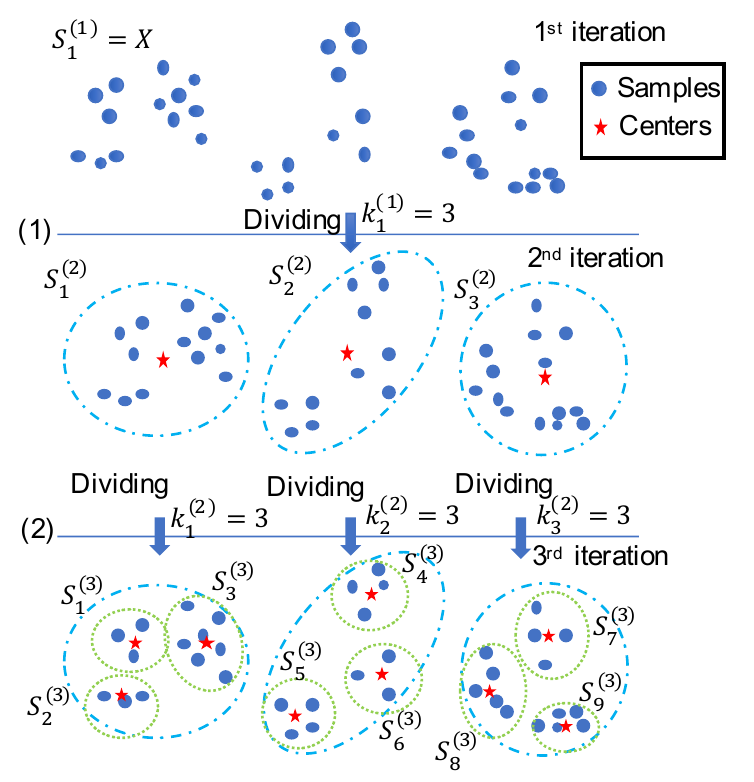}
    \caption{An illustration of the divide-and-conquer base landmark selection: (1) The dataset is initially divided into $k_1^{(1)} =3$ initial subsets; (2) Each subset is further divided into 3 smaller subsets and the total number of subsets reaches $p=9$. Finally, the centers of subsets are turned as landmarks.
    }
    \label{fig:DnC Selection}
\end{figure}

We propose a divide-and-conquer based landmark selection method, which aims to find a set of high-quality landmarks efficiently.
Instead of directly dividing data points into $p$ subsets like \eqref{eq: opt} for landmark selection, we first divide data points into $\alpha$ subsets, then recursively divide each subset into $k$ smaller subsets ($k\leq \alpha$), until the total number of subsets reaches $p$.
Denote $\alpha$ as a small number and $\alpha\ll p \ll n$. 
We define $\alpha$ as \textit{selection rate parameter} that is the upper boundary of desired subset number in each dividing process to limit the computational complexity.
Figure \ref{fig:DnC Selection} gives a simple example. The data points are recursively divided into $\alpha$ subsets until the total number of subsets is $p$, which avoids directly applying $k$-means to obtain too many subsets at once.

In the divide-and-conquer strategy, the number of desired subsets in each iteration is much smaller than $p$, and the subsets are smaller and smaller during iterations than directly applying $k$-means to datasets.
Suppose each dividing process will divide $\alpha$ subsets of the same size, the number of subsets increases exponentially and reaches $p$ subsets in $\lceil\log_\alpha p \rceil$ rounds.
Denote $S_1^{(1)}=X$ is the initial subset.
In first round of dividing process, $N$ data points are divided into $\alpha$ subsets $S_1^{(2)},S_2^{(2)},\dots,S_\alpha^{(2)}$ with computational complexity of $O(N\alpha dt)$, which is illustrated in the (1) of Figure \ref{fig:DnC Selection}.
There are $\alpha$ subsets currently, each current subset is then divided into $\alpha$ new subsets in the second round with computational complexity of $O(\frac{N}{\alpha}\alpha dt)$.
For example, $S_1^{(2)}$ is further divided into $\alpha$ new subsets $S_1^{(3)},S_2^{(3)},\dots,S_\alpha^{(3)}$, which is illustrated in the (2) of Figure \ref{fig:DnC Selection}.
The total computational complexity of the second round is $O(\alpha \frac{N}{\alpha}\alpha dt)=O(N\alpha dt)$.
The dividing process of second round generates $\alpha^2$ new subsets $S_1^{(3)},S_2^{(3)},\dots,S_{\alpha^2}^{(3)}$ totally.
Similarly, the computational complexity of any $i$-th ($i=1,2,\dots,\lceil\log_\alpha p \rceil$) dividing round will be $O(\alpha^{i-1} \frac{N}{\alpha^{i-1}}\alpha dt)=O(N\alpha dt)$.
Therefore, the total computational complexity of divide-and-conquer strategy is $O(\lceil\log_\alpha p \rceil N\alpha dt)$.
Note that $\log_\alpha p$ is a small value and can be treated as a constant, e.g., $\log_\alpha p$= 1.77 when setting $p=1000$ and $\alpha = 50$.
We further simplify the total computational cost as $O(N\alpha dt)$.
Compared with $k$-means based landmark selection, the divide-and-conquer strategy can naturally reduce the computational complexity from $O(N p d t)$ to $O(N\alpha d t)$ ($\alpha \ll p$).
Moreover, we design an efficient dividing algorithm, named light-$k$-means, to further accelerate the whole process into $O(N\alpha d)$.

\subsubsection{Divide-and-conquer Selection Strategy}

Before starting landmark selection, we first review the optimization problem \eqref{eq: opt}.
The variables in \eqref{eq: opt} are the $r_1, \dots, r_p$ which are used to map the unique $S_1,\dots, S_p$.
By setting the subsets $S_1,\dots, S_p$ and landmark number as the variables, we can rewrite the optimization problem \eqref{eq: opt} into a function form as follows:
\begin{equation}
    \label{eq: obj function}
    g(X, p)= \argmin_{S_1, \dots, S_p} \sum_{i = 1}^p\sum_{x_j\in S_i}\left\|x_j-r_i\right\|^{2},
\end{equation}
where $g(\cdot)$ indicates a centralized clustering problem that divides $X$ into $p$ subsets and $r_i$ is the center of subset $S_i$.
For any dividing problem that divides $Q$ into $h$ subsets, the function $g(Q, h)$ can be used to describe the dividing problem, and its computational complexity is $O(\|Q\|hdt)$, where $\|Q\|$ is the total number of samples in $Q$. 
More importantly, function $g(Q, h)$ can be used to derive the recursive function as follows:
\begin{align}
    &g(Q,h) = \bigcup_{i =1}^m g(A_i, k_i), \label{eq: ai}\\
    &\{A_1, \dots, A_m\} = g(Q,m), \label{eq: gm}
\end{align}
where $A_i$ is a subset of $Q$ and $Q = \bigcup_{i=1}^m A_i$; $k_i$ is the desired subset number of subset $A_i$ and $h = \sum_{i=1}^m k_i$; $m<h$.
\eqref{eq: ai} can simply divide any optimization problem into $m$ sub-problems, which builds a bridge between global problem $g(Q,h)$ and local problem $g(A_i, k_i)$.
We can recursively apply \eqref{eq: ai} and \eqref{eq: gm} to divide the optimization problem \eqref{eq: obj function} into the sub-problems small enough and solve them locally and efficiently. 

Denote $c_i$ as the total number of subsets during $i$-th iteration.
We will stop the recursive process in the $j$-th iteration when $c_j$ reaches the desired total number of subsets $p$.
Initially, we assign all data points as one subset.
As we only have one subset ($c_1=1<p$), the dividing process happens.
Let $k^{(i)}_j$ be the desired number of subsets for dividing process on $j$-th subset during $i$-th iteration.
We naturally set the desired number $k^{(1)}_1 =p$.
However, directly apply $g(X, p)$ may be time-consuming.
For $k>\alpha$, we force $k=\alpha$ to obtain subsets partially.
As a result, we have the initial setting as follows:
\begin{equation}
\label{eq:k1} 
    \left\{\begin{matrix}
S_1^{(1)} = X \\
k^{(1)}_1=\alpha
\end{matrix}\right.
\end{equation}
In the first iteration, we divide $S_1^{(1)}$ into $k^{(1)}_1$ subsets in as follows: 
\begin{equation}
    \label{eq: subset0}
    \{{S}_1^{(2)}, {S}_2^{(2)},\cdots,\ {S}_{c_2}^{(2)}\} = g(S_1^{(1)}, k^{(1)}_1),
\end{equation}
where ${S}_i^{(1) }$ indicates the $i$-th subset during first iteration and $c_2 =\sum_{i =1}^{c_1} k^{(i)}_1 = k^{(1)}_1$ is the total number of subsets.

From the second iteration, there are more and more subsets being obtained. 
Thus, we need a subset number allocation strategy to determine the desired number of subsets and guide iteration dynamically.
We define the residual sum of squares (RSS) of subset ${S}_i^{(j)}$ as 
\begin{equation}
\label{eq: zeta_i_j}
    \zeta_i^{(j)} = \sum_{x_l \in {S}_i^{(j)}} \left\|x_l- r_i \right\|^2,
\end{equation}
where $r_i$ is the center of the subset ${S}_i^{(j)}$.
Consider the global problem \eqref{eq: opt}, the desired number of subsets should be proportional to their RSS.
We propose a dynamical allocation strategy as follows: 
\begin{equation}
  \label{eq:DA}
  k_i^{(j)} =
  \begin{cases}
    \frac{\zeta_i^{(j)}}{\sum \zeta_i^{(j)}} p, & \text{if } \frac{\zeta_i^{(j)}}{\sum \zeta_i^{(j)}} p <\alpha, \\
    \alpha,                   & \text{otherwise},
  \end{cases}
\end{equation}
where $k_i^{(j)}$ is the allocated dividing number for subset ${S}_i^{(j) }$.
Then, all $k_i^{(j)}$ are turned as integers and fix the $c_{j+1}<= p$, where $c_{j+1}=\sum_i k_i^{(j)}$.
After obtaining $k_1^{(j)},\dots,k_{c_j}^{(j)}$ ($c_j<p$), we will divide each subset ${S}_i^{(j)}$ into $k_i^{(j)}$ smaller subsets via $g({S}_i^{(j)}, k_i^{(j)})$.
We then collect all subsets as follows: 
\begin{equation}
\label{eq:newcluter}
    \{{S}_1^{(j+ 1)}, {S}_2^{(j+ 1))},\cdots,\ {S}_{c_{j+1}}^{(j+ 1)}\} = \bigcup_{i =1}^{c_j} g({S}_i^{(j)}, k_i^{(j)}).
\end{equation}
We repeat the above process until $p$ subsets have been produced and set the $p$ subset centers as the landmarks.

Take an example using Figure \ref{fig:DnC Selection}, where we set $\alpha=3$ and $p=9$.
In the first iteration, we assign all data points as one subset. 
Since the desired landmark number $p>\alpha$, we set $k_1^{(1)}=3$.
Then we initially divide the dataset $S_1^{(1)}=X$ into $k_1^{(1)}=3$ subsets. 
In the second iteration, there are three subsets $S_1^{(2)}$, $S_2^{(2)}$ and $S_3^{(2)}$.
According to \eqref{eq:DA}, we compute the $k_1^{(2)}$, $k_2^{(2)}$, $k_2^{(2)}$.
Suppose $k_1^{(2)}=k_2^{(2)}=k_2^{(2)}=3$, we then divide $S_1^{(2)}$, $S_2^{(2)}$, $S_3^{(2)}$ into $k_1^{(2)},k_2^{(2)},k_2^{(2)}$ smaller subset respectively.
In the third iteration, there are 9 subsets $S_1^{{(3)}}, \dots, S_9^{{(3)}}$.
Since the total number of subsets $c_3 = 9$ reaches the desired landmark number $p=9$, we stop the recursive process in the third iteration. 
Finally, compute the subset centers $r_1, r_2, \dots, r_p$ of $S_1^{{(3)}}, \dots, S_9^{{(3)}}$ and set them as the landmarks.

Note that the dividing process $g(\cdot)$ can be directly solved by the $k$-means method.
However, directly apply $k$-means on large data is time-consuming.
To further reduce the complexity, we propose a modified $k$-means method, named light-$k$-means.
When dataset size is large, we conduct the dividing process via light-$k$-means. 
Otherwise, we use the traditional $k$-means method.
We summary the divide-and-conquer based landmark selection method in Algorithm \ref{ag: divide-and-conquer based landmark selection method}.

\begin{algorithm}
  \label{ag: divide-and-conquer based landmark selection method}
  \caption{Divide-and-conquer based landmark selection method}
  \SetAlgoLined
  \KwIn{Dataset $X$, the number of landmarks $p$, selection rate $\alpha$;}
  \KwOut{Landmarks ${R}$;}
  Initialize: Set the $S_1^{(1)}$ and $k^{(1)}_1$ via \eqref{eq:k1};\\   
  $c_{1}=1$;\\
  $j=1$;\\
  \While {total number of subsets $c_j < p$}{
    Set the $k_1^{(j)},\dots,k_c^{(j)}$ via \eqref{eq:DA};\\
    \ForEach(){$S_i^{(j)}$}{
    \eIf{the size of $S_i^{(j)}$ is larger than $p'$}
     {Conduct \eqref{eq:newcluter} via \textit{light-$k$-means};}
    {Conduct \eqref{eq:newcluter} via $k$-means;}
    }
    $c_{j+1}=\sum_i k_i^{(j)}$;\\ 
    $j = j+1$;
  }
  Collect the latest cluster centers as $R$.
\end{algorithm}
\subsubsection{Light-$k$-means Algorithm}

We define $p'$ as an upper bound.
When the size of ${S}_i^{(j)}$ is larger than $p'$, we will use light-$k$-means to compute $g({S}_i^{(j)}, k_i^{(j)})$.
The light-$k$-means is performed as the following steps:
\begin{enumerate}
\item Randomly select $p'$ representatives from ${S}_i^{(j)}$ and denote them in a set as $H$ and the complement of $H$ is $H^c$.
\item Conduct $k$-means to divide $H$ into $k_i^{(j)}$ subsets;
\item Find the nearest subset centers for the remained data points in $H^c$;
\item Assign the remained data points in $H^c$ to their nearest subsets (with the center nearest to these points).
\end{enumerate}

\begin{figure}
    \centering
    \includegraphics[width=1.0\linewidth]{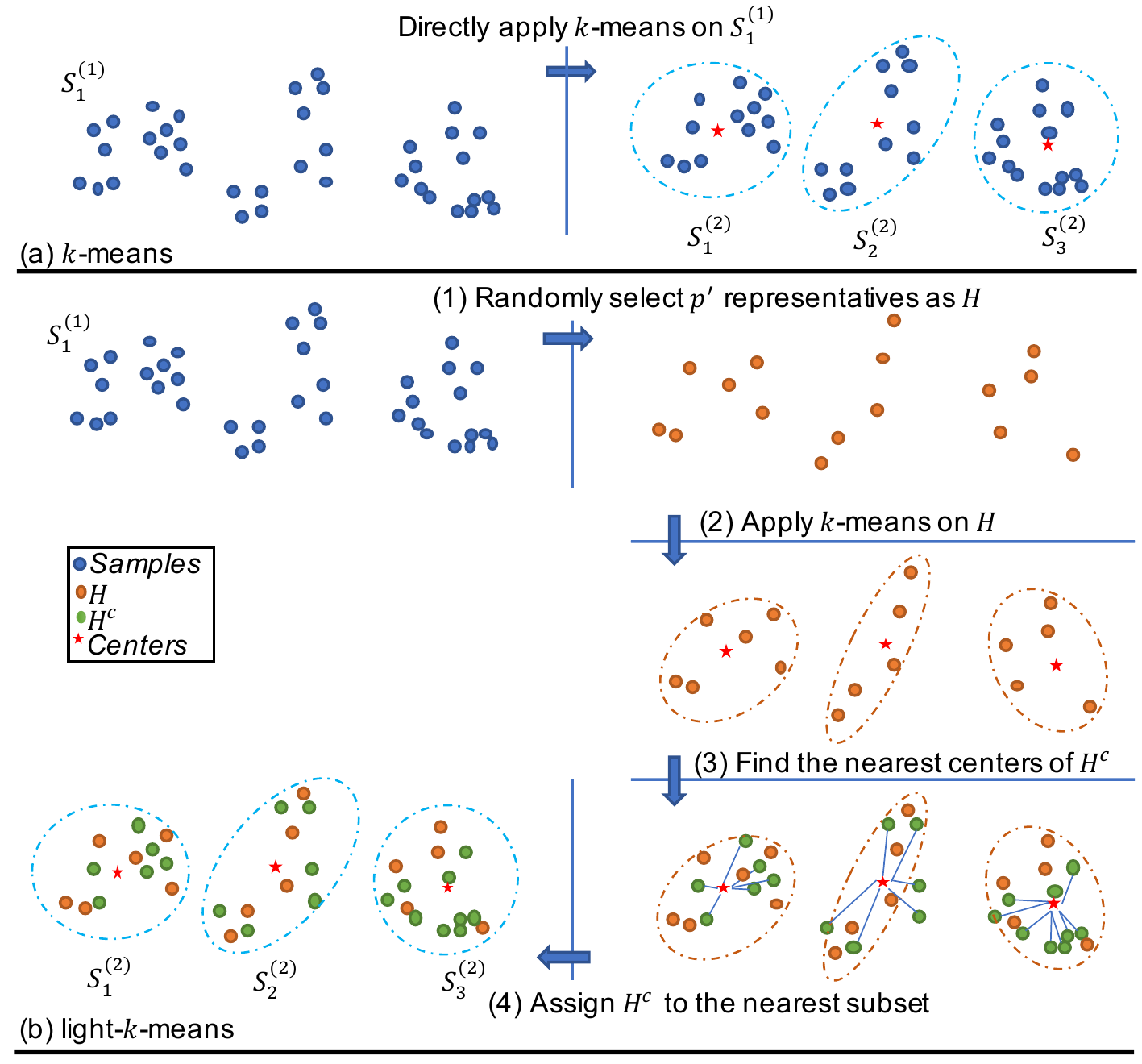}
    \caption{An comparison between $k$-means and light-$k$-means. (a) $k$-means directly divide all samples $S_1^{(1)}$ into 3 subsets.
    (b) In light-$k$-means, $k$-means is applied on $p'$ representatives, which significantly reduces the complexity on large data.
    }
    \label{fig:light_k_means}
\end{figure}

Figure \ref{fig:light_k_means} shows a comparison between $k$-means and light-$k$-means method, which are the implementation examples of (1) in Figure \ref{fig:DnC Selection}.
Given a subset $S_1^{(1)}$, the light-$k$-means first randomly select $p'$ data points and denotes them as $H$ and the complement is $H^c$ ($S_1^{(1)}=H\cup{H^c}$).
Then the $k$-means is used to divide $H$ into $k_1^{(1)}$ subsets, i.e., $A_1, A_2, A_3$.
For each data points in $H^c$, find its nearest center and assign it to the subset, i.e., $A_1^c, A_2^c, A_3^c$, according to its nearest center.
Finally, return the combined subsets $S_1^{2},S_2^{2},S_3^{2}$ as the results of this dividing process.

\begin{algorithm}[]
  \label{ag: Light-$k$-means}
  \SetAlgoLined
  \KwIn{Data ${S}_i^{(j)}$, the number of cluster $k_i^{(j)}$, number of samples $p'$;}
  \KwOut{$k_i^{(j)}$ subsets;}
    Randomly select $p'$ samples from ${S}_i^{(j)}$ and denote them as $H$;\\
    Denote the complement of $H$ as $H^c$;\\
    Apply $k$-means to divide $H$ into $k_i^{(j)}$ subsets;\\
    Find the nearest center of samples in $H^c$;\\
    Assign the samples in $H^c$ to the subset according to their nearest centers.
  \caption{Light-$k$-means}
\end{algorithm}

Denote $n_i^{(j)}$ as the number of samples in ${S}_i^{(j)}$.
The computational complexity of light-$k$-means for the dividing process $g({S}_i^{(j)},k_i^{(j)})$ should be $O(p'k_i^{(j)}t)+O((n_i^{(j)}-p')k_i^{(j)}d)) = O(n_i^{(j)}k_i^{(j)}d + p'k_i^{(j)}d(t-1))$, where $O(n_i^{(j)}k_i^{(j)}d)$ is the dominant term.
While, $k$-means costs $O(n_i^{(j)}k_i^{(j)}dt)$ for the same dividing process.
Compared with $k$-means, light-$k$-means significantly alleviates the computational complexity of iterative optimization.
Empirically, the number of $p'$ is suggested to be several times larger than $p$, e.g., $p'=10p$, to provide enough samples for the $k$-means algorithm.
Since our landmark selection focuses more on the local dividing, the light-$k$-means can effectively divide the large subsets into small ones and find more accurate subsets locally.
Finally, we summarise the light-$k$-means method in Algorithm \ref{ag: Light-$k$-means}.

\begin{figure}[]
  \begin{center}
    {\subfigure[]
      {\includegraphics[width=0.31\linewidth]{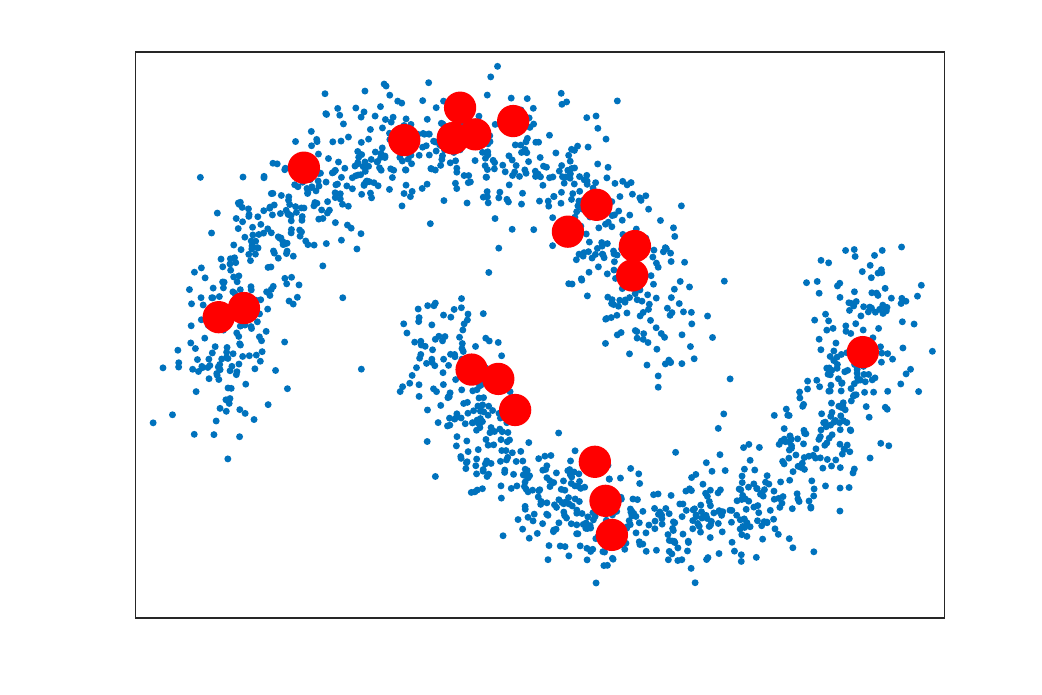}\label{fig:cmpSelStrategy_random}}}
    {\subfigure[]
      {\includegraphics[width=0.31\linewidth]{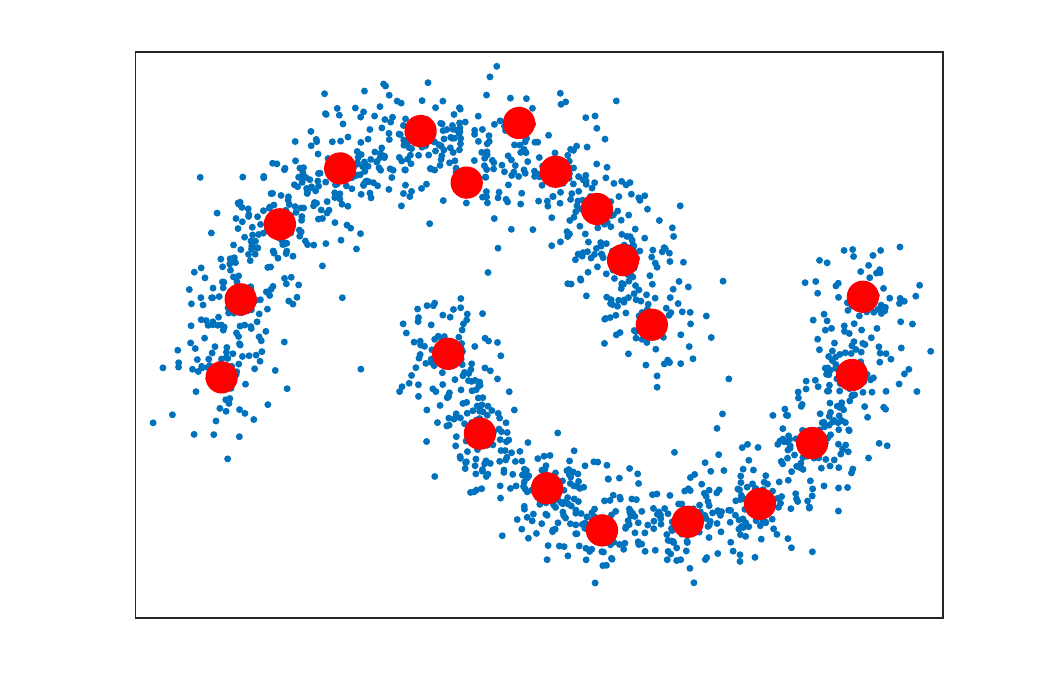}}}
    {\subfigure[]
      {\includegraphics[width=0.31\linewidth]{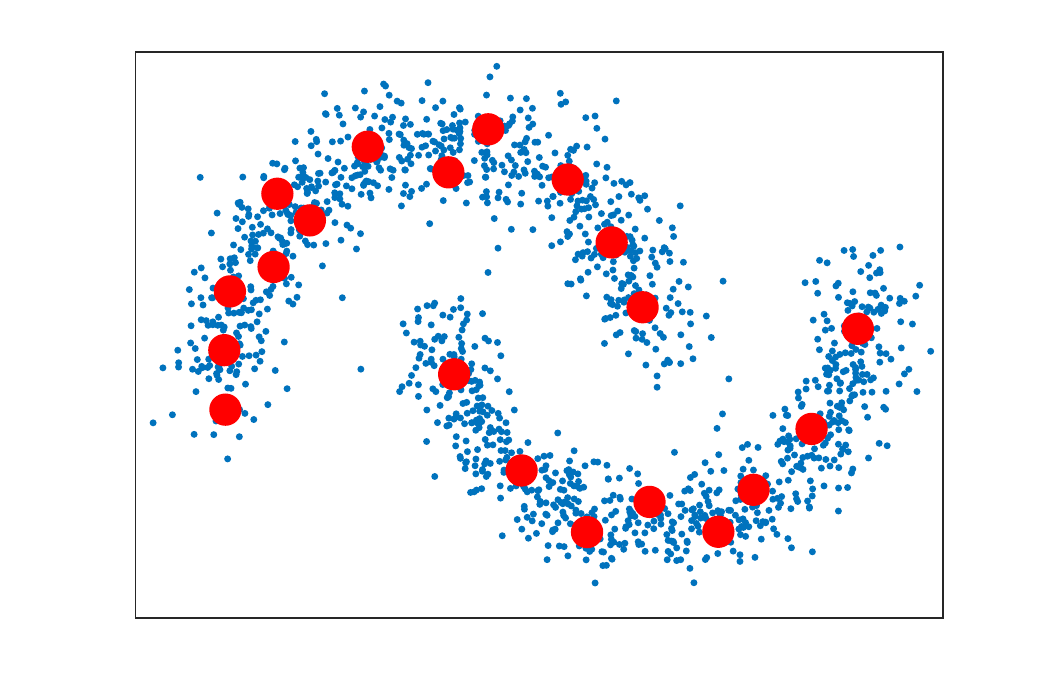}}}
    \caption{Comparison of the landmarks produced by (a) random selection, (b) $k$-means based selection, and (c) Divide-and-conquer based selection.}
    \label{fig:cmpSelStrategy_all3}
  \end{center}
\end{figure}

By introducing the divide-and-conquer based landmark selection, the complexity of landmark selection is reduced to $O(N\alpha d)$ from $O(Npdt)$ of $k$-means based selection.
Figure \ref{fig:cmpSelStrategy_all3} illustrates that the proposed divide-and-conquer based landmark selection can better represent data distribution than the random selection and has similar performance $k$-means based selection, but it has the lower complexity than $k$-means based selection.

\subsection{Approximate Similarity Matrix Construction}
\label{sec:approx_similiarty}

After landmark selection, the next object is to construct a similarity matrix between entire data points and the landmarks.
Instead of dense similarity matrix, we design a similarity matrix $B\in \mathbb{R}^{N \times p}$ according to $K$-nearest neighbor as follows: 
% We construct the sparse similarity sub-matrix using the widely used the Gaussian kernel \eqref{eq:gaussian kernel} as follow,
\begin{align}
  b_{ij} & = \begin{cases}\exp( \frac{-\left \| x_i -r_j \right \|^2}{2\sigma^2})
    ,  & \text{if $r_j\in N_K(x_i)$},\label{eq:gaussian kernel}
    \\
    0, & \text{otherwise,}
  \end{cases}
\end{align}
where $N_K(x_i)$ denotes the set of $K$-nearest landmarks of $x_i$ and $\sigma $ is the bandwidth of Gaussian kernel.
Note that there are only $NK$ non-zero entries in the sparse matrix $B$.

\begin{figure}
    \centering
    \includegraphics[width=0.5\linewidth]{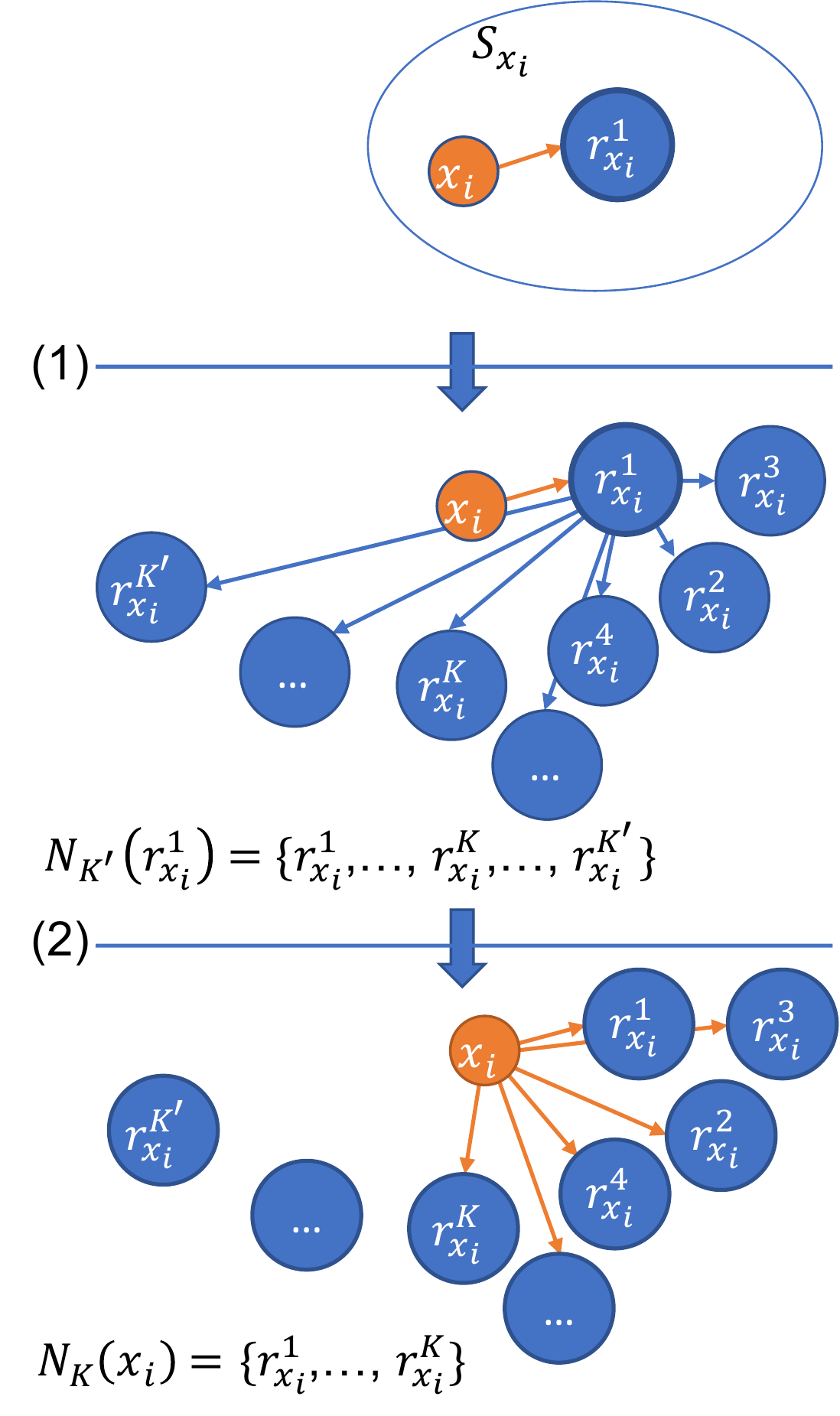}
    \caption{An illustration of our $K$-nearest landmarks approximation: 
    (1) Find the $K'$-nearest ($K'>K$) landmarks of $r_{x_i}^1$, where $x_i\in S_{x_i}$ and $r_{x_i}^1$ is the subset center of $S_{x_i}$;
    (2) Find the $K$-nearest landmarks among $K'$-nearest landmarks of $r_{x_i}^1$.
    }
    \label{fig:illustion_aknn}
\end{figure}

To estimate $N_K(x_i)$, we propose a new $K$-nearest landmarks approximation method.
The main idea is to use the subset centers' nature of landmarks to estimate the possible nearest candidates, as shown in Figure \ref{fig:illustion_aknn}.
Formally, we denote $S_{x_i}$ as the subset that $x_i$ belongs to, and the landmark $r_{x_i}^1$ as the center of $S_{x_i}$.
Since $r_{x_i}^1$ is the subset center of $x_i$, it essentially is the nearest landmark of $S_{x_i}$ according to \eqref{eq: opt}.
Take the advantage of this landmark nature, we search the $K$-nearest landmarks of each data point $x_i\in X$ according to the following two steps:
\paragraph{Step 1: Find $K'$ possibles candidates.}
As (1) of Figure \ref{fig:illustion_aknn} shows, we find the $K'$-nearest ($K'>K$) landmarks of $r_{x_i}^1$ and denoted them as ${N}_{K'}(r_{x_i}^{1})=\left\{ r_{x_i}^{1}, \dots, r_{x_i}^{K'}\right\}$.
Since the exact $K$-nearest landmarks of $x_i$ are highly possible closed to $r_{x_i}^{1}$, we treat ${N}_{K'}(r_{x_i}^1)$ as possible candidates set.
Empirically, the number of $K'$ is suggested to be several times larger than $K$, e.g., $K'= 10K$, to provide enough candidates to search $N_K(x_i)$.
\paragraph{Step 2: Search the $K$-nearest landmarks.}
As (2) of Figure \ref{fig:illustion_aknn} shows,
we search the $K$-nearest landmarks of $x_i$ among ${N}_{K'}(r_{x_i}^1)$ and denote them as $N_K(x_i)$.

After the $K$-nearest landmarks approximation, we compute the similarity matrix $B$ according to \eqref{eq:gaussian kernel}.
For all data points, the complexity of \textit{step 1} is $O(p^2(d+K))$ and \textit{step 2} is $O(NK(d+k))$.
The computational complexity of our similarity construction is $O(p^2(d+K)+ NK(d+k))$.
As $k, K \ll p \ll n$, the dominant term in the complexity is $O(NKd)$.
Compared with the exact similarity construction of $O(Npd)$ \cite{cai2014large,ye2018large}, our method is much faster.

\subsection{Bipartite Graph Partitioning}
\label{sec:bipartite_partition}
After obtaining the similarity matrix $B$, we conduct graph partitioning on the graph Laplacian.
The similarity matrix $B$ reflects the relationships between $X$ and ${R}$.
Therefore $B$ can be interpreted as the cross-similarity matrix of the bipartite graph:
\begin{equation}
  G=\{X, {R}, B\},
\end{equation}
where $X\cup{R}$ is the node-set.
As a result, the objective is changed to a bipartite graph partitioning problem.

We apply transfer cuts to efficiently compute the spectrum of graph Laplacian for spectral clustering. 
The $k$ bottom eigenvectors $u_1,u_2,\dots,u_k$ for $X$ side can be computed according to \eqref{eq:general_eigen_problem_reduced} and \eqref{eq:u}.
Let $U\in \mathbb{R}^{N\times k}$ be the matrix containing the vectors $u_1,\dots,u_k$ as columns, then matrix $U$ will be the spectrum of graph Laplacian.
In practice, we normalized $U$ by its 1-norm as $\tilde{U}$, then apply $k$-means clustering on $\tilde{U}$ to obtain the final clustering results \cite{ng2002spectral}.
The $k$-means clustering is then performed on this embedding to obtain the $k$ clusters as the final clustering result with $O(Nk^2t)$ computational complexity.
We summarize the divided-and-conquer based large-scale spectral clustering in Algorithm \ref{ag: divided-and-conquer based large-scale spectral clustering}.

\begin{algorithm}
  \label{ag: divided-and-conquer based large-scale spectral clustering}
  \caption{Divided-and-conquer based large-scale spectral clustering}
  \SetAlgoLined
  \KwIn{Dataset $X$, the number of landmarks $p$, selection rate $\alpha$, the number of nearest landmarks $K$, the number of cluster $k$;}
  \KwOut{Cluster labels;}
  Obtain $p$ landmarks by divide-and-conquer based landmark selection method;\\
  \ForEach(){$x_i \in X$}{
     Let $S_{x_i}$ be the subset that $x_i$ belongs to and $r_{x_i}^1$ be the center of $S_{x_i}$;\\
     Obtain $K'$-nearest ($K'>K)$ landmarks of $r_{x_i}^1$, denoted as ${N}_{K'}(r_{x_i}^1)$;\\
     Find the $K$-nearest landmarks of $x_i$ from ${N}_{K'}(r_{x_i}^1)$, denoted as $N_K(x_i)$;\\
    }
  Construct sparse similarity sub-matrix by \eqref{eq:gaussian kernel};\\
  Calculate $v$ by solve the eigen-problem \eqref{eq:general_eigen_problem_reduced};\\
  Obtain spectral embedding $u$ by \eqref{eq:u};\\
  Conduct $k$-means on the bottom $k$ eigenvectors of $u$ to obtain final clustering results.
\end{algorithm}

\section{Discussion}

\subsection{Computational Complexity Analysis}
\label{sec:dnc-sc_complexity}
In this section, we summary the computational cost of the proposed method in each phase.

The divide-and-conquer based landmark selection takes $O(N\alpha d)$ time. The similarity construction takes $O(NKd+p^2(d+K))$ time. The eigen-decomposition takes $O(NK(K+k)+p^3)$ time. The $k$-means discretization takes $O(Nk^2t)$ time. With consideration to $k,K,\alpha\ll p\ll N$, the overall computational complexity of DnC-SC is $O(N(\alpha d+K^2+Kk+Kd+k^2t)+p^3+p^2(d+K))$, where $O(N(\alpha)d)$ is the dominant term. Table~\ref{table:cmp_complexity} provides a comparison of computational complexity of our DnC-SC algorithm against several other large-scale spectral clustering algorithms.
The space complexity of DnC-SC is $O(NK)$.

\begin{table}%[!t]
  \centering
  \caption{Comparison of the computational complexity of several large-scale spectral clustering methods.}
  \label{table:cmp_complexity}
  \begin{threeparttable}
    \begin{tabular}{p{1.4cm}<{\centering}p{1.65cm}<{\centering}p{1.6cm}<{\centering}p{2.435cm}<{\centering}}
      \toprule
      Method                                 & landmark selection & Similarity construction  & Eigen-decomposition \\
      \midrule
      Nystr\"{o}m  & /                        & $O(Npd)$               & $O(Np+p^3)$         \\
      LSC-R  & /                        & $O(Npd)$               & $O(Np^2+p^3)$       \\
      LSC-K & $O(Npdt)$                & $O(Npd)$               & $O(Np^2+p^3)$       \\
      U-SPEC           & $O(p^2dt)$               & $O(Np^{\frac{1}{2}}d)$ & $O(NK(K+k)+p^3)$    \\
      DnC-SC                                 & $O(N\alpha d)$                 & $O(NKd)$               & $O(NK(K+k)+p^3)$    \\
      \bottomrule
    \end{tabular}
    \begin{tablenotes}
      \item[*] The final $k$-means is $O(Nk^2t)$ for each method.
    \end{tablenotes}
  \end{threeparttable}
\end{table}

\subsection{Relations with Other Methods}
As a large-scale spectral clustering method, the proposed method is closely related to the methods in \cite{cai2014large, huang2019ultra}. 
We compare the proposed method with the two methods to discuss the improvements of the proposed method. 

Firstly, we compare them on the landmark selection methods.
Both the two methods \cite{cai2014large, huang2019ultra} directly or indirectly apply $k$-means based landmark selection.
LSC-K method \cite{cai2014large} directly conduct $k$-means algorithm to select landmarks within a high computational complexity $O(Npdt)$.
While the U-SPEC \cite{huang2019ultra} indirectly conduct $k$-means algorithm on a random set of samples to select landmarks, which finds a balance between $k$-means and random selection within $O(p^2dt)$ time cost. 
Despite U-SPEC can efficiently find landmarks, it also has two limitations:
1) The quality of landmarks highly depends on how good the random set of samples is set up;
2) Since landmarks are not the centers for all data points, the center's nature of landmark can not be used to approximate the similarity matrix.
The proposed method uses the divide-and-conquer based landmark selection, which can effectively produce high-quality landmarks. 
We design a objection function \eqref{eq: opt} is to find the landmarks that best represent all data points with minimum RSS.
We then propose a divide-and-conquer strategy to divide \eqref{eq: opt} into local sub-problems and use light-$k$-means to effectively solve them.
Finally, we combine all sub-problems and obtain landmarks.
The our landmark selection produces landmarks within $O(N\alpha d)$ computational time.
Moreover, the our landmarks are essentially the centers of subsets for all data points, which can be used to approximate the similarity matrix next.

Secondly, we compare them on the similarity construction.
LSC-K needs to cost $O(Npd)$ to compute the dense similarity matrix at first to conduct the $K$-nearest neighbor sparse.
The U-SPEC method indirectly computes the sparse similarity sub-matrix in a coarse-to-fine mechanism to approximate the $K$-nearest landmarks within $O(Np^{\frac{1}{2}}d)$ time cost.
U-SPEC first cluster $p$ landmarks into $p^{\frac{1}{2}}$ clusters and then compute the distances between data points and $p^{\frac{1}{2}}$ cluster centers to find the possible range of nearest landmarks.
For the proposed method, since the landmarks essentially are the cluster centers of data points, we can easily identify a highly possible range of $K$-nearest landmarks according to the centers' nature of landmarks and find $K$-nearest landmarks in this range. 
The proposed $K$-nearest landmarks search method costs $O(NKd)$ computational time.

Overall, DnC-SC consists of divide-and-conquer based landmark selection, approximate similarity construction, and bipartite graph partition.
It conducts spectral clustering tasks within $O(N\alpha d)$ computational complexity and $O(NK)$ space complexity, which is faster than most existing large-scale spectral clustering methods. 

\section{Experiments}
\label{sec:experiment}

In this section, we conduct experiments on five real and five synthetic datasets to evaluate the performance of the proposed DnC-SC methods.
The comparison experiments against several state-of-the-art spectral clustering methods show better performance on clustering quality and efficiency for DnC-SC methods.
Besides that, the analysis of the parameters is performed.
For each experiment, the test method is repeated 20 times, and the average performance is reported.
All experiments are conducted in Matlab R2020a on a Mac Pro with 3 GHz 8-Core Intel Xeon E5 and 16 GB of RAM.

\subsection{Datasets and Evaluation Measures}

\begin{table}[!t]
  \centering
  \caption{Properties of the real and synthetic datasets.}
  \label{table:datasets}
  \begin{center}
    \begin{tabular}{p{1.2cm}<{\centering}|p{1.3cm}<{\centering}|p{1.5cm}<{\centering}p{1.2cm}<{\centering}p{1.2cm}<{\centering}}
      \toprule
      \multicolumn{2}{c|}{Dataset}         &\#Object     &\#Dimension      &\#Class\\
      \midrule
                                        & \emph{USPS}      & 9298       & 256 & 10 \\
      \multirow{5}{*}{\emph{Real}}      & \emph{PenDigits} & 10,992     & 16  & 10 \\
                                        & \emph{Letters}   & 20,000     & 16  & 26 \\
                                        & \emph{MNIST}     & 70,000     & 784 & 10 \\
                                        & \emph{Covertype} & 581,012    & 54  & 7  \\
      \midrule
      \multirow{5}{*}{\emph{Synthetic}} & \emph{TS-60K}    & 600,000    & 2   & 3  \\
                                        & \emph{TM-1M}     & 1,000,000  & 2   & 2  \\
                                        & \emph{TC-6M}     & 6,000,000  & 2   & 3  \\
                                        & \emph{CG-10M}    & 10,000,000 & 2   & 11 \\
                                        & \emph{FL-20M}    & 20,000,000 & 2   & 13 \\
      \bottomrule
    \end{tabular}
  \end{center}
\end{table}

\begin{figure}%[!t]
  \begin{center}
    {\subfigure[\emph{TS-60K} ($1\%$)]
      {\includegraphics[width=0.31\columnwidth]{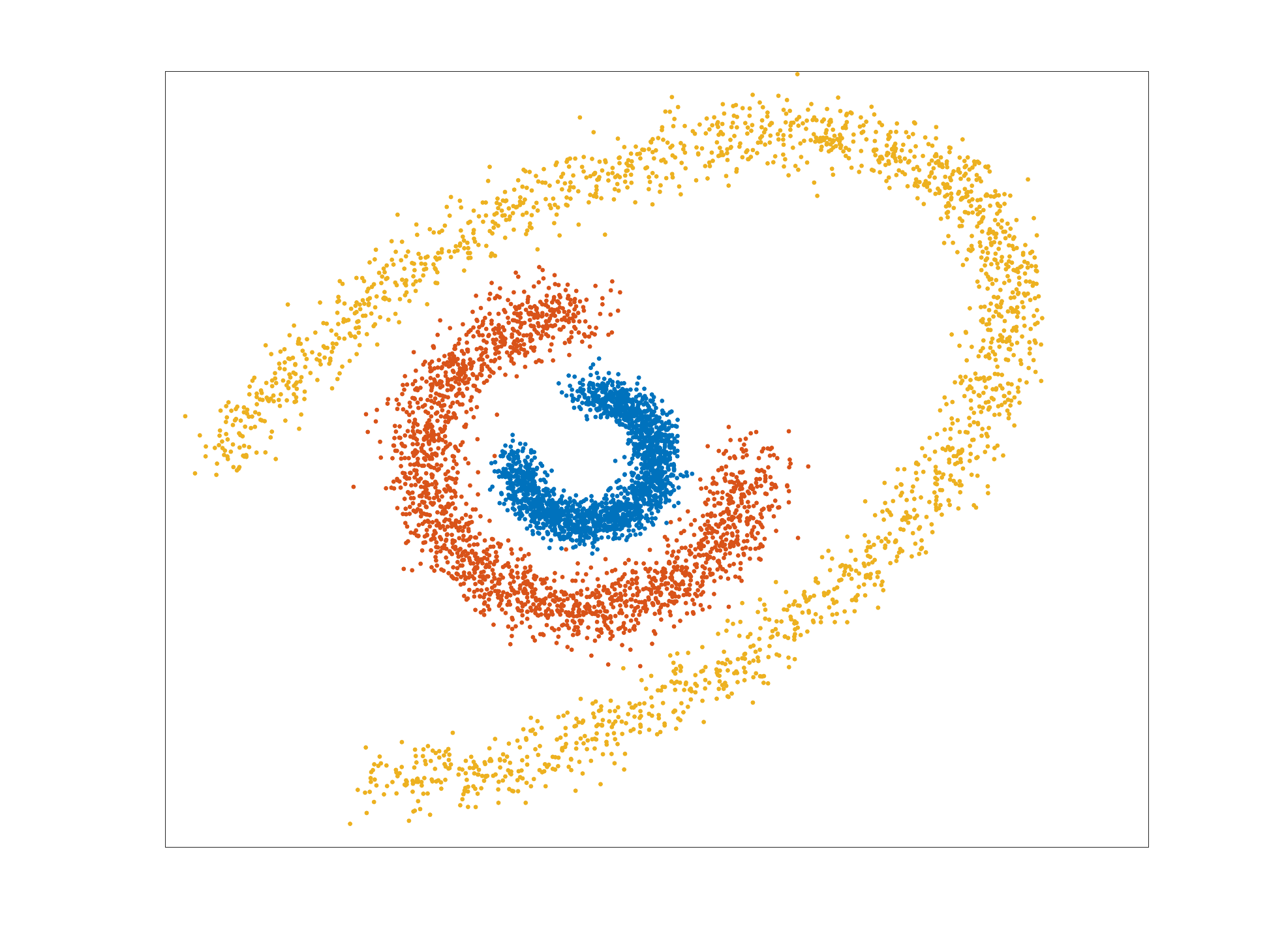}}}
    {\subfigure[\emph{TM-1M} ($1\%$)]
      {\includegraphics[width=0.31\columnwidth]{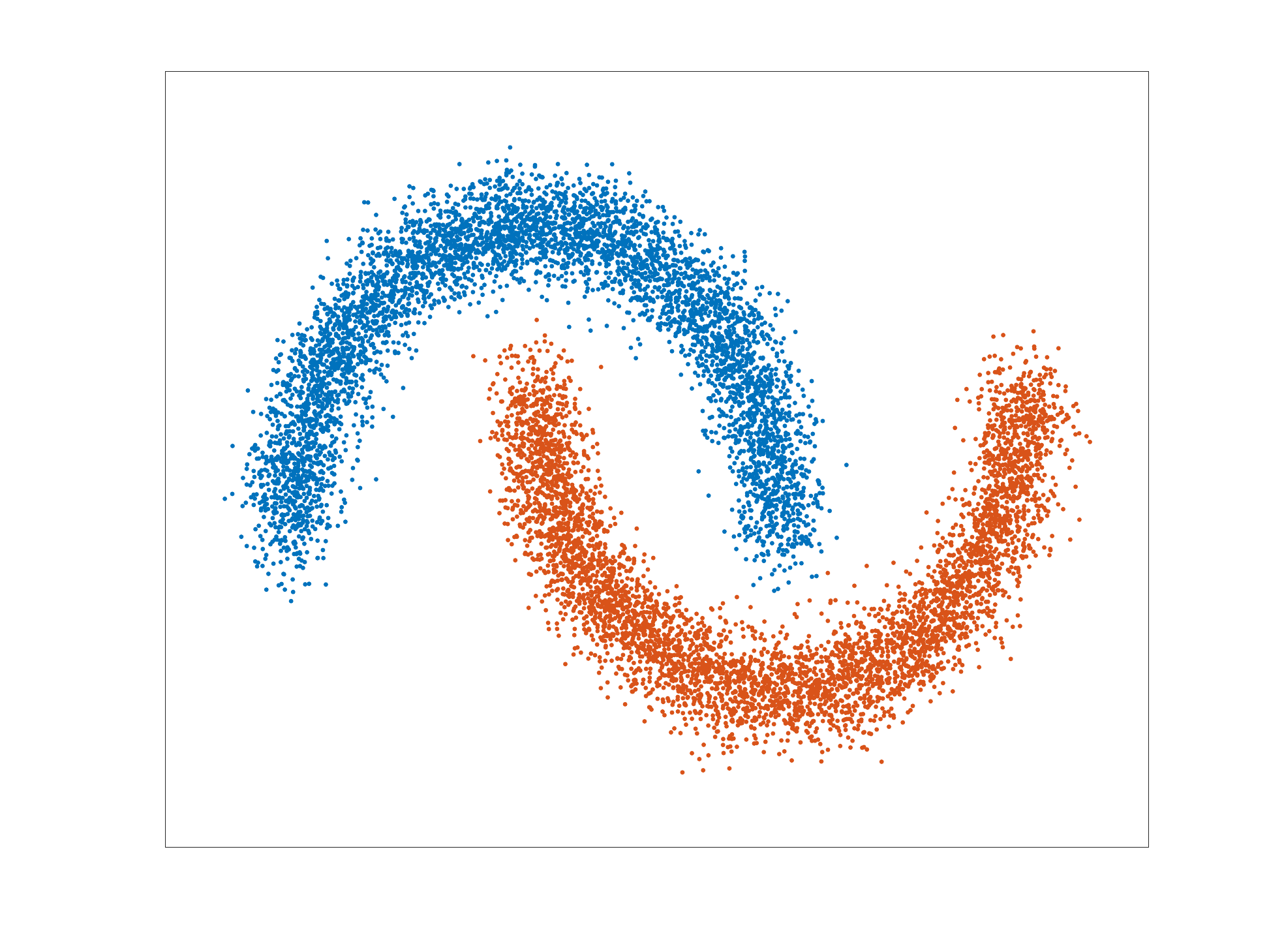}}}
    {\subfigure[\emph{TC-6M} ($1\%$)]
      {\includegraphics[width=0.31\columnwidth]{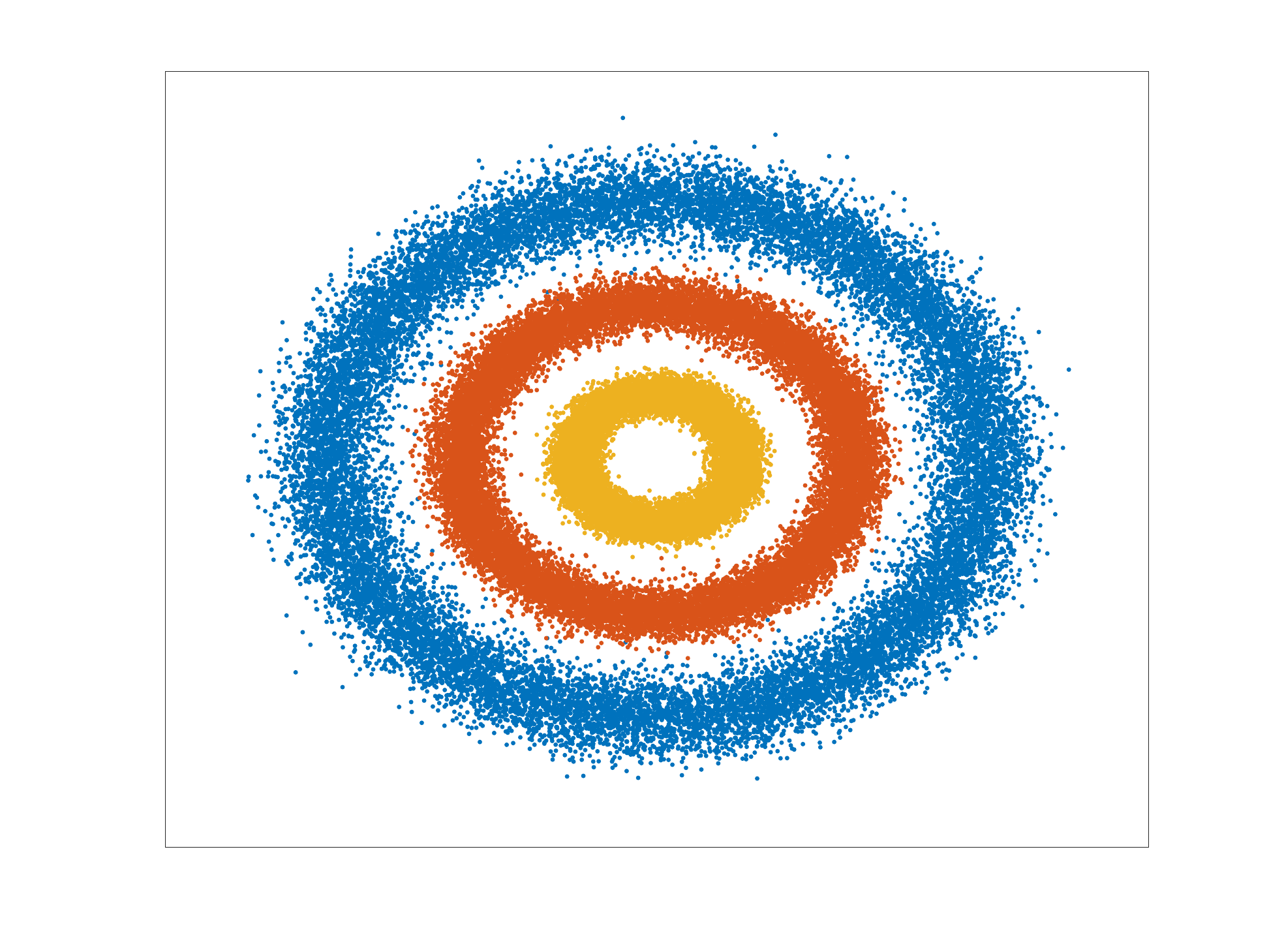}}}
    {\subfigure[\emph{CG-10M} ($0.1\%$)]
      {\includegraphics[width=0.31\columnwidth]{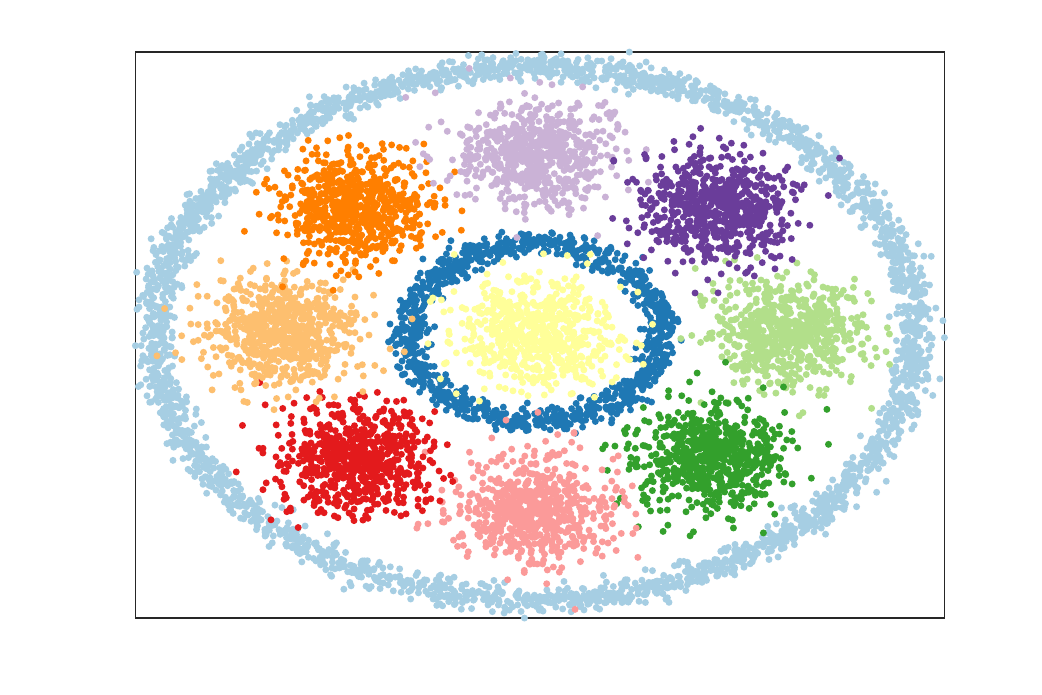}}}
    {\subfigure[\emph{FL-20M} ($0.1\%$)]
      {\includegraphics[width=0.31\columnwidth]{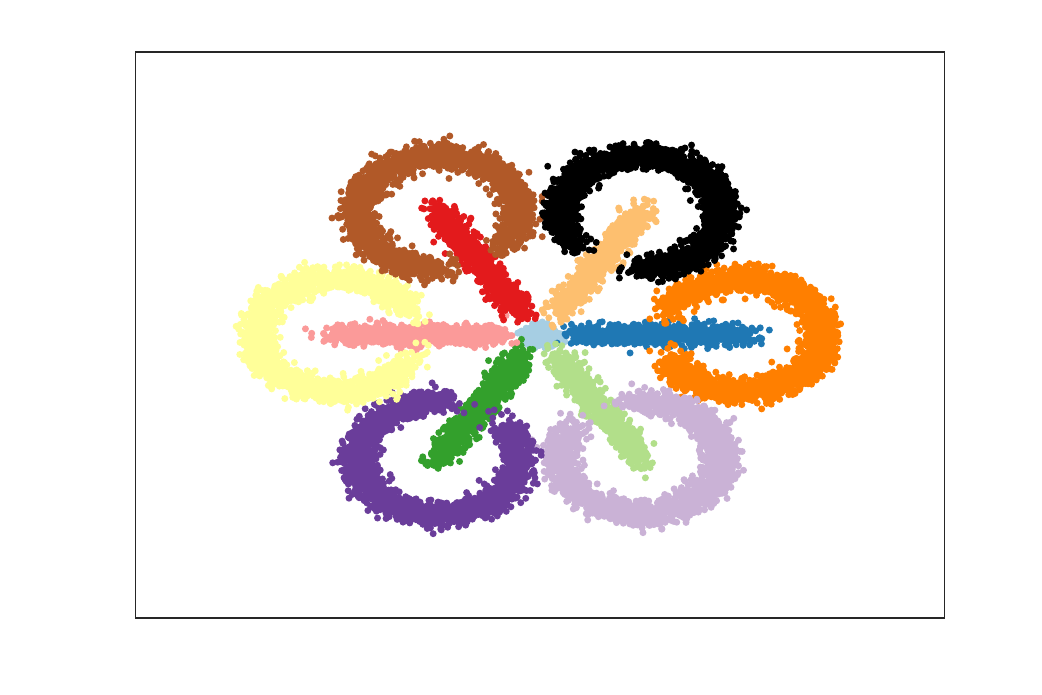}}}
    \caption{Illustration of the five synthetic datasets. Note that only a $1\%$ or $0.1\%$ samples of each dataset is plotted.}
    \label{fig:fiveSynDS}
  \end{center}
\end{figure}

Our experiments are conducted on ten large-scale datasets, varying from nine thousand to as large as twenty million data points. Specifically, the five real datasets are \emph{USPS} \cite{cai2010graph} \footnote{\label{cai_deng_data} http://www.cad.zju.edu.cn/home/dengcai/Data/MLData.html}, \emph{PenDigits} \cite{asuncion2007uci} \footnote{https://archive.ics.uci.edu/ml/datasets/Pen-Based+Recognition+of+Handwritten+Digits}, \emph{Letters} \cite{frey1991letter} \footnote{https://archive.ics.uci.edu/ml/datasets/Letter+Recognition}, \emph{MNIST} \cite{cai2011speed} \footref{cai_deng_data}, and \emph{Covertype} \cite{blackard1999comparative} \footnote{https://archive.ics.uci.edu/ml/datasets/covertype}. The five synthetic datasets are \emph{Two Spiral-60K} (\emph{TS-60K}), \emph{Two
  Moons-1M} (\emph{TM-1M}), \emph{Three Circles-6M} (\emph{TC-6M}), \emph{Circles and Gaussians-10M} (\emph{CG-10M}) \cite{huang2019ultra} \footnote{\label{huang}https://www.researchgate.net/publication/330760669}, \emph{Flower-20M} (\emph{FL-20M}) \cite{huang2019ultra} \footref{huang}. Figure ~\ref{fig:fiveSynDS} shows the synthetic datasets.
The properties of the datasets are summarized in Table~\ref{table:datasets}.

We adopt two widely used evaluation metrics, i.e., Normalized Mutual Information (NMI) \cite{slonim2000agglomerative} and Accuracy (ACC) \cite{yan2009fast}, to evaluate the clustering results.
Let $X=[x_1,x_2,...,x_n]$ be the data matrix. For each data point $x_i$, denote $t_i$ and $c_i$ as the cluster label of ground truth and obtained cluster label from clustering methods, respectively. The ACC is defined as:
\begin{equation}
  \text{ACC}=\frac{\sum_{i=1}^n \delta(t_i,\text{map}(c_i))}{n},
\end{equation}
where $n$ is the number of data  and $\delta(t_i,c_i)$ is a function to check $t_i$ and $c_i$ are equal or not,  returning 1 if equals otherwise returning 0. The map$(c_i)$ is a best mapping function that maps each predicted label to the most possibly true cluster label by permuting operations \cite{xu2003document}.

Let $T$ denote a set of clusters of ground truth and $C$ obtained from clustering methods. Mutual information (MI) is defined as
\begin{equation}
  MI(T,C)=\sum_{t_i\in T,c_i\in C}p(t_i,c_i)\text{ln}\frac{p(t_i,c_i)}{p(t_i)p(c_i)},
\end{equation}
where $p(t_i)$ and $p(c_i)$ are marginal probabilities that a sample happens to belong to cluster $t_i$ or $c_i$ while $p(t_i,c_i)$ is the joint probabilities that a sample happens to belong to cluster both $t_i$ and $c_i$.
The NMI is the normalization of MI by the joint entropy as follow:
\begin{equation}
  \label{NMI}
  NMI(T,C)= \frac{\sum_{t_i\in T,c_i\in C}p(t_i,c_i)\text{ln}\frac{p(t_i,c_i)}{p(t_i)p(c_i)}}
  {-\sum_{t_i\in T,c_i\in C}p(t_i,c_i)\text{ln}(p(t_i,c_i))}
  ,\end{equation}

A better clustering result will provide a larger value of NMI/ACC. Both NMI and ACC are in the range of $[0,1]$.

\subsection{Baseline Methods and Experimental Settings}

In this experiment, we compare the proposed method with two baseline clustering methods, which are $k$-means clustering and spectral clustering (SC) \cite{chen2010parallel}, as well as six state-of-the-art large-scale spectral clustering methods. The compared spectral clustering methods are listed as follows:

\begin{enumerate}
  \item \textbf{SC} \cite{chen2010parallel}: original spectral clustering \footnote{\label{psc}http://alumni.cs.ucsb.edu/~wychen/sc.html}.
  \item \textbf{Nystr\"{o}m} \cite{fowlkes2004spectral}: Nystr\"{o}m spectral clustering \footref{psc}.
  \item \textbf{LSC-K} \cite{cai2014large}: landmark based spectral clustering using $k$-means based landmark selection \footnote{\label{LSC}http://www.cad.zju.edu.cn/home/dengcai/Data/Clustering.html}.
  \item \textbf{LSC-R} \cite{cai2014large}: landmark based spectral clustering using random landmark selection \footref{LSC}.
  \item \textbf{LSC-KH} \cite{ye2018large}: Landmark-based spectral clustering using $k$-means partition to find the hubs as the landmarks \footnote{\label{mycode}https://github.com/Li-Hongmin/MyPaperWithCode}.
  \item \textbf{LSC-RH} \cite{ye2018large}: Landmark-based spectral clustering using random partition to find the hubs as the landmarks \footref{mycode}.
  \item \textbf{U-SPEC} \cite{huang2019ultra}: Ultra-Scalable Spectral Clustering \footref{huang}.
\end{enumerate}

There are several common parameters among the methods mentioned above. We set these parameters as follow:
\begin{itemize}
  \item We set the number of landmarks or representatives as $p=1000$ for DnC-SC, U-SPEC, Nystr\"{o}m, LSC-K, and LSC-R methods. The parameter analysis on $p$ will be further conducted in Section~\ref{sec:para_p}.
  \item We set the $K=5$ for the number of nearest neighbors for DnC-SC, U-SPEC, LSC-K, and LSC-R.
        The parameter analysis on $K$ will be further conducted in Section~\ref{sec:para_K}.
  \item The DnC-SC method has a unique parameter $\alpha$. In the experiments, $\alpha = 200$ is used for the datasets whose size is less than 100,000, otherwise $\alpha=50$.
\end{itemize}

\begin{table*}[]
  \centering
  \caption{Clustering performance (ACC\% $\pm$ std) for large-scale spectral clustering methods}
  \label{table:compare_spectrals_acc}
  \resizebox{0.95\textwidth}{!}{%
    \begin{threeparttable}
      \begin{tabular}{@{}c||c||ccccccc|c@{}}
        \toprule
        Dataset          & KM                    & SC                 & Nystr{\"{o}}m       & LSC-K                       & LSC-R               & LSC-KH              & LSC-RH              & U-SPEC                      & DnC-SC                      \\
        \midrule
        \emph{USPS}      & 67.01$_{\pm 0.70}$    & 73.21$_{\pm 3.10}$ & 69.47$_{\pm 1.38}$  & 74.02$_{\pm 7.34}$          & 73.90$_{\pm 4.42}$  & 73.66 $_{\pm 5.18}$ & 73.89 $_{\pm 4.27}$ & 80.79$_{\pm 3.13}$          & \textbf{82.55$_{\pm 1.96}$} \\
        \emph{PenDigits} & 64.40$_{\pm 4.73}$    & 67.23$_{\pm 4.35}$ & 72.46$_{\pm 0.18}$  & 82.30$_{\pm 2.95}$          & 81.55$_{\pm 3.79}$  & 82.17 $_{\pm 4.09}$ & 81.55 $_{\pm 5.12}$ & 81.74$_{\pm 4.95}$          & \textbf{82.27$_{\pm 1.33}$} \\
        \emph{Letters}   & 25.56$_{\pm 1.00}$    & 31.21$_{\pm 0.76}$ & 31.30$_{\pm 0.40}$  & 33.20$_{\pm 2.52}$          & 32.34$_{\pm 0.15}$  & 31.13 $_{\pm 0.88}$ & 31.60 $_{\pm 1.67}$ & 33.20$_{\pm 1.16}$          & \textbf{33.54$_{\pm 1.21}$} \\
        \emph{MINST}     & 56.60$_{\pm 2.71}$    & N/A                & 57.02$_{\pm 3.66}$  & \textbf{80.96$_{\pm 0.10}$} & 62.00$_{\pm 3.99}$  & 66.59 $_{\pm 5.33}$ & 67.60 $_{\pm 6.02}$ & 72.00$_{\pm 3.33}$          & 74.24$_{\pm 2.14}$          \\
        \emph{Covertype} & 24.04$_{\pm 0.22}$    & N/A                & 21.65$_{\pm 1.30}$  & \textbf{24.71$_{\pm 1.45}$} & 23.62$_{\pm 1.10}$  & N/A                 & N/A                 & 24.40$_{\pm 2.20}$          & 23.48$_{\pm 1.86}$          \\
        \emph{TS-60K}    & 56.96$_{\pm 0.00}$    & N/A                & 55.94$_{\pm 10.17}$ & 70.37$_{\pm 4.57}$          & 62.91$_{\pm 13.74}$ & N/A                 & N/A                 & 65.78$_{\pm 13.63}$         & \textbf{81.00$_{\pm 9.29}$} \\
        \emph{TM-1M}     & 75.21$_{\pm 0.00}$    & N/A                & 64.63$_{\pm 8.40}$  & 51.76$_{\pm 0.54}$          & 66.41$_{\pm 26.68}$ & N/A                 & N/A                 & \textbf{99.96$_{\pm 0.01}$} & \textbf{99.96$_{\pm 0.01}$} \\
        \emph{TC-6M}     & 33.34$_{\pm 0.00}$    & N/A                & N/A                 & N/A                         & N/A                 & N/A                 & N/A                 & 99.86$_{\pm 0.03}$          & \textbf{99.87$_{\pm 0.02}$} \\
        \emph{CG-10M}    & 60.47$_{\pm 2.91}$    & N/A                & N/A                 & N/A                         & N/A                 & N/A                 & N/A                 & 66.77 $_{\pm 3.97}$         & \textbf{66.83$_{\pm 4.61}$} \\
        \emph{FL-20M}    & 50.07$_{\pm 2.91}$    & N/A                & N/A                 & N/A                         & N/A                 & N/A                 & N/A                 & 80.17 $_{\pm 3.97}$         & \textbf{81.90$_{\pm 5.61}$} \\
        \midrule
        \midrule
        Avg. score       & \multicolumn{1}{c}{-} & N/A                & N/A                 & N/A                         & N/A                 & N/A                 & N/A                 & 70.45                       & \textbf{72.5}9              \\
        \midrule
        \midrule
        Avg. rank        & \multicolumn{1}{c}{-} & 5.80               & 5.10                & 2.80                        & 3.90                & 4.70                & 4.5                 & 2.30                        & \textbf{1.50}               \\
        \bottomrule
      \end{tabular}%
      \begin{tablenotes}
        \item[*] N/A denotes the case when MATLAB reports the error of out of memory.
      \end{tablenotes}
    \end{threeparttable}
  }

\end{table*}

\begin{table*}[]
  \centering
  \caption{Clustering performance (NMI\% $\pm$ std) for large-scale spectral clustering methods}
  \label{table:compare_spectrals_nmi}
  \resizebox{0.95\textwidth}{!}{%
    \begin{tabular}{@{}c||c||ccccccc|c@{}}
      \toprule
      Dataset           & KM                    & SC                 & Nystr{\"{o}}m       & LSC-K                       & LSC-R               & LSC-KH              & LSC-RH              & U-SPEC              & DnC-SC                      \\
      \midrule
      \emph{USPS     }  & 61.28$_{\pm 0.42}$    & 77.90$_{\pm 0.55}$ & 65.07$_{\pm 1.23}$  & 81.37$_{\pm 1.92}$          & 76.22$_{\pm 0.76}$  & 76.41 $_{\pm 1.74}$ & 76.24 $_{\pm 1.00}$ & 81.86$_{\pm 1.95}$  & \textbf{82.86$_{\pm 0.21}$} \\
      \emph{PenDigits}  & 67.65$_{\pm 1.18}$    & 71.70$_{\pm 1.21}$ & 65.48$_{\pm 0.21}$  & 80.78$_{\pm 0.55}$          & 79.15$_{\pm 1.74}$  & 80.78 $_{\pm 0.55}$ & 79.15 $_{\pm 1.74}$ & 81.68$_{\pm 2.33}$  & \textbf{82.01$_{\pm 1.08}$} \\
      \emph{Letters  }  & 34.95$_{\pm 0.54}$    & 34.96$_{\pm 0.63}$ & 40.07$_{\pm 0.41}$  & 44.68$_{\pm 1.56}$          & 42.36$_{\pm 0.86}$  & 42.31 $_{\pm 0.75}$ & 42.20 $_{\pm 1.30}$ & 45.11$_{\pm 0.54}$  & \textbf{45.37$_{\pm 0.85}$} \\
      \emph{MINST    }  & 50.90$_{\pm 1.10}$    & N/A                & 49.05$_{\pm 1.55}$  & \textbf{76.81$_{\pm 0.18}$} & 62.53$_{\pm 1.87}$  & 65.08 $_{\pm 2.16}$ & 65.14 $_{\pm 2.47}$ & 69.15$_{\pm 0.76}$  & 72.00$_{\pm 0.51}$          \\
      \emph{Covertype}  & 7.55$_{\pm 0.00}$     & N/A                & 7.98$_{\pm 0.98}$   & \textbf{9.21$_{\pm 0.14}$}  & 8.06$_{\pm 0.07}$   & N/A                 & N/A                 & 8.19$_{\pm 0.04}$   & 8.30$_{\pm 0.30}$           \\
      \emph{TS-60K   }  & 22.22$_{\pm 0.00}$    & N/A                & 21.64$_{\pm 14.69}$ & 39.16$_{\pm 9.25}$          & 39.80$_{\pm 17.52}$ & N/A                 & N/A                 & 62.52$_{\pm 17.01}$ & \textbf{73.84$_{\pm 5.08}$} \\
      \emph{TM-1M    }  & 19.21$_{\pm 0.00}$    & N/A                & 8.03$_{\pm 8.58}$   & 0.10$_{\pm 0.05}$           & 28.11$_{\pm 48.63}$ & N/A                 & N/A                 & 99.52$_{\pm 0.08}$  & \textbf{99.52$_{\pm 0.05}$} \\
      \emph{TC-6M    }  & 34.95$_{\pm 0.54}$    & N/A                & N/A                 & N/A                         & N/A                 & N/A                 & N/A                 & 99.14$_{\pm 0.19}$  & \textbf{99.15$_{\pm 0.08}$} \\
      \emph{CG-10M    } & 64.94$_{\pm 1.61}$    & N/A                & N/A                 & N/A                         & N/A                 & N/A                 & N/A                 & 79.98$_{\pm 2.10}$  & \textbf{80.91$_{\pm 3.59}$} \\
      \emph{FL-20M}     & 65.02$_{\pm 2.91}$    & N/A                & N/A                 & N/A                         & N/A                 & N/A                 & N/A                 & 86.77 $_{\pm 3.97}$ & \textbf{87.67$_{\pm 3.18}$} \\
      \midrule
      \midrule
      Avg. score        & \multicolumn{1}{c}{-} & N/A                & N/A                 & N/A                         & N/A                 & N/A                 & N/A                 & 71.39               & \textbf{72.39}              \\
      \midrule
      \midrule
      Avg. rank         & \multicolumn{1}{c}{-} & 5.40               & 5.30                & 3.10                        & 4.20                & 4.50                & 4.70                & 2.00                & \textbf{1.40}               \\
      \bottomrule
    \end{tabular}%
  }
\end{table*}

\begin{table*}[]
  \centering
  \caption{Time costs(s) of large-scale spectral clustering methods.}
  \label{table:compare_spectrals_time}
  \resizebox{0.95\textwidth}{!}{%
    \begin{tabular}{@{}c||c||ccccccc|c@{}}
      \toprule
      Dataset           & KM                    & SC                 & Nystr{\"{o}}m          & LSC-K                   & LSC-R                      & LSC-KH              & LSC-RH              & U-SPEC                  & DnC-SC                         \\
      \midrule
      \emph{USPS     }  & 0.37$_{\pm 0.18}$     & 3.15$_{\pm 0.18}$  & 1.44$_{\pm 0.04}$      & 1.35$_{\pm 0.09}$       & \textbf{0.64$_{\pm 0.14}$} & 0.71 $_{\pm 0.06}$  & 0.88 $_{\pm 0.07}$  & 3.36$_{\pm 0.25}$       & 1.25$_{\pm 0.07}$              \\
      \emph{PenDigits}  & 0.05$_{\pm 0.05}$     & 3.15$_{\pm 0.11}$  & 1.61$_{\pm 0.10}$      & 1.20$_{\pm 0.37}$       & 0.77$_{\pm 0.34}$          & 0.71 $_{\pm 0.05}$  & 0.68 $_{\pm 0.07}$  & 2.07$_{\pm 0.95}$       & \textbf{0.64$_{\pm 0.08}$}     \\
      \emph{Letters  }  & 0.26$_{\pm 0.05}$     & 13.67$_{\pm 2.35}$ & 4.70$_{\pm 0.17}$      & 3.89$_{\pm 0.28}$       & 2.03$_{\pm 0.34}$          & 2.26 $_{\pm 0.17}$  & 2.63 $_{\pm 0.28}$  & 1.58$_{\pm 0.06}$       & \textbf{0.90$_{\pm 0.10}$}     \\
      \emph{MINST    }  & 21.40$_{\pm 1.02}$    & N/A                & 6.54$_{\pm 0.11}$      & 17.29$_{\pm 0.82}$      & 5.80$_{\pm 0.31}$          & 18.04 $_{\pm 2.35}$ & 15.38 $_{\pm 2.43}$ & 11.96$_{\pm 0.32}$      & \textbf{5.11$_{\pm 0.51}$}     \\
      \emph{Covertype}  & 14.02$_{\pm 4.39}$    & N/A                & 571.69$_{\pm 144.60}$  & 354.74$_{\pm 90.80}$    & 41.00$_{\pm 12.38}$        & N/A                 & N/A                 & 15.96$_{\pm 1.44}$      & \textbf{13.15$_{\pm 3.00}$}    \\
      \emph{TS-60K   }  & 1.39$_{\pm 0.18}$     & N/A                & 1283.33$_{\pm 248.12}$ & 167.29$_{\pm 39.99}$    & 16.35$_{\pm 1.62}$         & N/A                 & N/A                 & 17.36$_{\pm 20.89}$     & \textbf{4.01$_{\pm 1.16}$}     \\
      \emph{TM-1M    }  & 1.12$_{\pm 0.17}$     & N/A                & 3401.61$_{\pm 410.03}$ & 3997.21$_{\pm 1436.73}$ & 591.02$_{\pm 127.86}$      & N/A                 & N/A                 & 7.85$_{\pm 0.21}$       & \textbf{6.46$_{\pm 1.13}$}     \\
      \emph{TC-6M    }  & 35.23$_{\pm 1.72}$    & N/A                & N/A                    & N/A                     & N/A                        & N/A                 & N/A                 & 30.46$_{\pm 1.52}$      & \textbf{25.05$_{\pm 3.04}$}    \\
      \emph{CG-10M    } & 134.42$_{\pm 9.28}$   & N/A                & N/A                    & N/A                     & N/A                        & N/A                 & N/A                 & 381.72$_{\pm 72.24}$    & \textbf{281.05$_{\pm 77.04}$}  \\
      \emph{FL-20M}     & 311.94$_{\pm 2.91}$   & N/A                & N/A                    & N/A                     & N/A                        & N/A                 & N/A                 & 1530.30 $_{\pm 578.44}$ & \textbf{837.38$_{\pm 213.70}$} \\
      \midrule
      \midrule
      Avg. score        & \multicolumn{1}{c}{-} & N/A                & N/A                    & N/A                     & N/A                        & N/A                 & N/A                 & 165.96                  & \textbf{117.50}                \\
      \midrule
      \midrule
      Avg. rank         & \multicolumn{1}{c}{-} & 5.80               & 4.50                   & 4.40                    & 2.60                       & 4.30                & 4.20                & 3.30                    & \textbf{1.50 }                 \\
      \bottomrule
    \end{tabular}%
  }
\end{table*}

\subsection{Comparison with Large-scale Spectral Clustering Methods}
\label{sec:cmp_spectral}

In this section, we compare the proposed DnC-SC method with five state-of-the-art spectral clustering methods, as well as the $k$-means clustering and original spectral clustering methods as the baseline methods.

We report the experimental results in Tables ~\ref{table:compare_spectrals_acc}, ~\ref{table:compare_spectrals_nmi} and ~\ref{table:compare_spectrals_time}, where we use N/A to denote the case when MATLAB reports the error of out of memory.
Only two methods (proposed DnC-SC and U-SPEC) pass all datasets because they can approximately compute the similarity matrix within a limited memory.
The proposed DnC-SC method achieves the best clustering performance of both ACC and NMI ten times on ten benchmark datasets according to Table ~\ref{table:compare_spectrals_acc} and ~\ref{table:compare_spectrals_nmi}.
The proposed DnC-SC method achieves the best efficiency nine times on ten benchmark datasets according to Table ~\ref{table:compare_spectrals_time}.

In addition, we report the average performance score and rank for each method in Tables ~\ref{table:compare_spectrals_acc}, ~\ref{table:compare_spectrals_nmi} and ~\ref{table:compare_spectrals_time}.
The proposed DnC-SC method achieves the best average scores of both ACC and NMI.
The DnC-SC method shows average ranks of 1.50 of ACC and 1.40 of NMI, which implies the best clustering quality in all spectral clustering methods.
Moreover, the DnC-SC method costs much less average time than the other competitors and achieves a rank of 1.50, which implies the most efficient method in this experiment.
Overall, the proposed DnC-SC method shows significant effectiveness and efficiency comparing with six state-of-the-art large-scale spectral clustering methods.

\begin{table*}%[!t]
  \centering
  \caption{Clustering performance (ACC(\%), NMI(\%), and time costs(s)) for different methods by varying number of landmark $p$.}
  \label{table:compare_para_p}
  \begin{threeparttable}
    \begin{tabular}{m{0.08\textwidth}<{\centering}|m{0.2\textwidth}<{\centering}m{0.2\textwidth}<{\centering}m{0.2\textwidth}<{\centering}m{0.2\textwidth}<{\centering}}
      \toprule
      \emph{Dataset} & \emph{Letters} & \emph{MNIST} & \emph{TS-60K} & \emph{TM-1M} \\
      \midrule
      \multirow{1}{*}{ACC}
      &\includegraphics[width=0.18\textwidth]{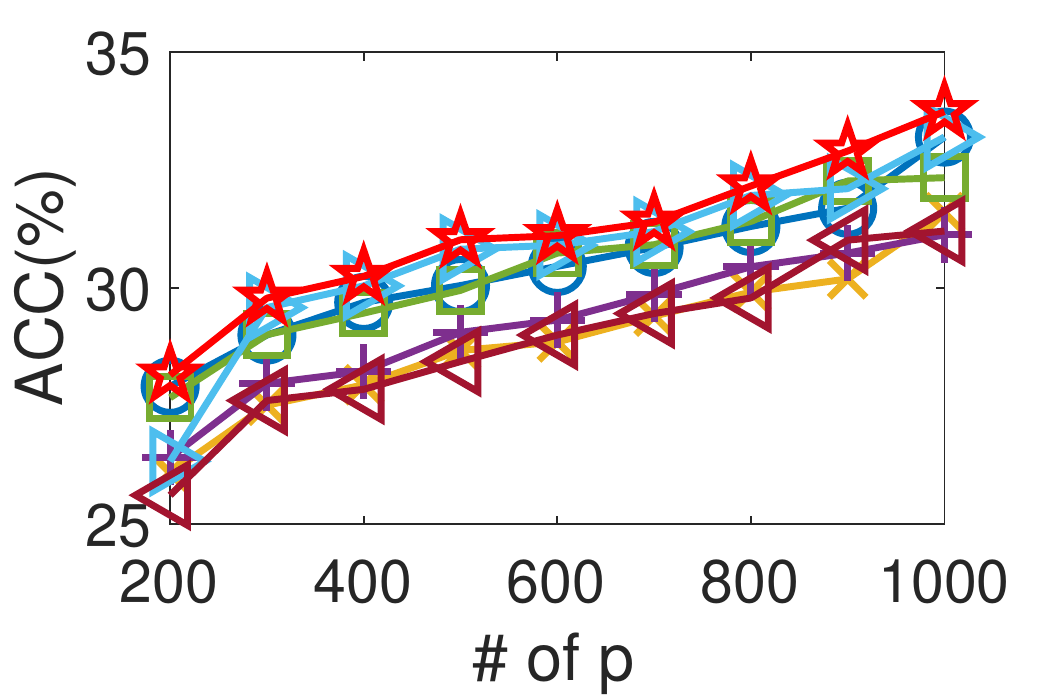}
      &\includegraphics[width=0.18\textwidth]{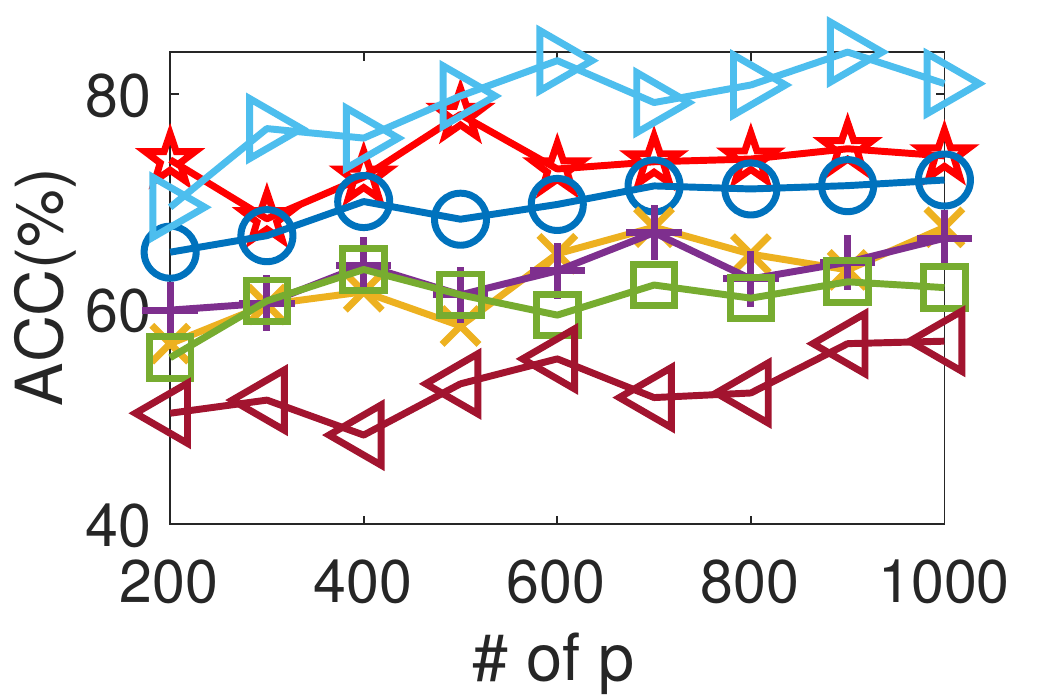}
      &\includegraphics[width=0.18\textwidth]{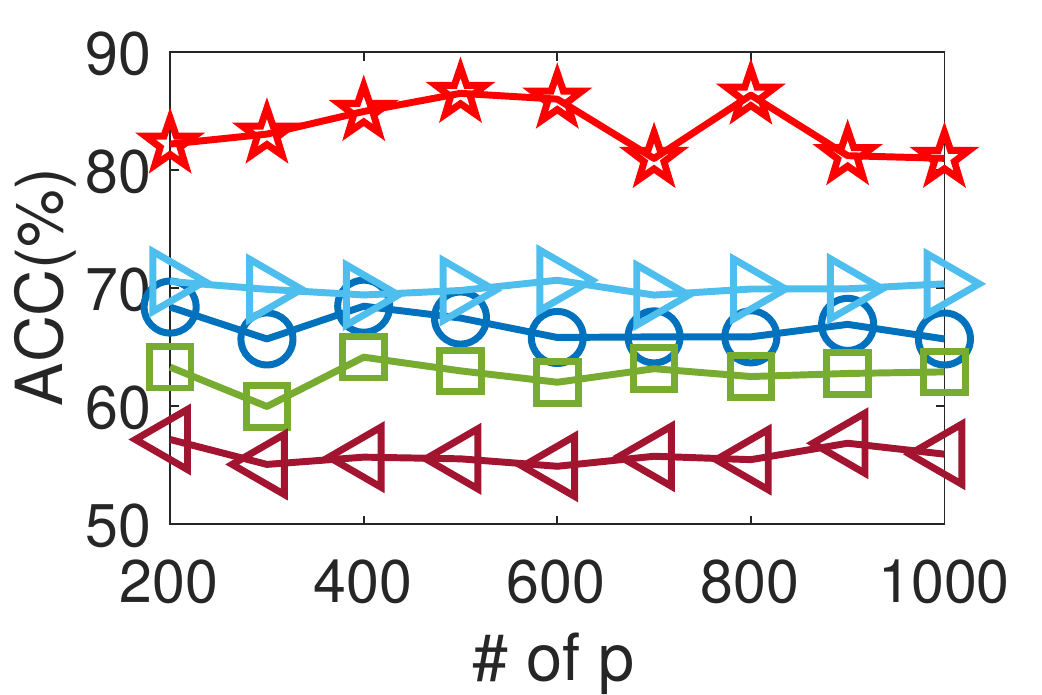}
      &\includegraphics[width=0.18\textwidth]{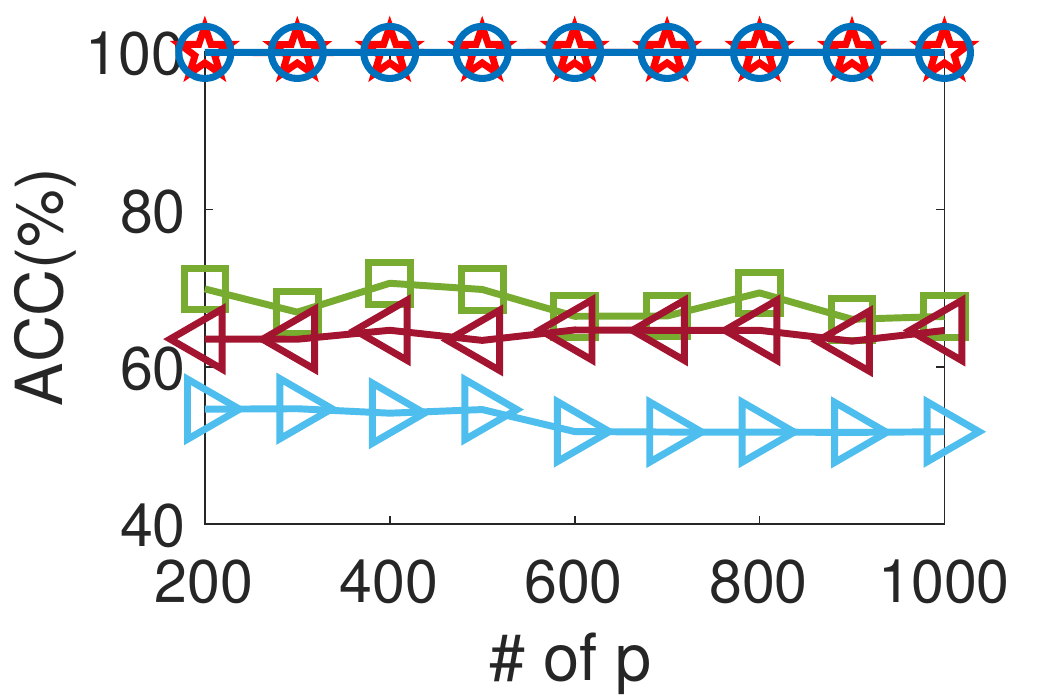}\\
      NMI
      &\includegraphics[width=0.18\textwidth]{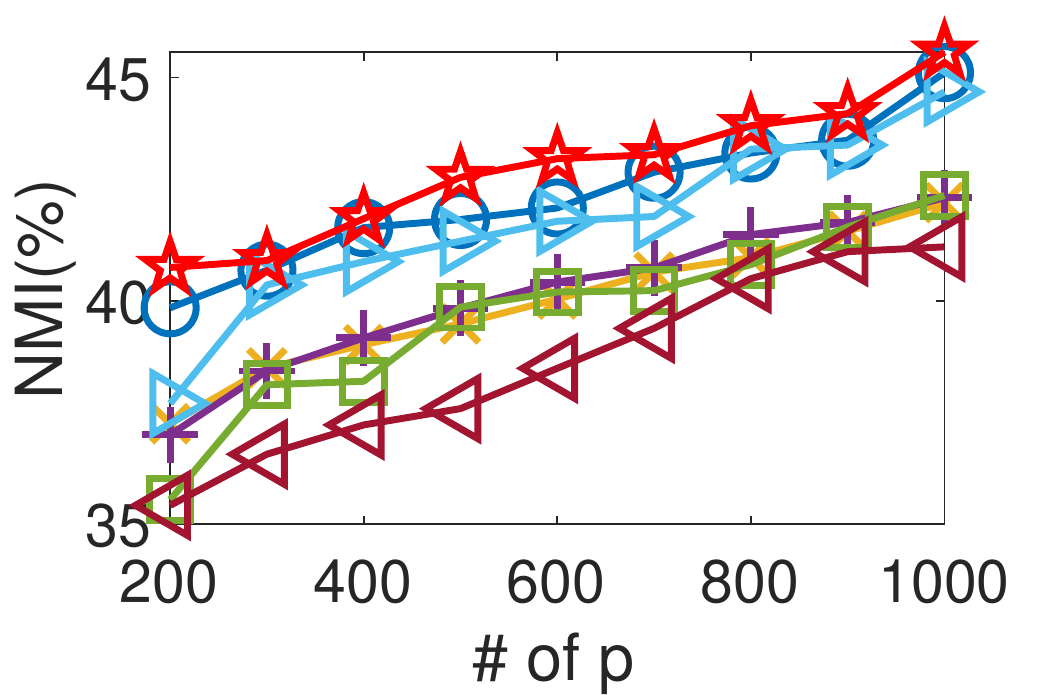}
      &\includegraphics[width=0.18\textwidth]{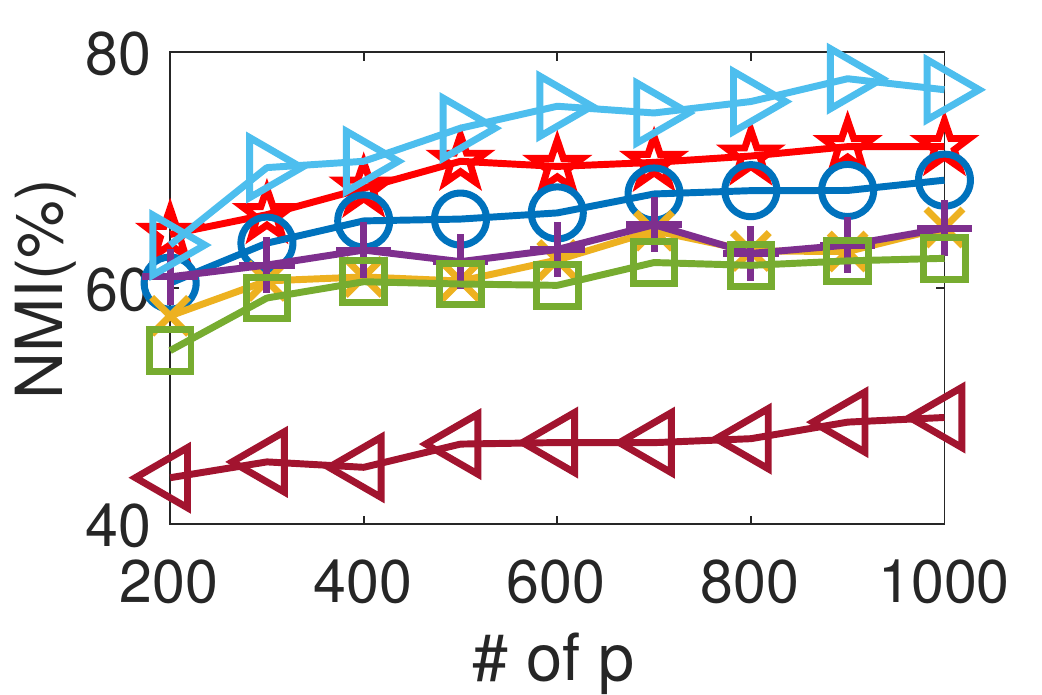}
      &\includegraphics[width=0.18\textwidth]{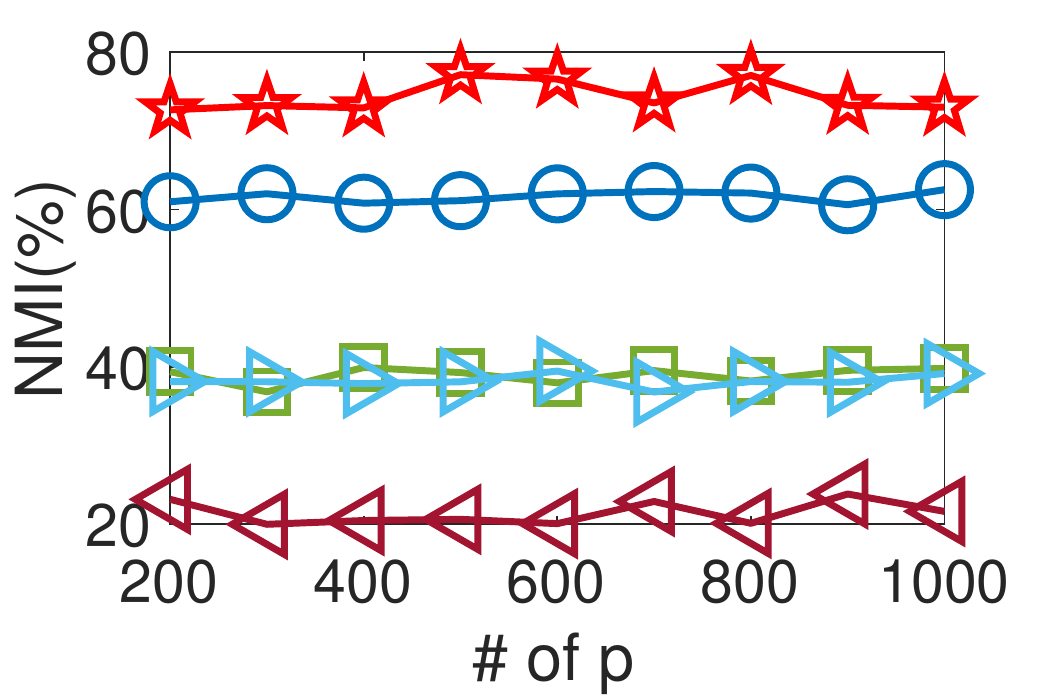}
      &\includegraphics[width=0.18\textwidth]{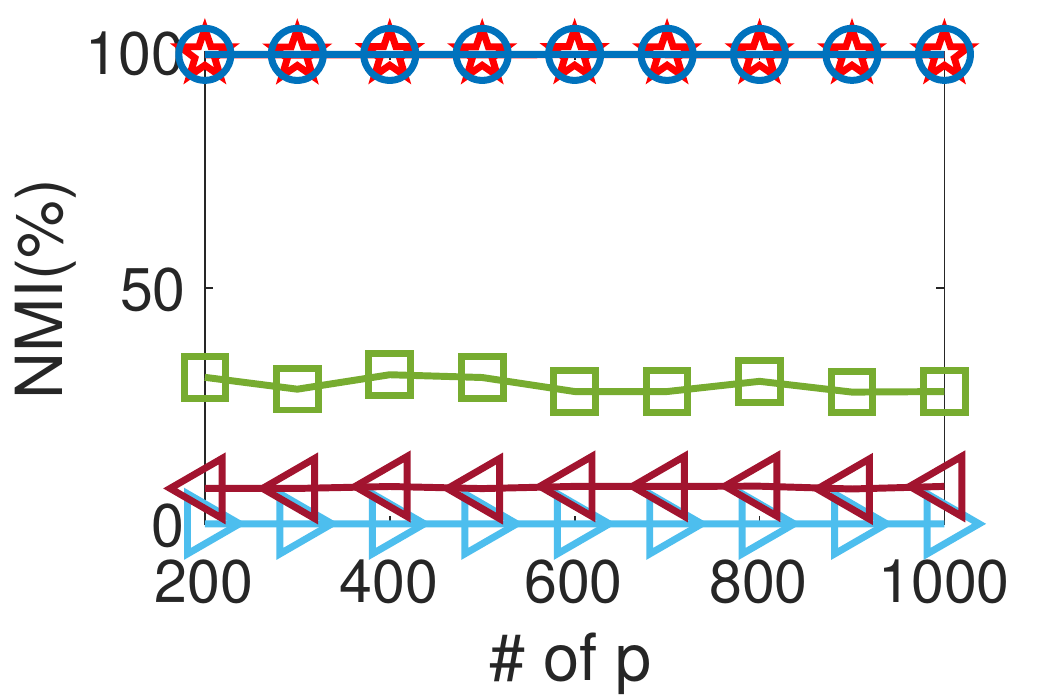}\\
      Time cost
      &\includegraphics[width=0.18\textwidth]{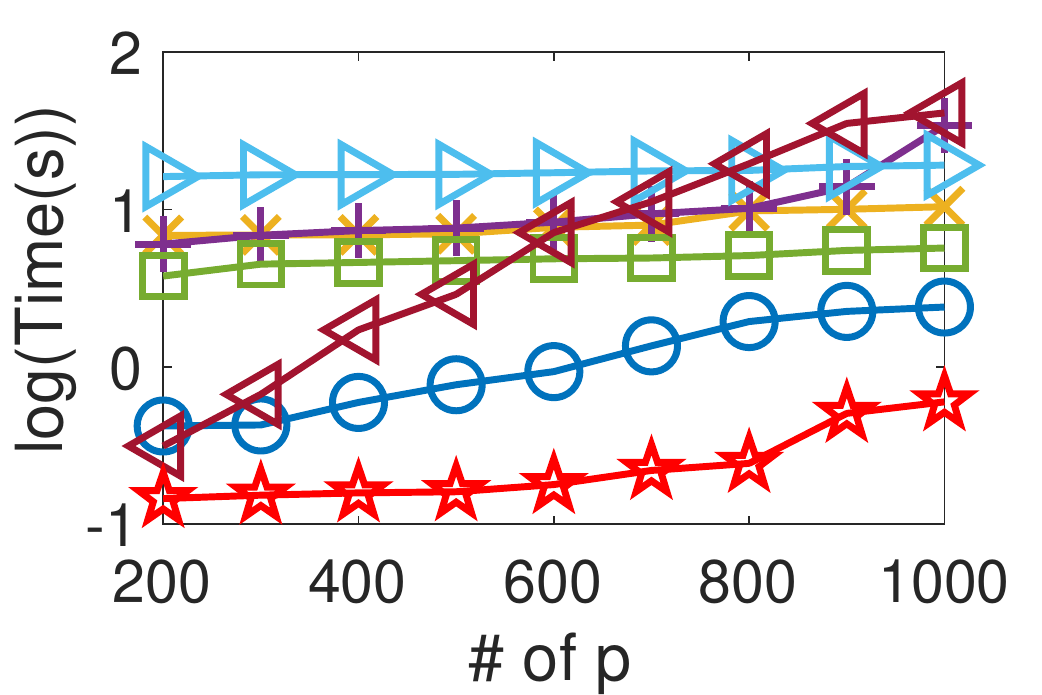}
      &\includegraphics[width=0.18\textwidth]{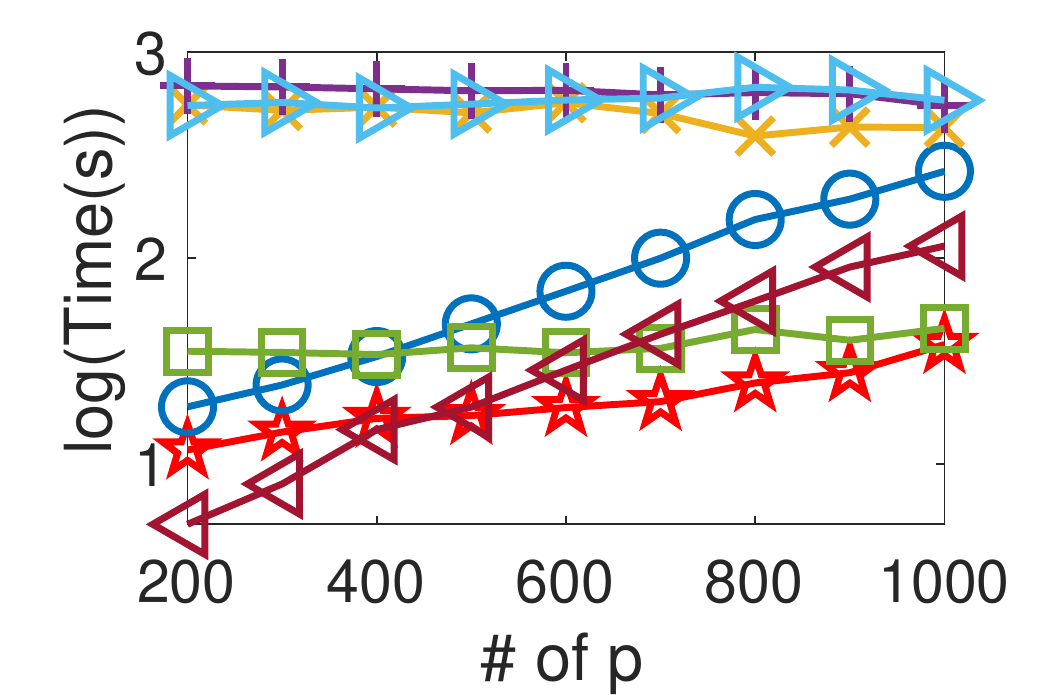}
      &\includegraphics[width=0.18\textwidth]{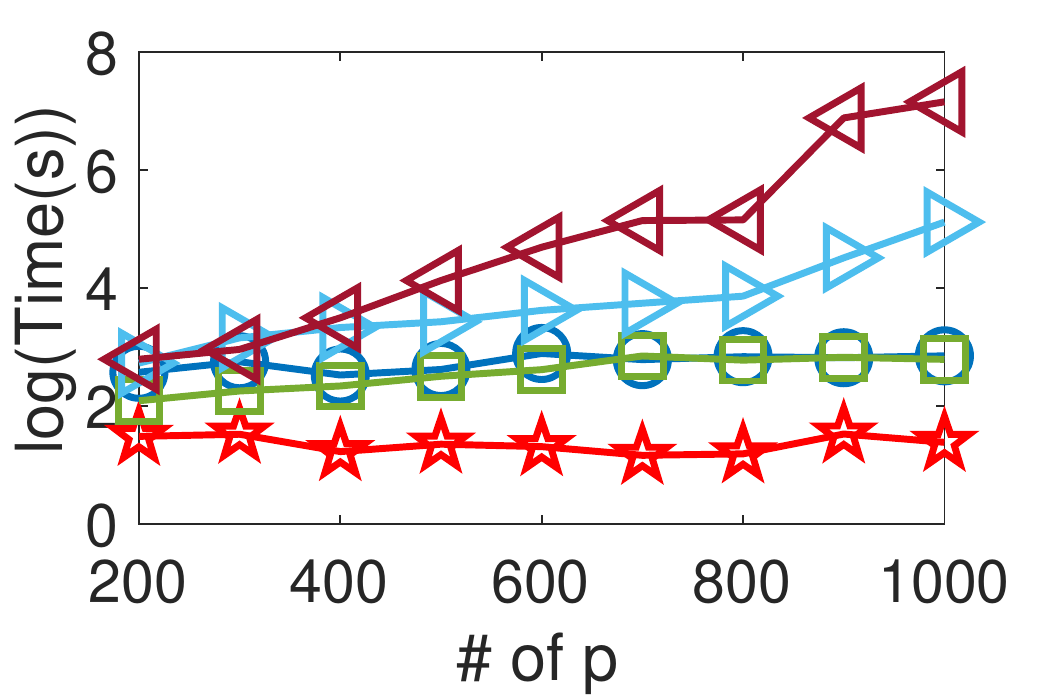}
      &\includegraphics[width=0.18\textwidth]{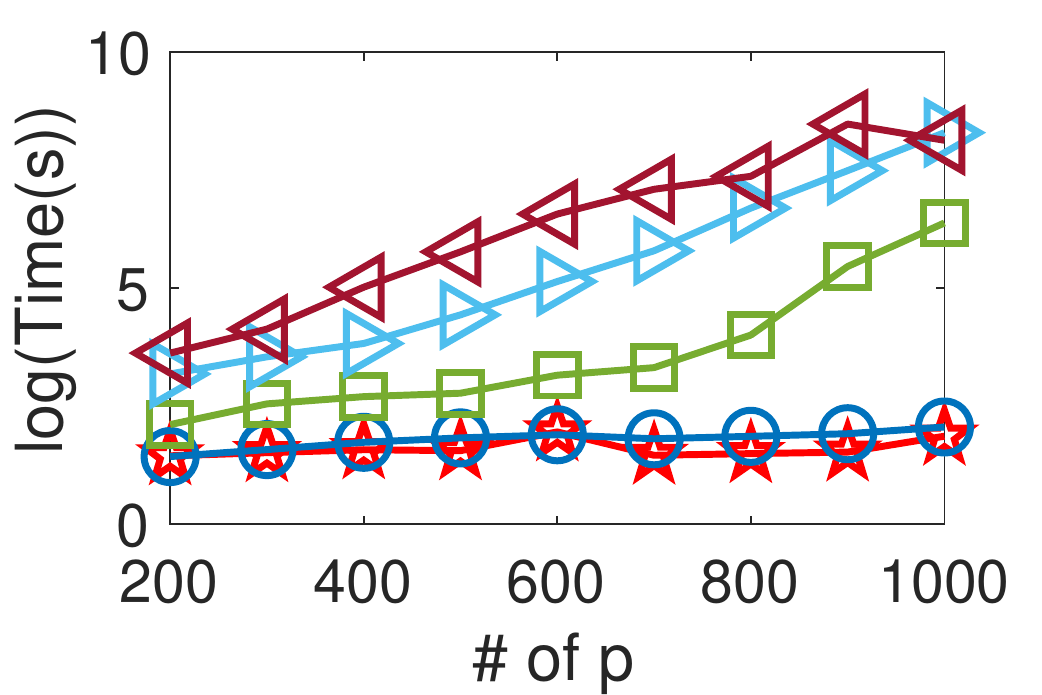}\\
      &\multicolumn{4}{c}{\includegraphics[width=0.8\textwidth]{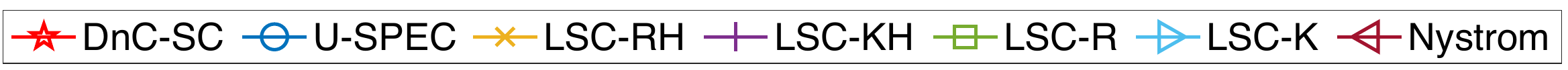}}\\
      \bottomrule
    \end{tabular}
    \begin{tablenotes}
      \item[*] LSC-KH and LSC-RH cannot be conduct on the \emph{TM-60K} and \emph{TM-1M} dataset due to the memory bottleneck.
    \end{tablenotes}
  \end{threeparttable}
\end{table*}

\begin{table*}%[!t]
  \centering
  \caption{Clustering performance (ACC(\%), NMI(\%), and time costs(s)) for different methods by varying number of nearest landmarks $K$.}
  \label{table:compare_para_Knn}
  \begin{threeparttable}
    \begin{tabular}{m{0.08\textwidth}<{\centering}|m{0.2\textwidth}<{\centering}m{0.2\textwidth}<{\centering}m{0.2\textwidth}<{\centering}m{0.2\textwidth}<{\centering}}
      \toprule
      \emph{Dataset} & \emph{Letters} & \emph{MNIST} & \emph{TS-60K} & \emph{TM-1M} \\
      \midrule
      \multirow{1}{*}{ACC}
      &\includegraphics[width=0.18\textwidth]{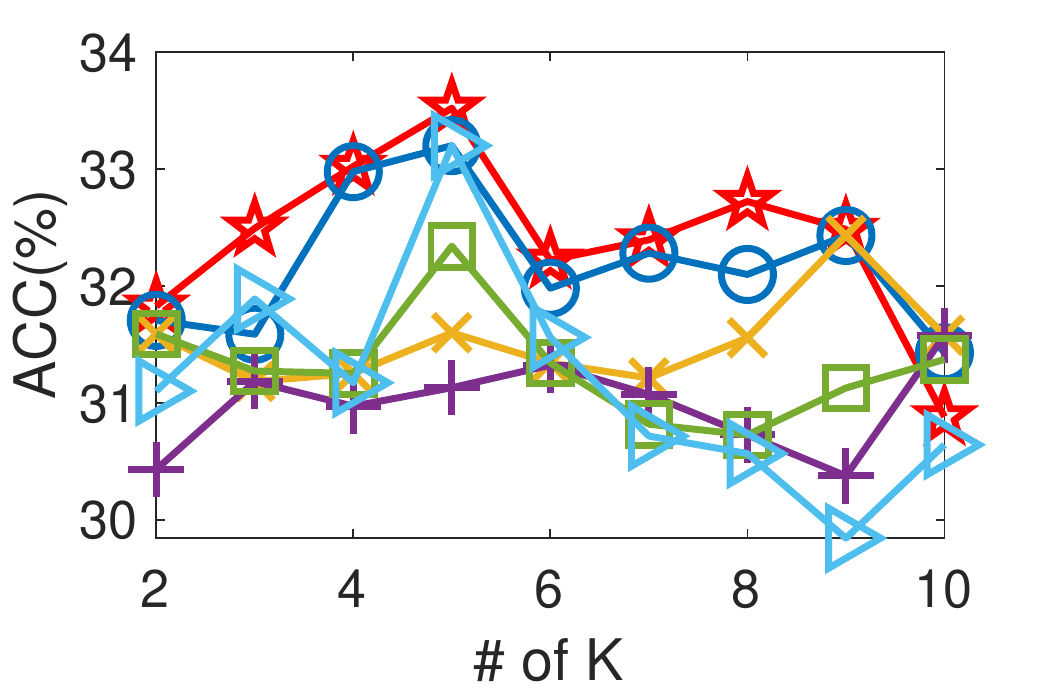}
      &\includegraphics[width=0.18\textwidth]{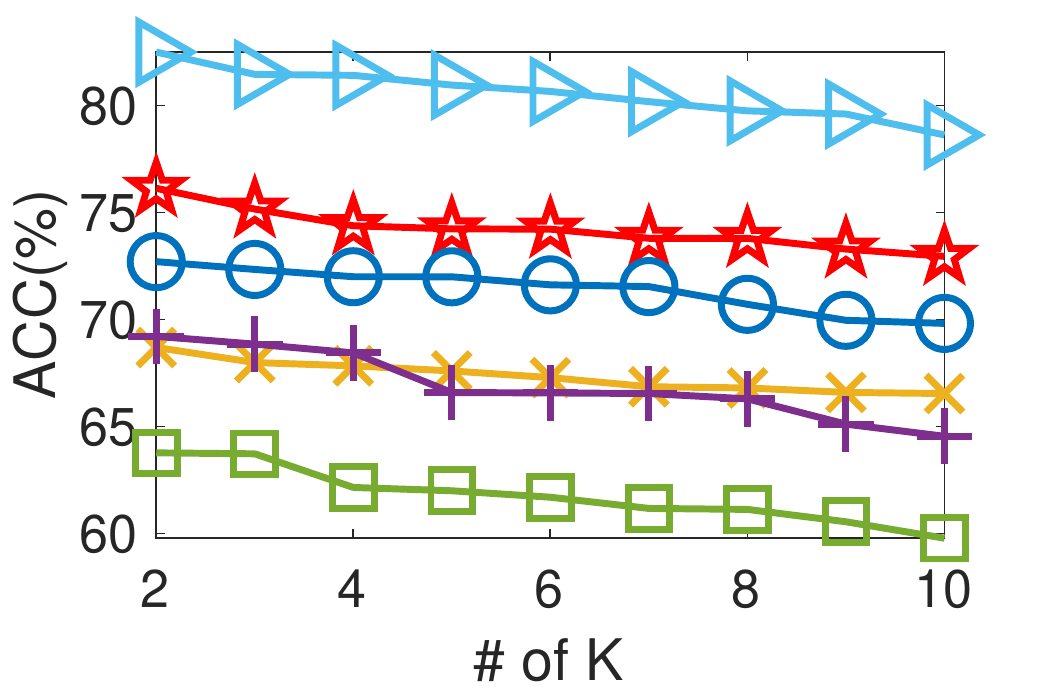}
      &\includegraphics[width=0.18\textwidth]{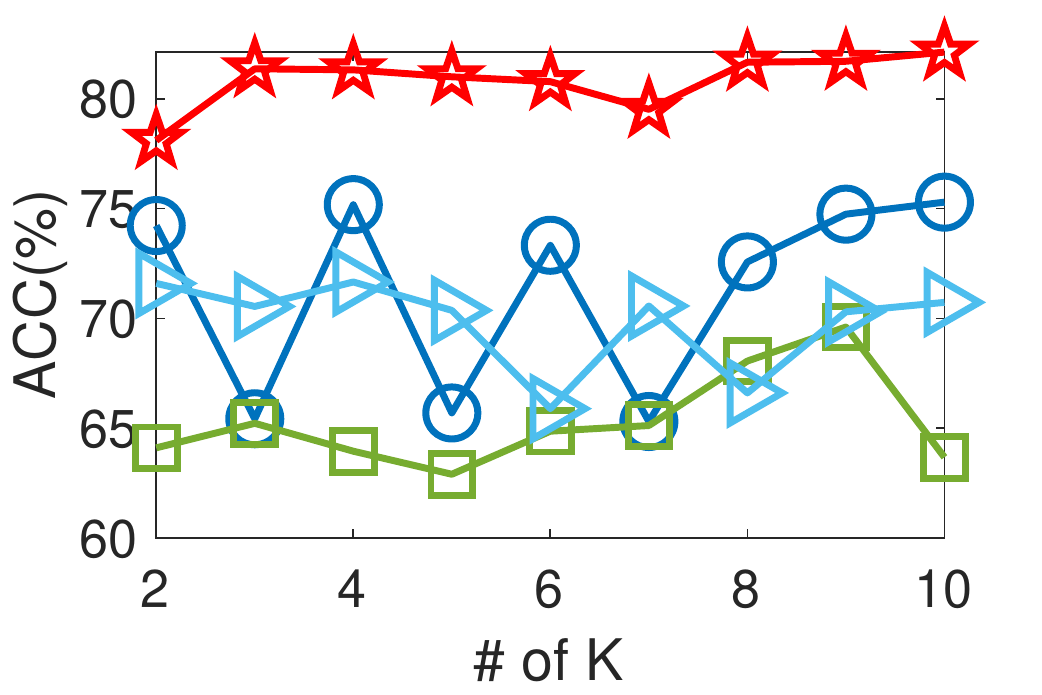}
      &\includegraphics[width=0.18\textwidth]{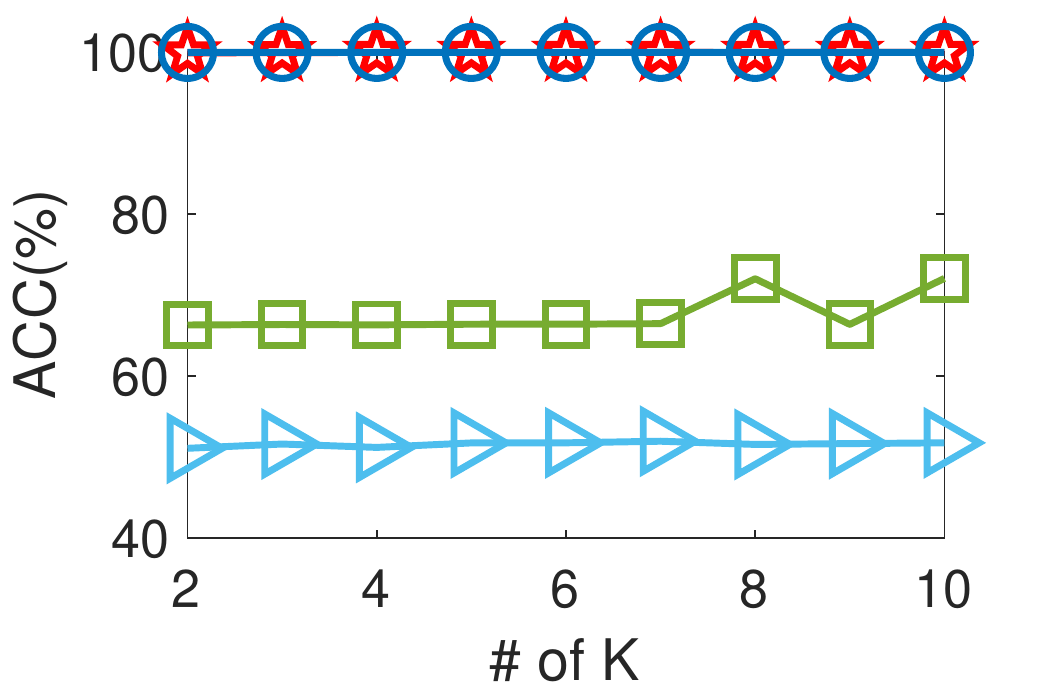}\\
      NMI
      &\includegraphics[width=0.18\textwidth]{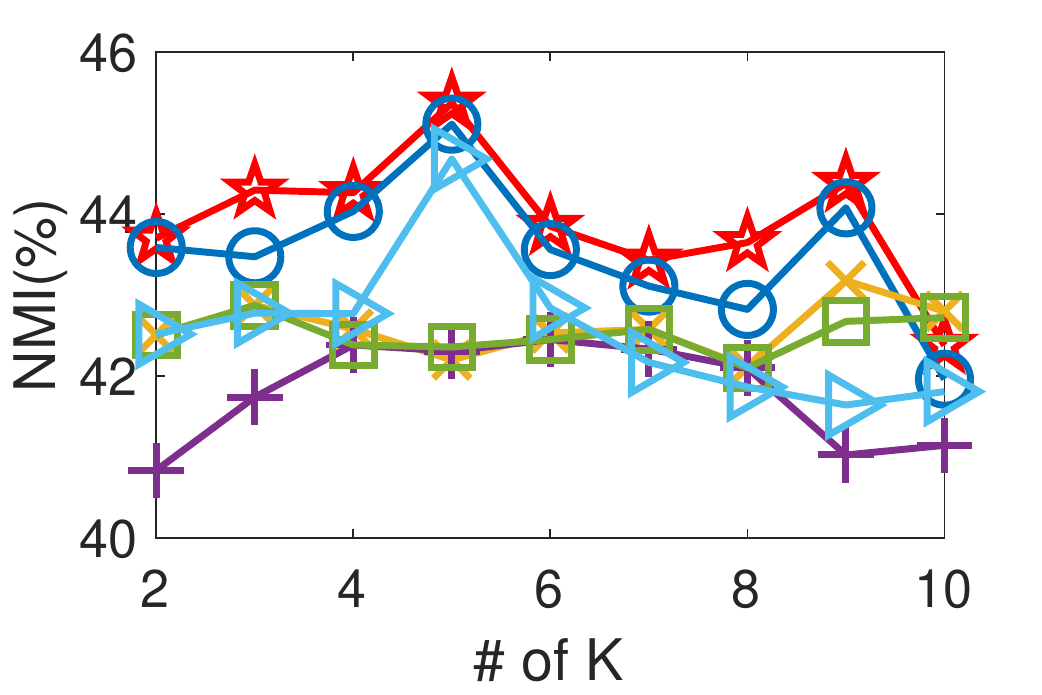}
      &\includegraphics[width=0.18\textwidth]{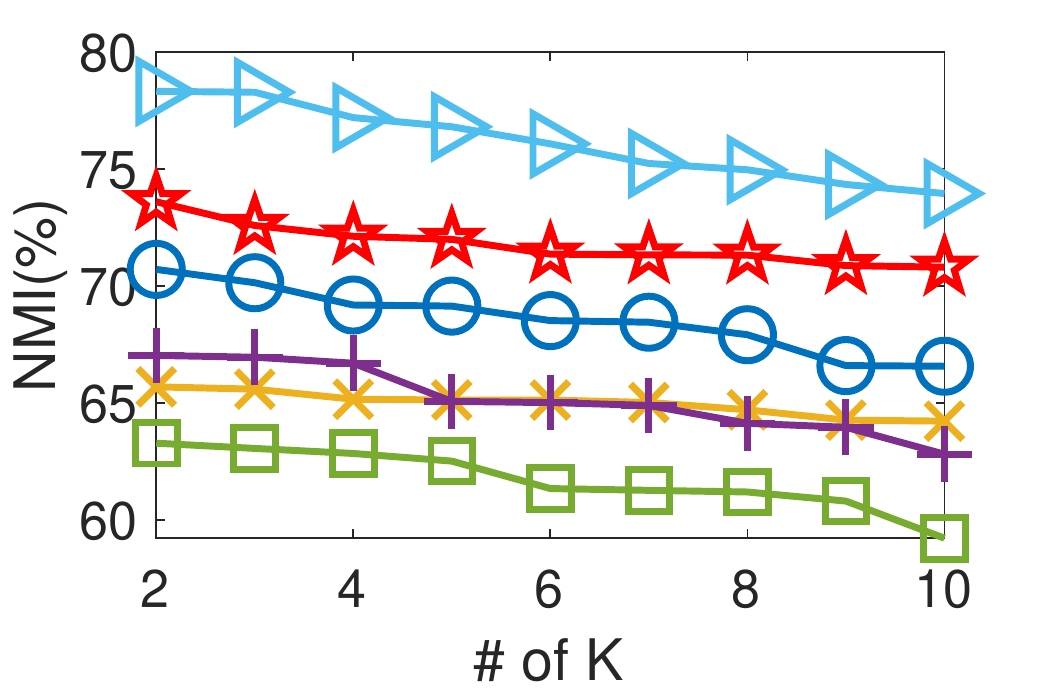}
      &\includegraphics[width=0.18\textwidth]{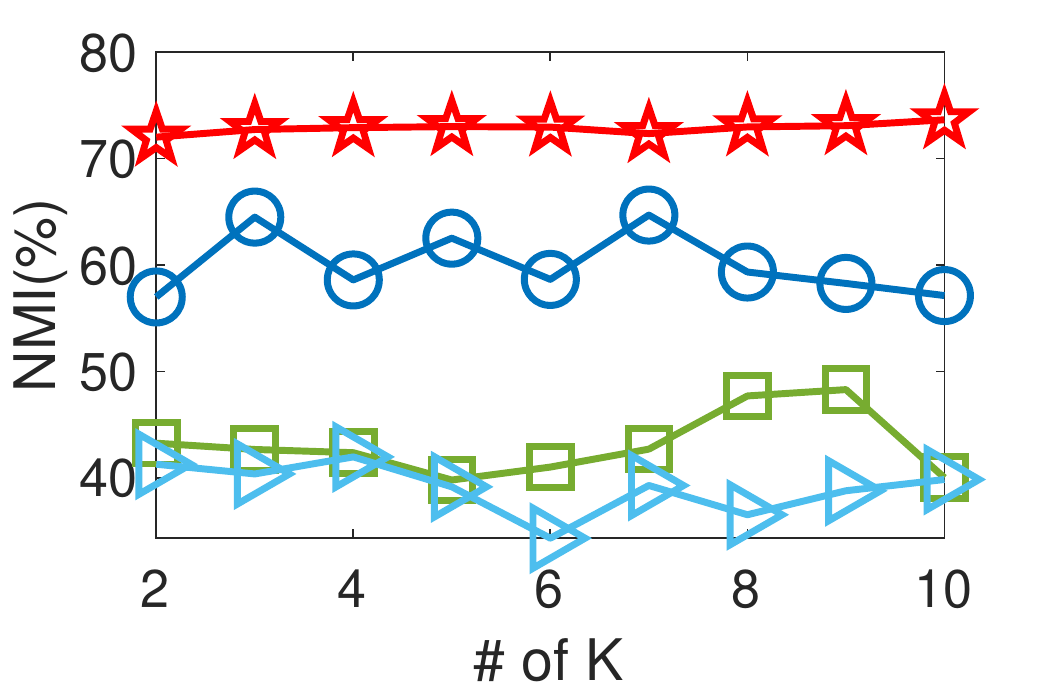}
      &\includegraphics[width=0.18\textwidth]{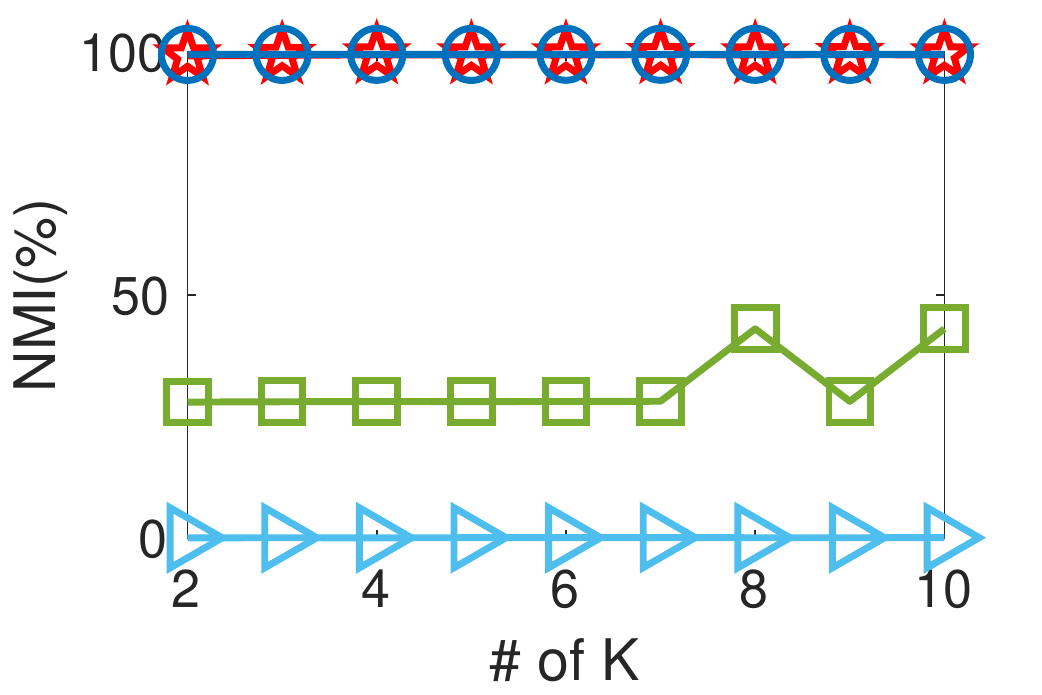}\\
      Time cost
      &\includegraphics[width=0.18\textwidth]{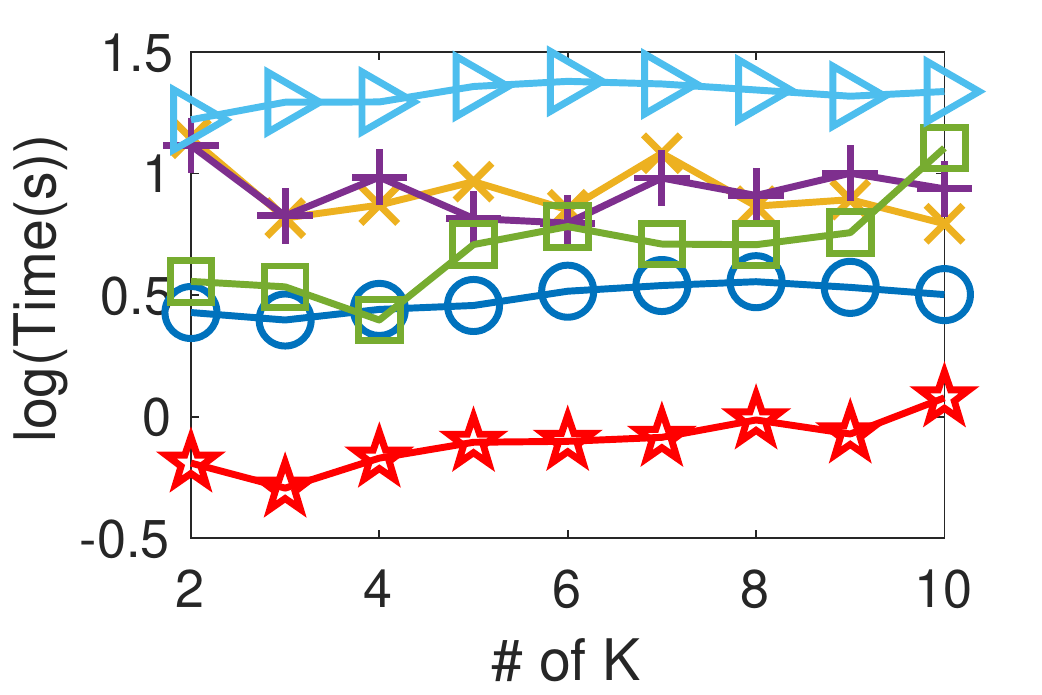}
      &\includegraphics[width=0.18\textwidth]{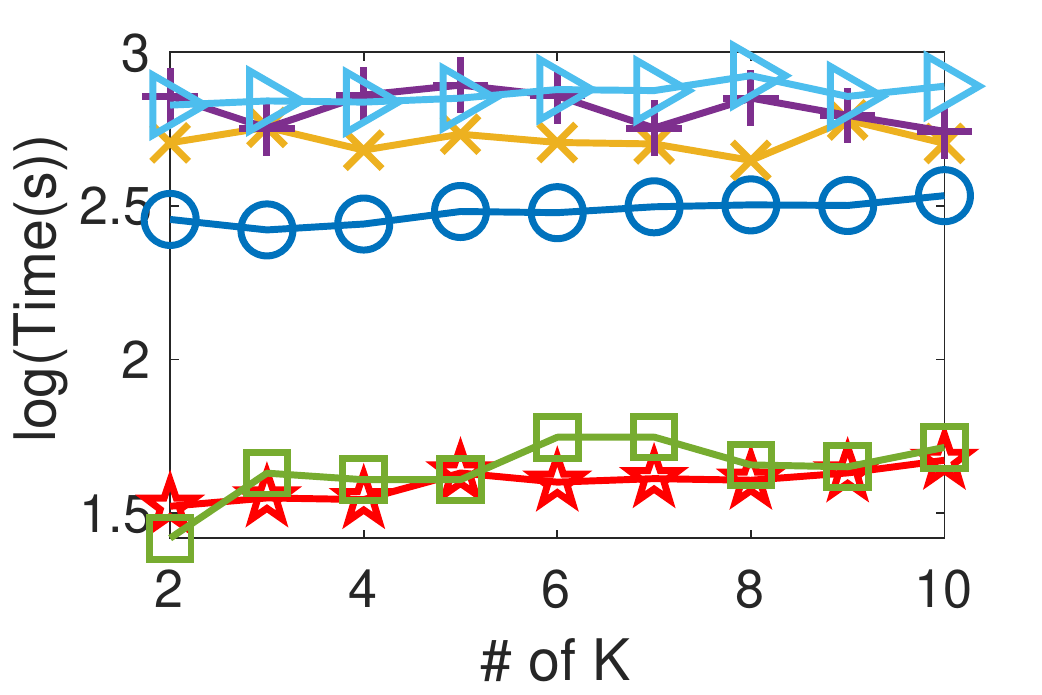}
      &\includegraphics[width=0.18\textwidth]{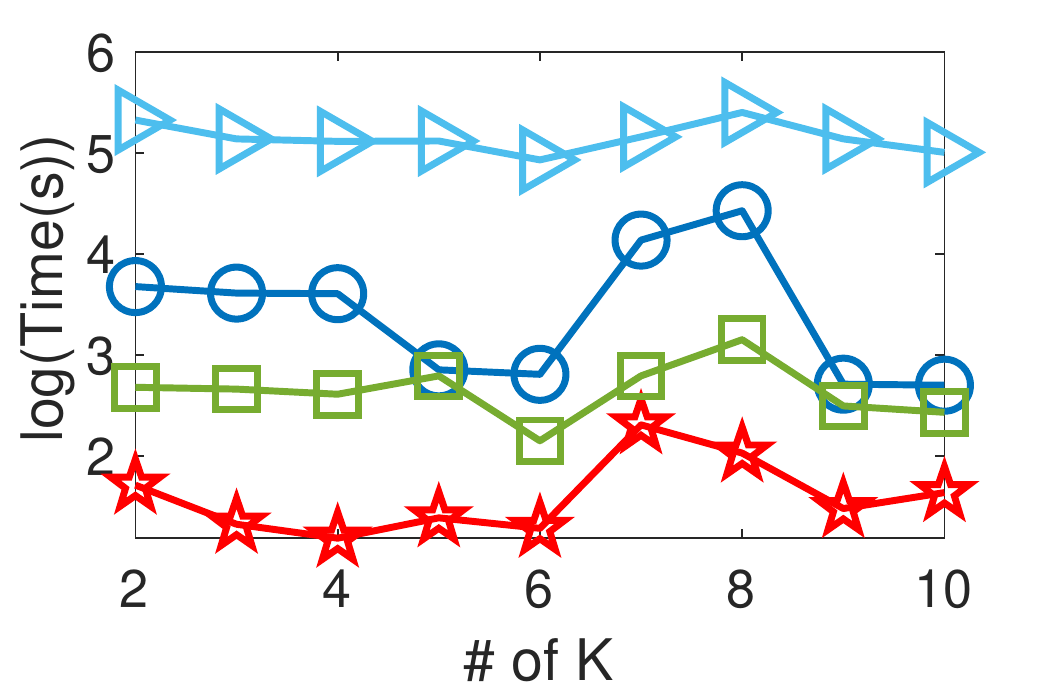}
      &\includegraphics[width=0.18\textwidth]{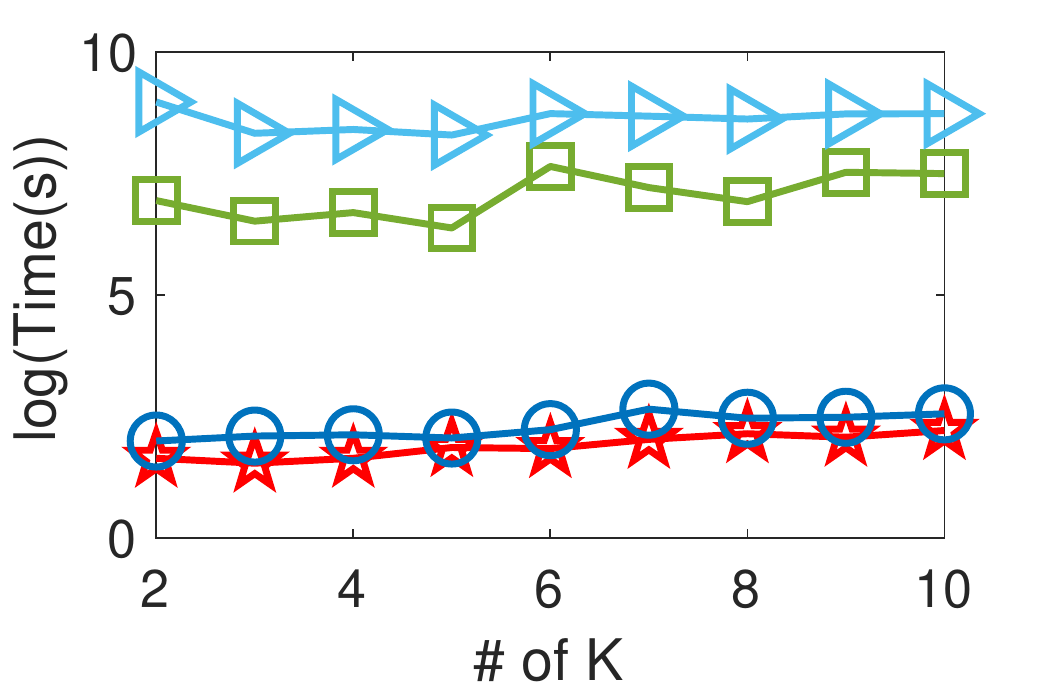}\\
      &\multicolumn{4}{c}{\includegraphics[width=0.6\textwidth]{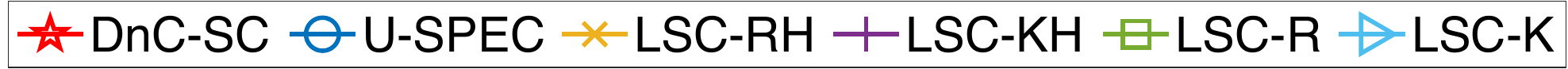}}\\
      \bottomrule
    \end{tabular}
  \end{threeparttable}
\end{table*}

\begin{table*}%[!t]
  \centering
  \caption{Clustering performance (ACC(\%), NMI(\%), and time costs(s)) for different methods by varying number of nearest landmark $K$ and selection rate $\alpha$.}
  \label{table:compare_para_K_u}
  \begin{threeparttable}
    \begin{tabular}{m{0.08\textwidth}<{\centering}|m{0.2\textwidth}<{\centering}m{0.2\textwidth}<{\centering}m{0.2\textwidth}<{\centering}m{0.2\textwidth}<{\centering}}
      \toprule
      \emph{Dataset} & \emph{Letters} & \emph{MNIST} & \emph{TS-60K} & \emph{TM-1M} \\
      \midrule
      \multirow{1}{*}{ACC}
      &\includegraphics[width=0.2\textwidth]{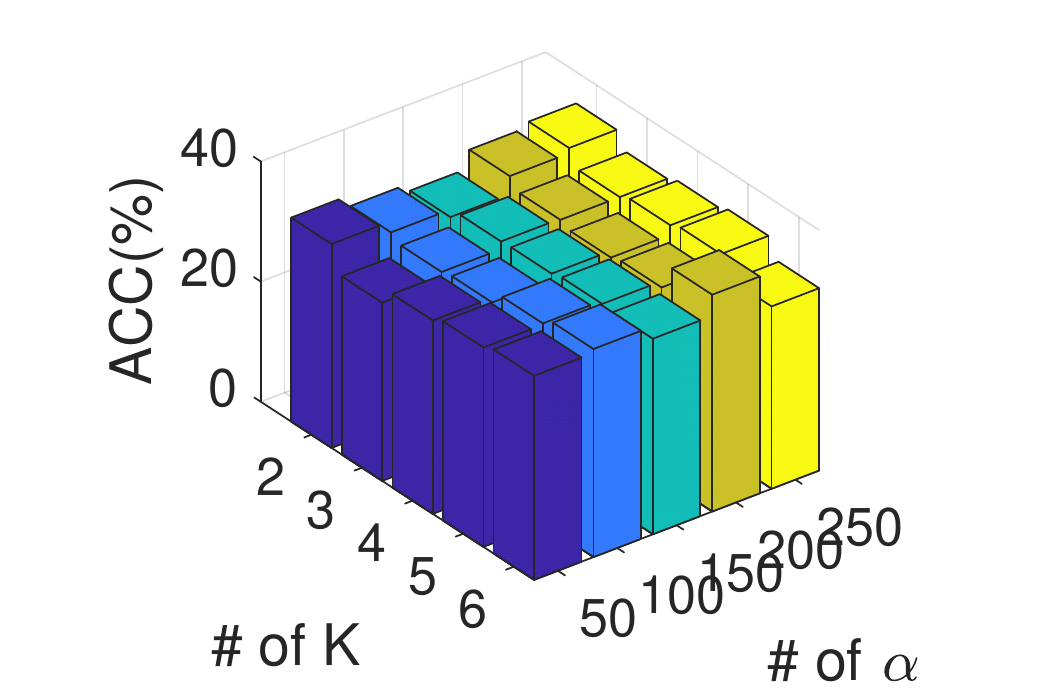}
      &\includegraphics[width=0.2\textwidth]{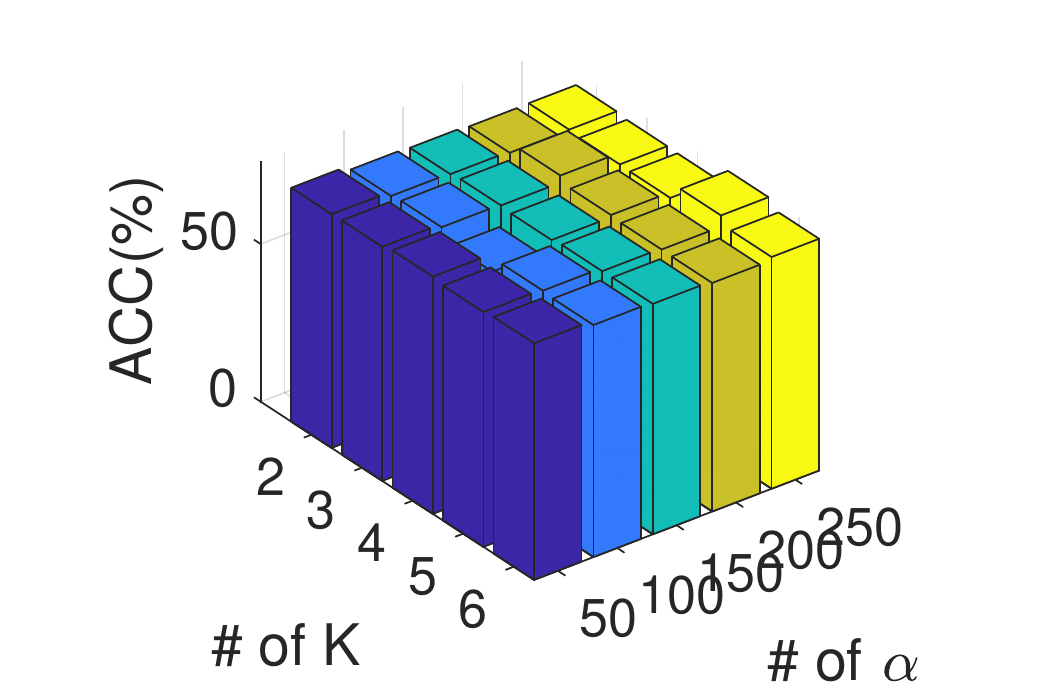}
      &\includegraphics[width=0.2\textwidth]{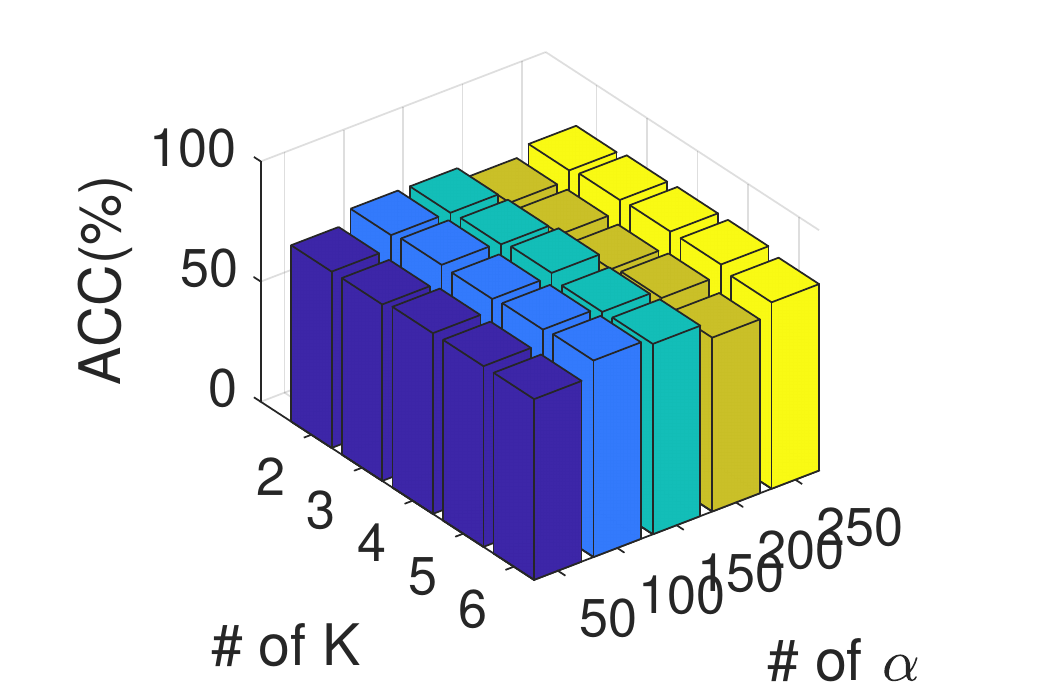}
      &\includegraphics[width=0.2\textwidth]{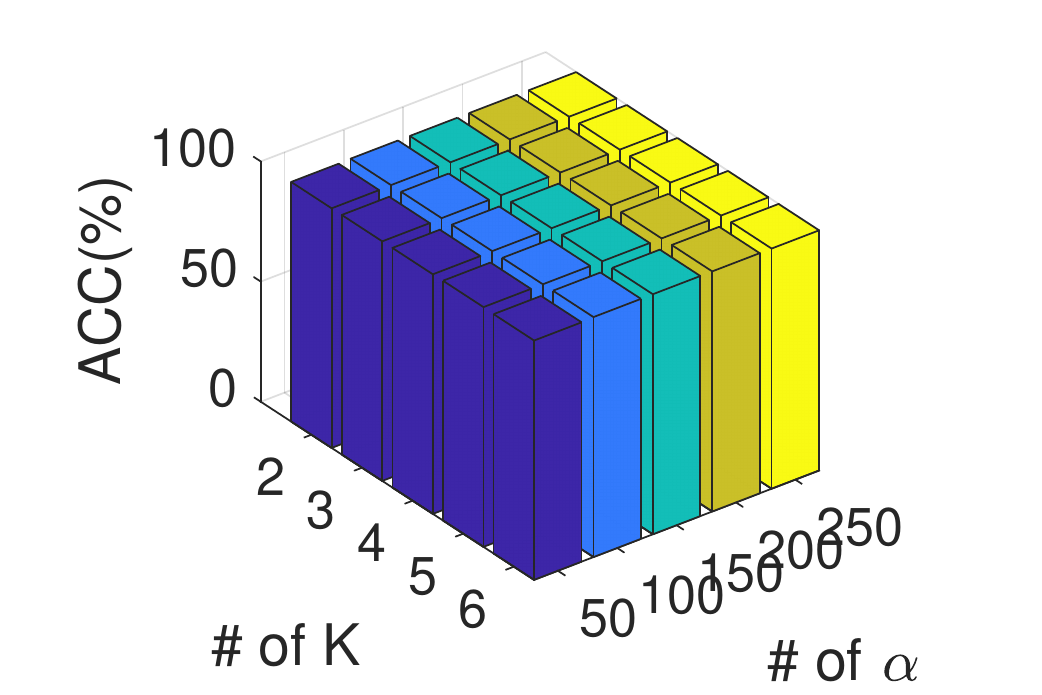}\\
      NMI
      &\includegraphics[width=0.2\textwidth]{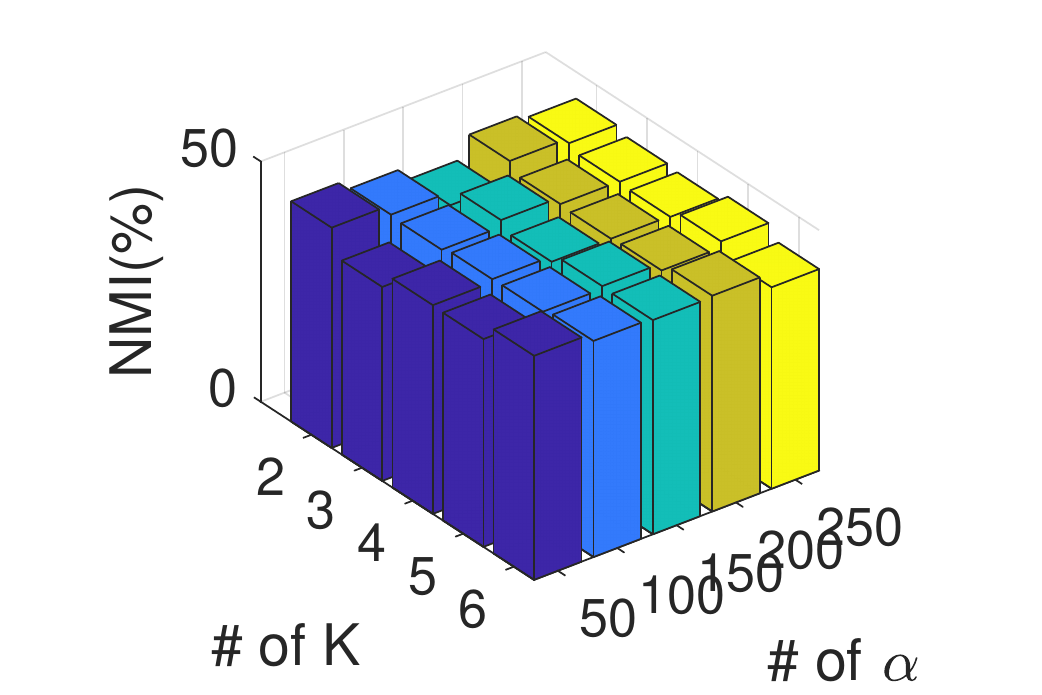}
      &\includegraphics[width=0.2\textwidth]{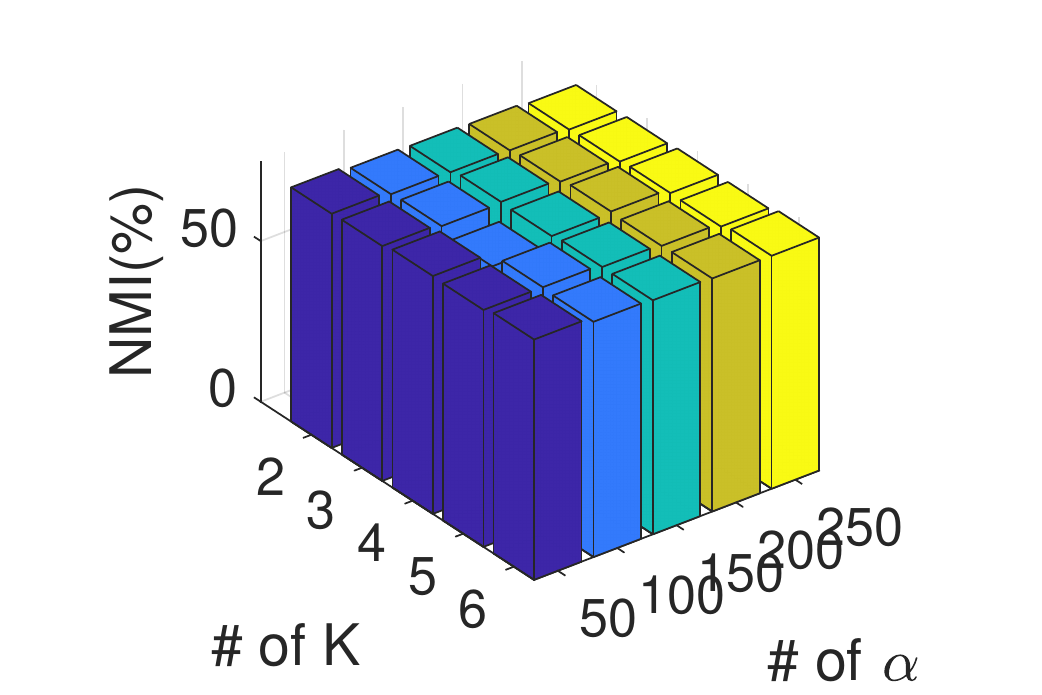}
      &\includegraphics[width=0.2\textwidth]{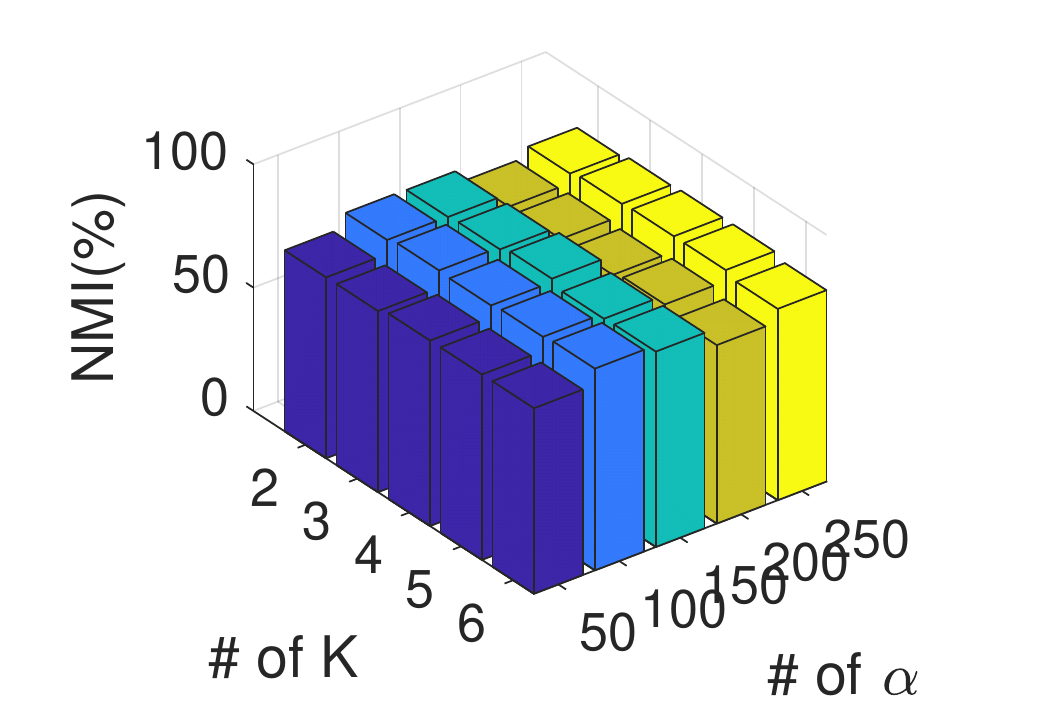}
      &\includegraphics[width=0.2\textwidth]{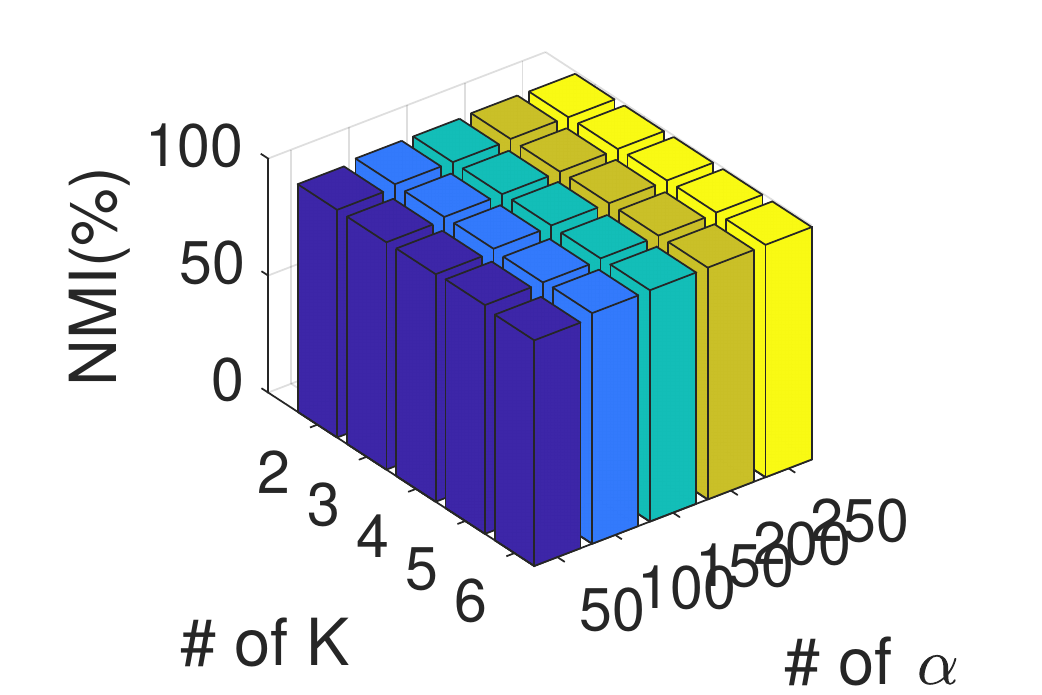}\\
      Time cost
      &\includegraphics[width=0.2\textwidth]{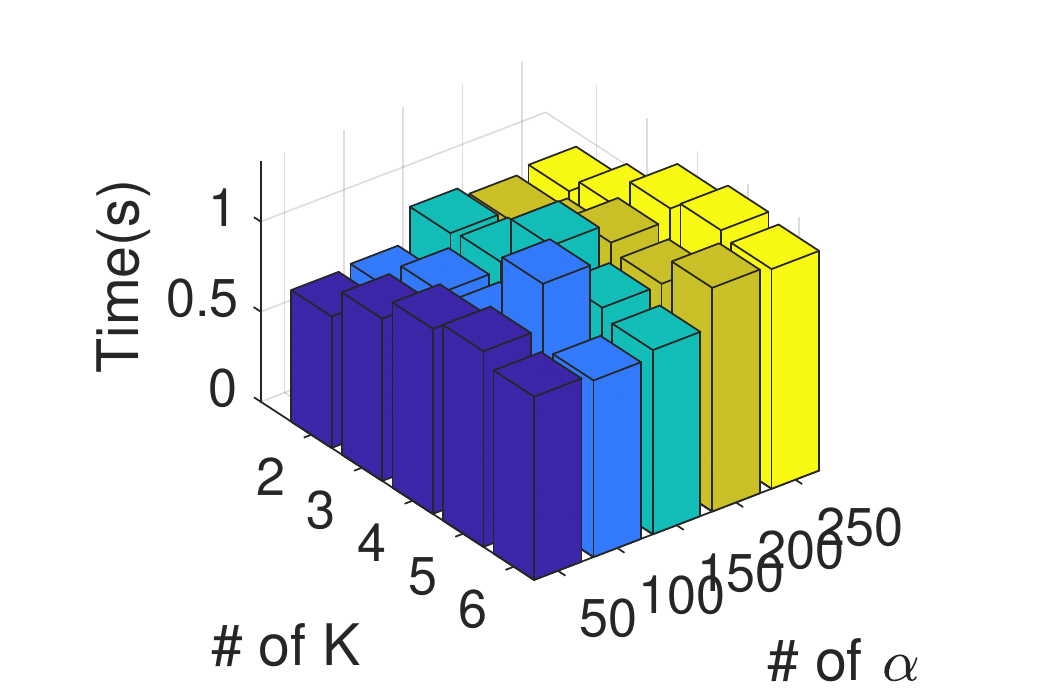}
      &\includegraphics[width=0.2\textwidth]{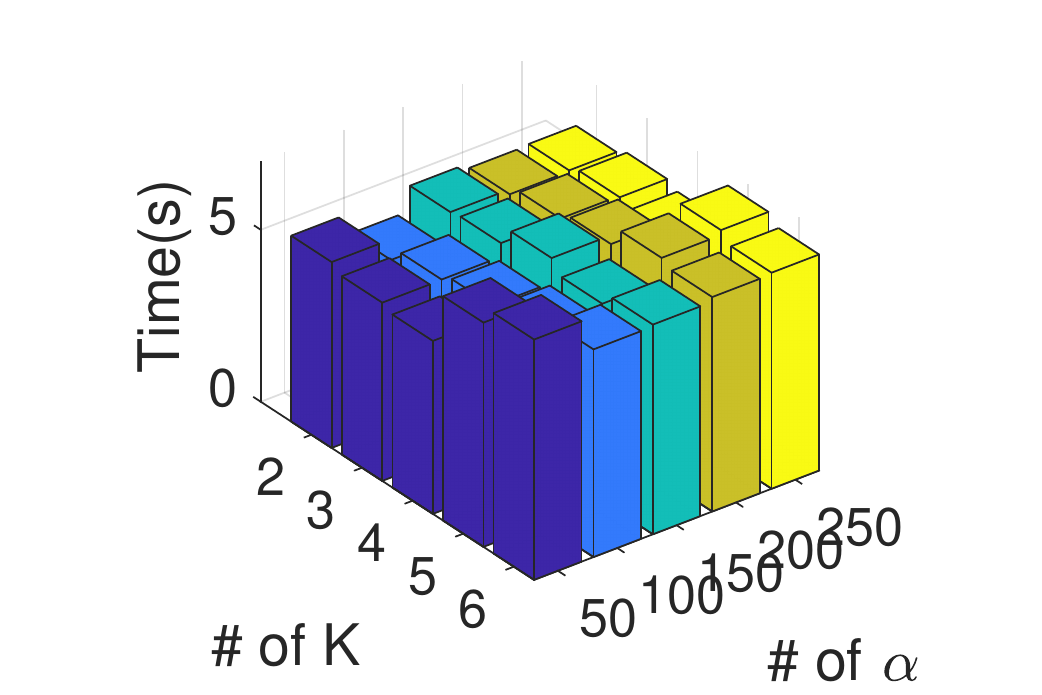}
      &\includegraphics[width=0.2\textwidth]{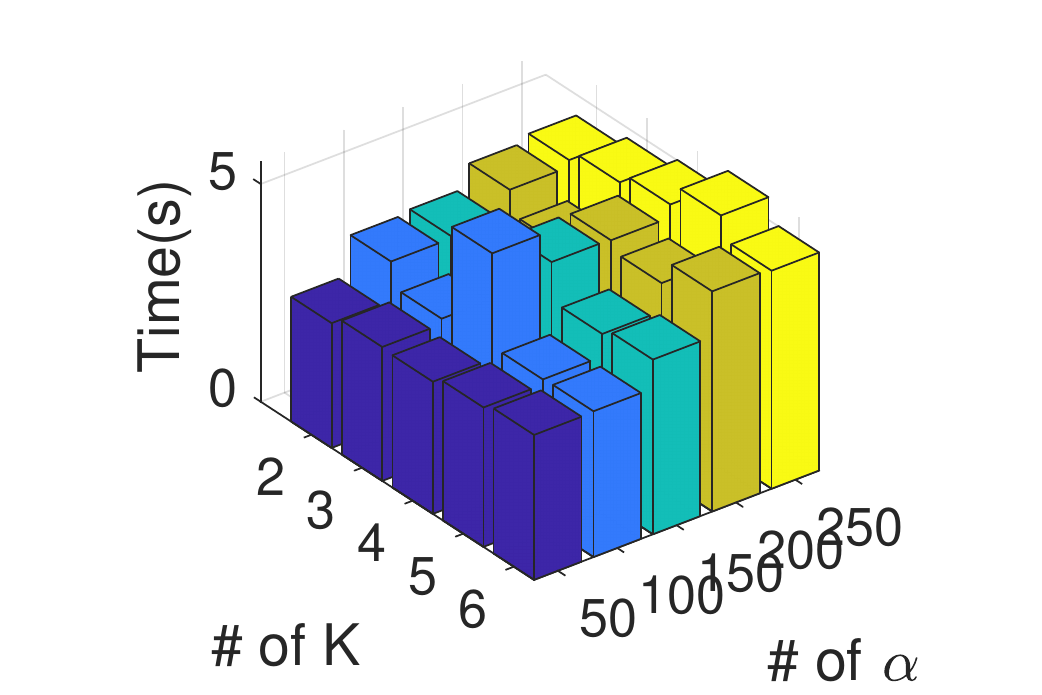}
      &\includegraphics[width=0.2\textwidth]{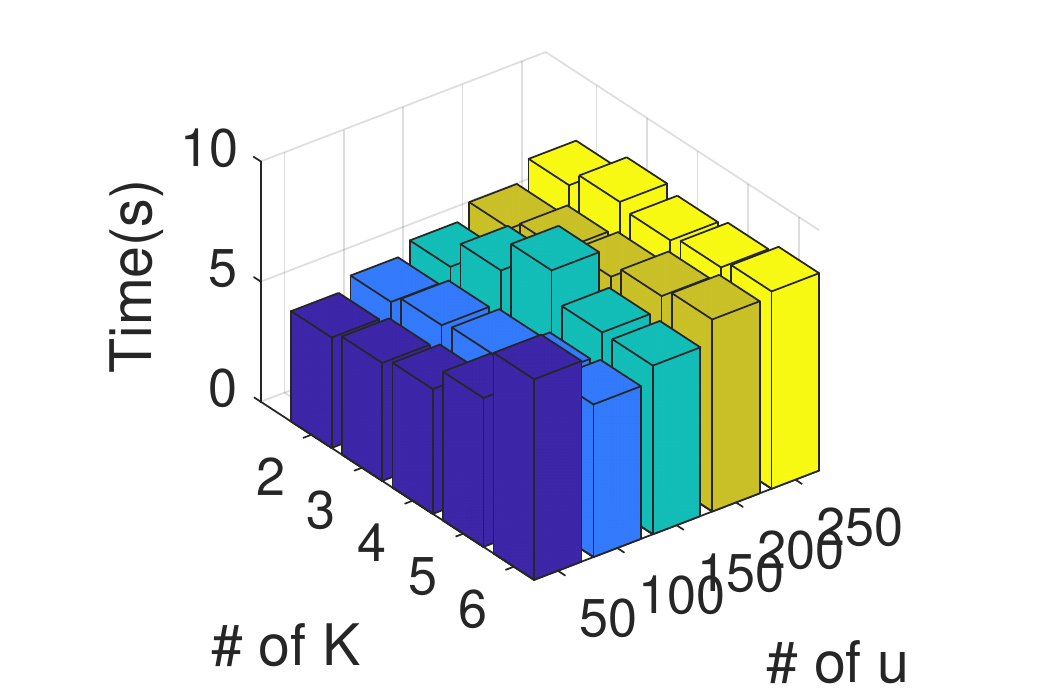}\\
      \bottomrule
    \end{tabular}
  \end{threeparttable}
\end{table*}

\begin{table}%[!t]
  \centering
  \caption{Clustering performance (ACC(\%), NMI(\%), and time costs(s)) for DnC-SC using divide-and-conquer based landmark selection and $k$-means based landmark selection.}
  \label{table:compare_sel_strategies}
  \begin{threeparttable}
    \begin{tabular}{m{0.75cm}<{\centering}|m{1.45cm}<{\centering}m{1.45cm}<{\centering}m{1.45cm}<{\centering}m{1.45cm}<{\centering}}
      \toprule
      \emph{Data} & \emph{Letters} & \emph{MNIST} & \emph{TS-60K} & \emph{TM-1M} \\
      \midrule
      \multirow{1}{*}{ACC}
      &\includegraphics[width=1.7cm]{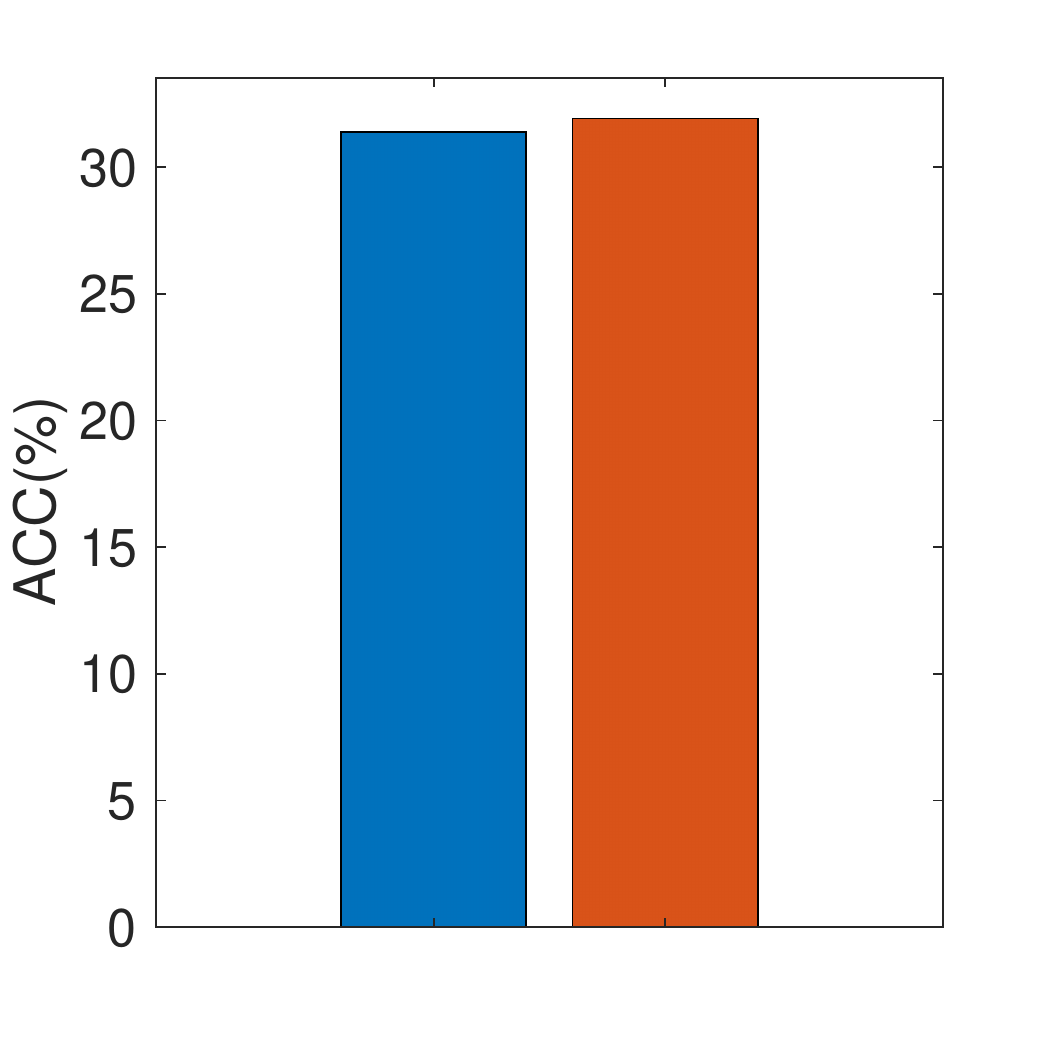}
      &\includegraphics[width=1.7cm]{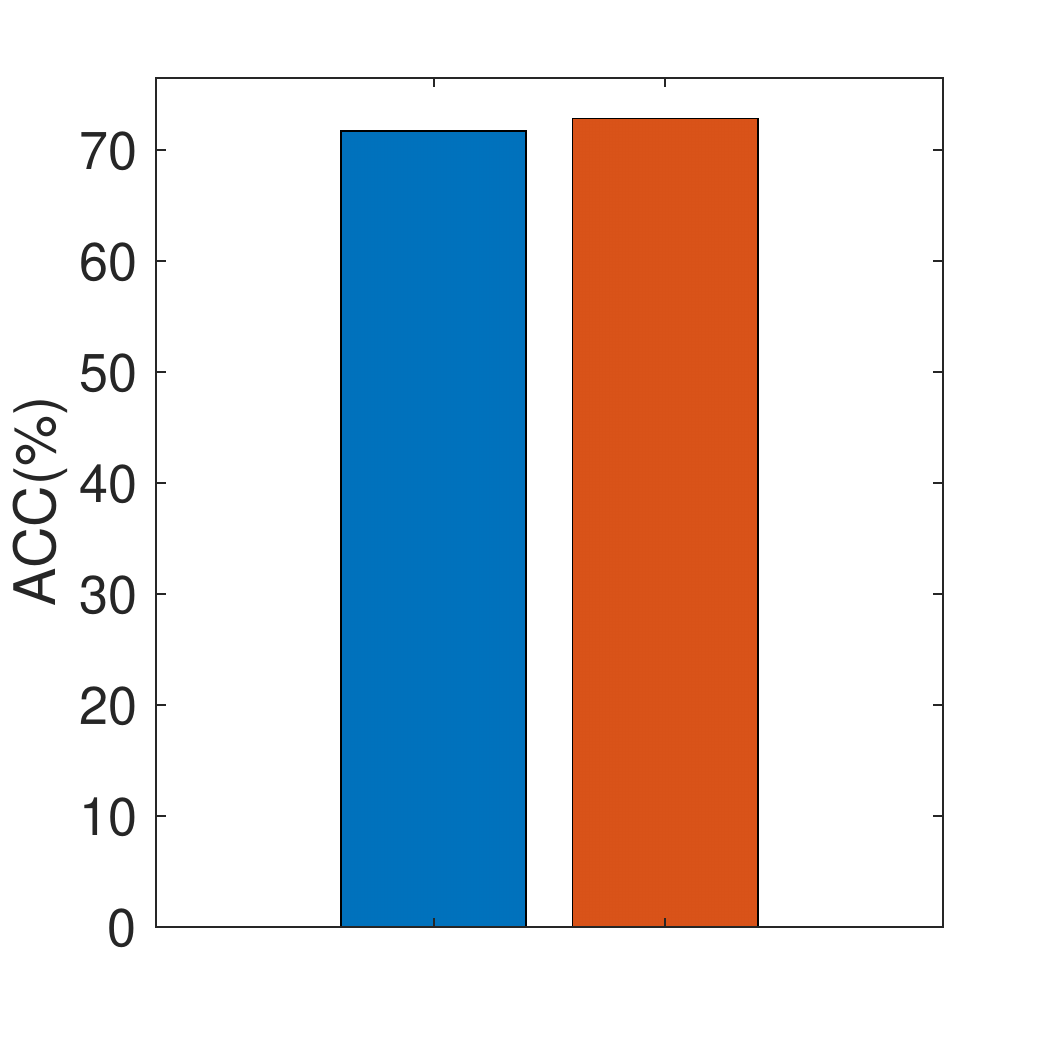}
      &\includegraphics[width=1.7cm]{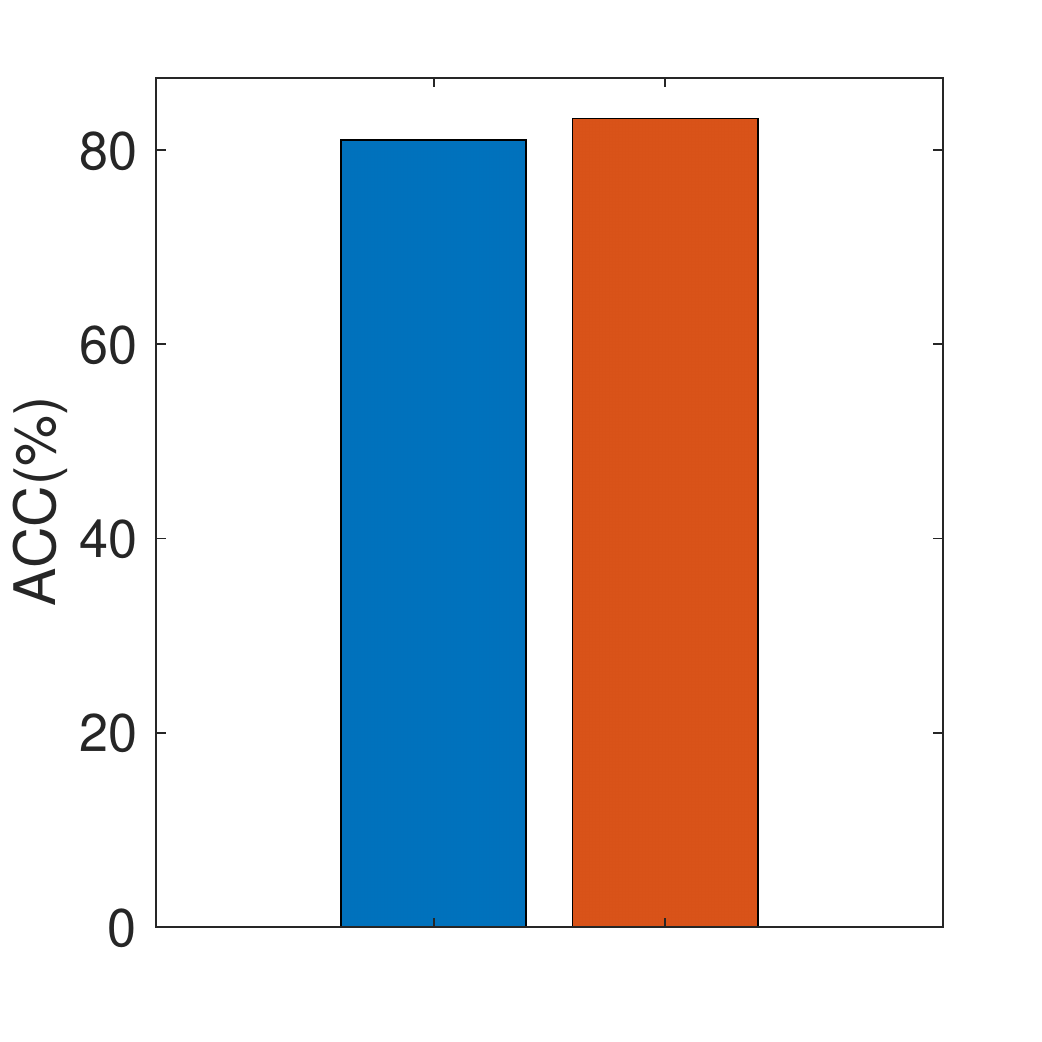}
      &\includegraphics[width=1.7cm]{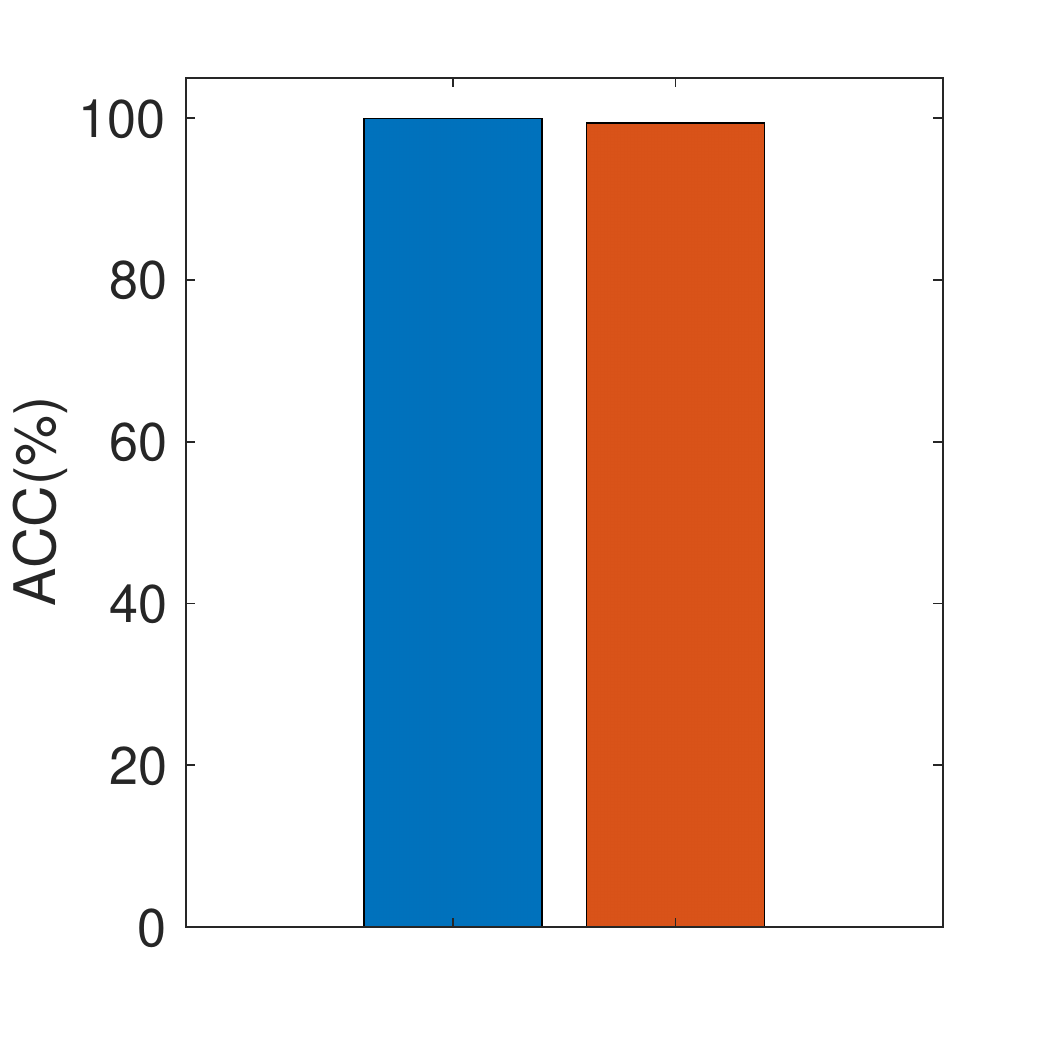}\\
      NMI
      &\includegraphics[width=1.7cm]{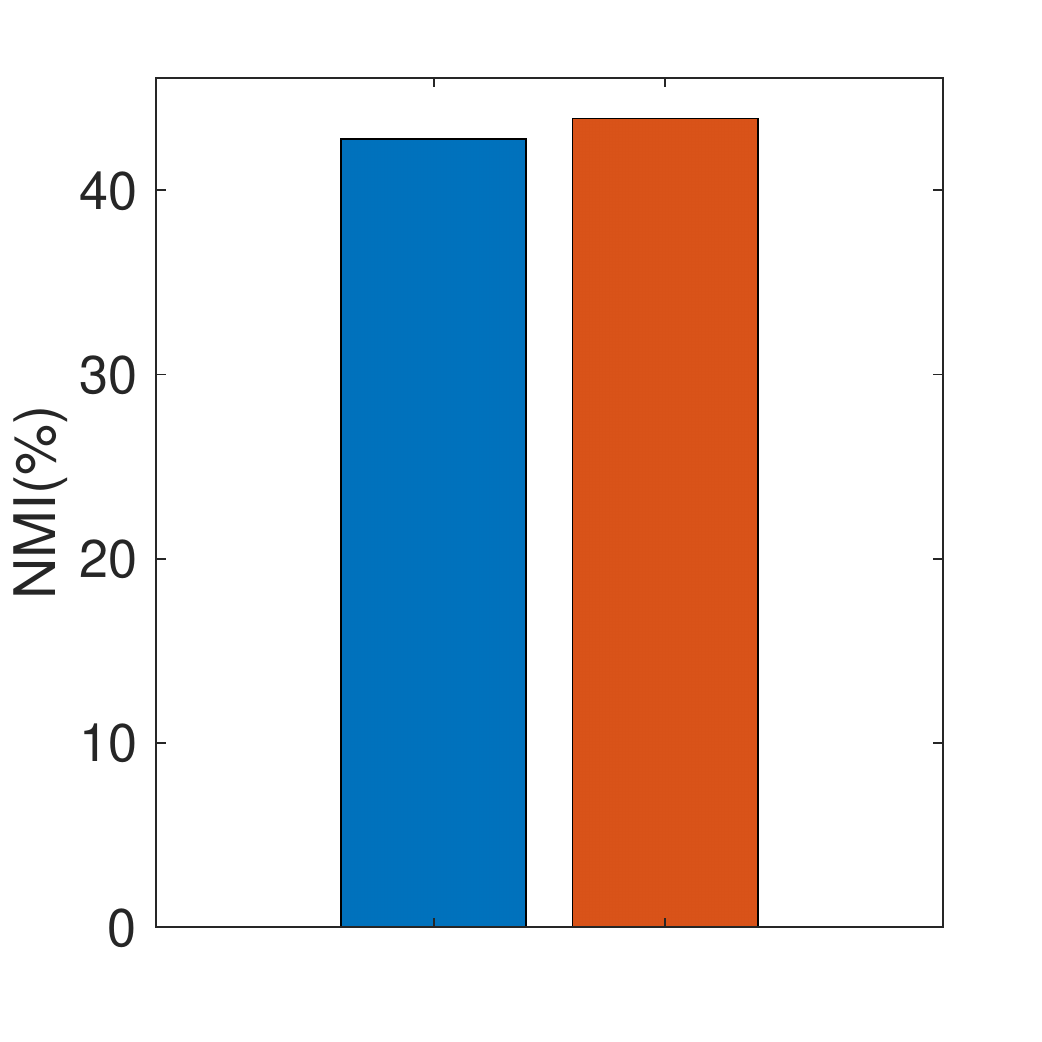}
      &\includegraphics[width=1.7cm]{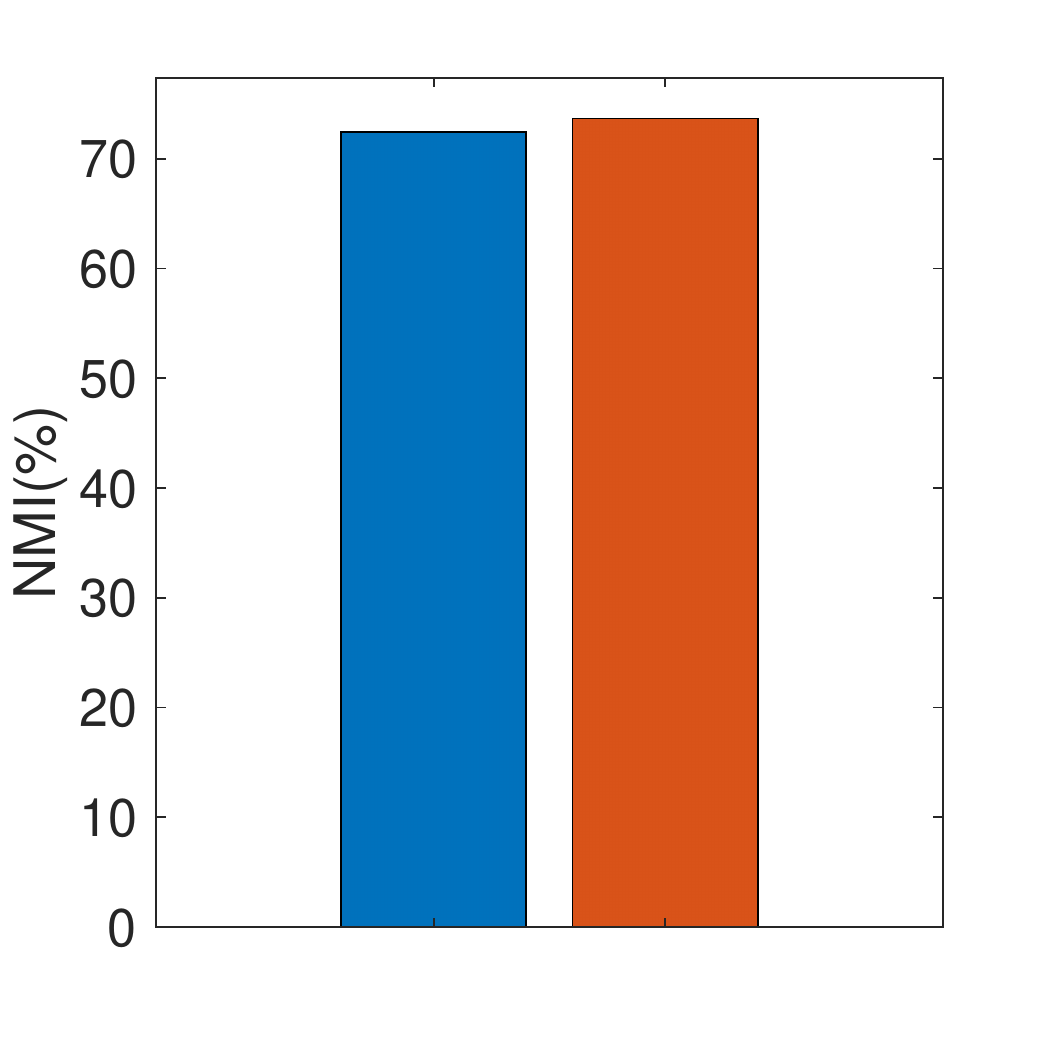}
      &\includegraphics[width=1.7cm]{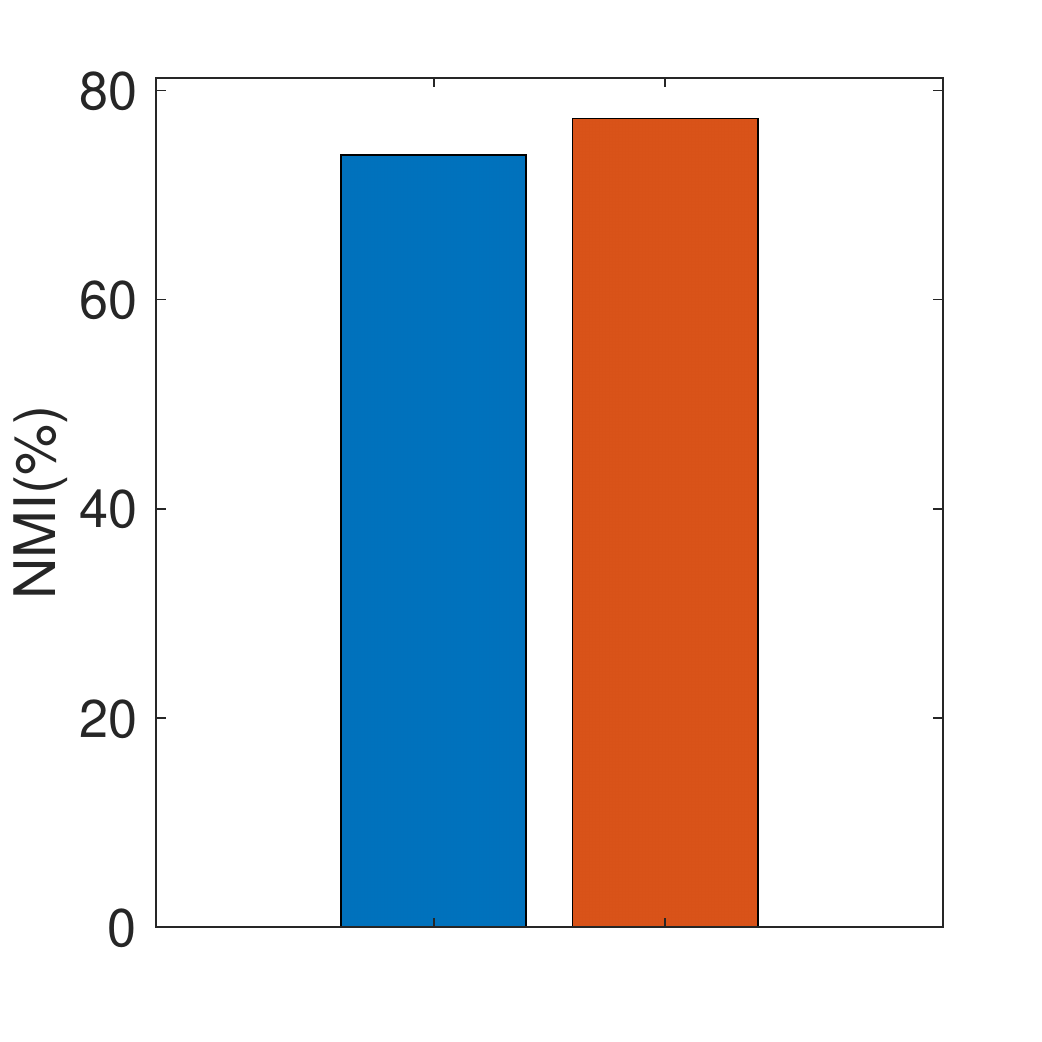}
      &\includegraphics[width=1.7cm]{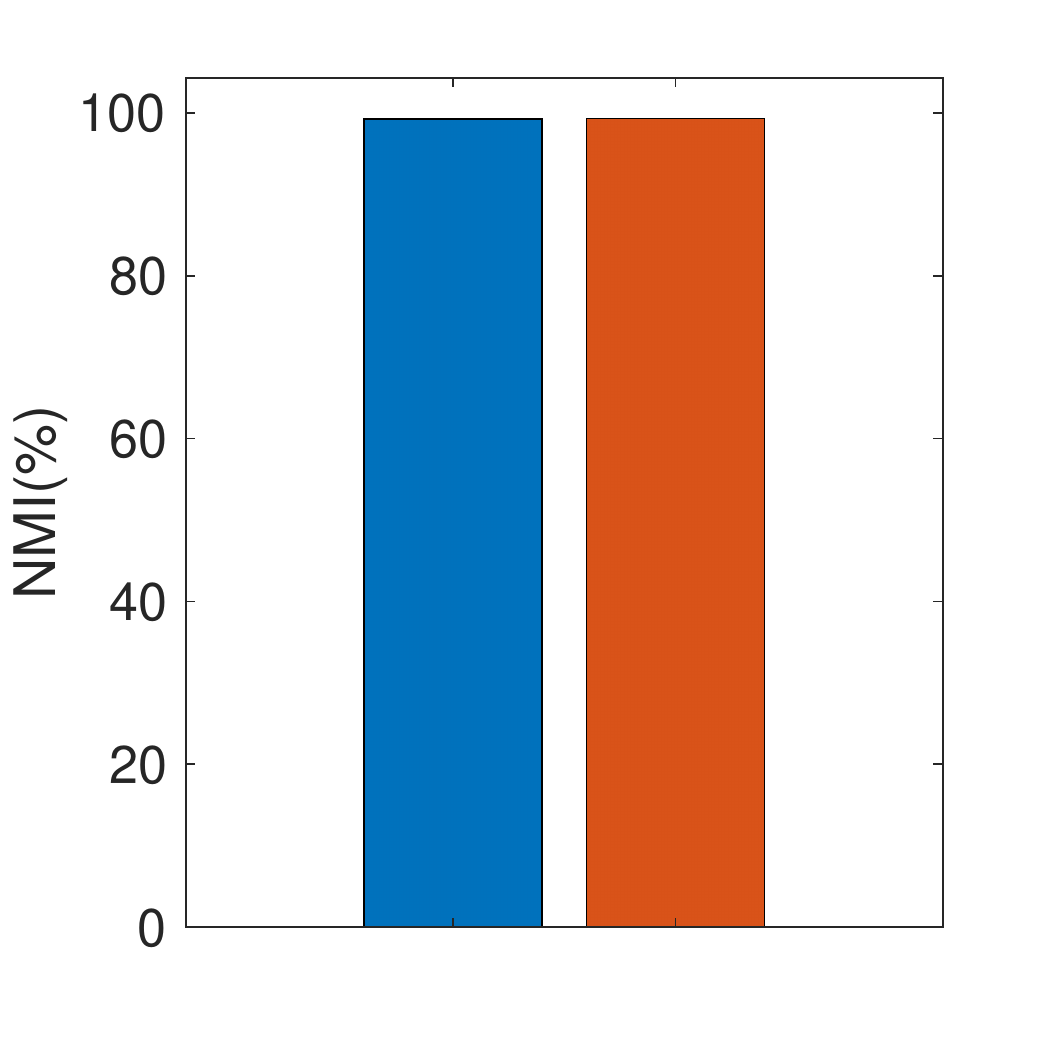}\\
      Time cost
      &\includegraphics[width=1.7cm]{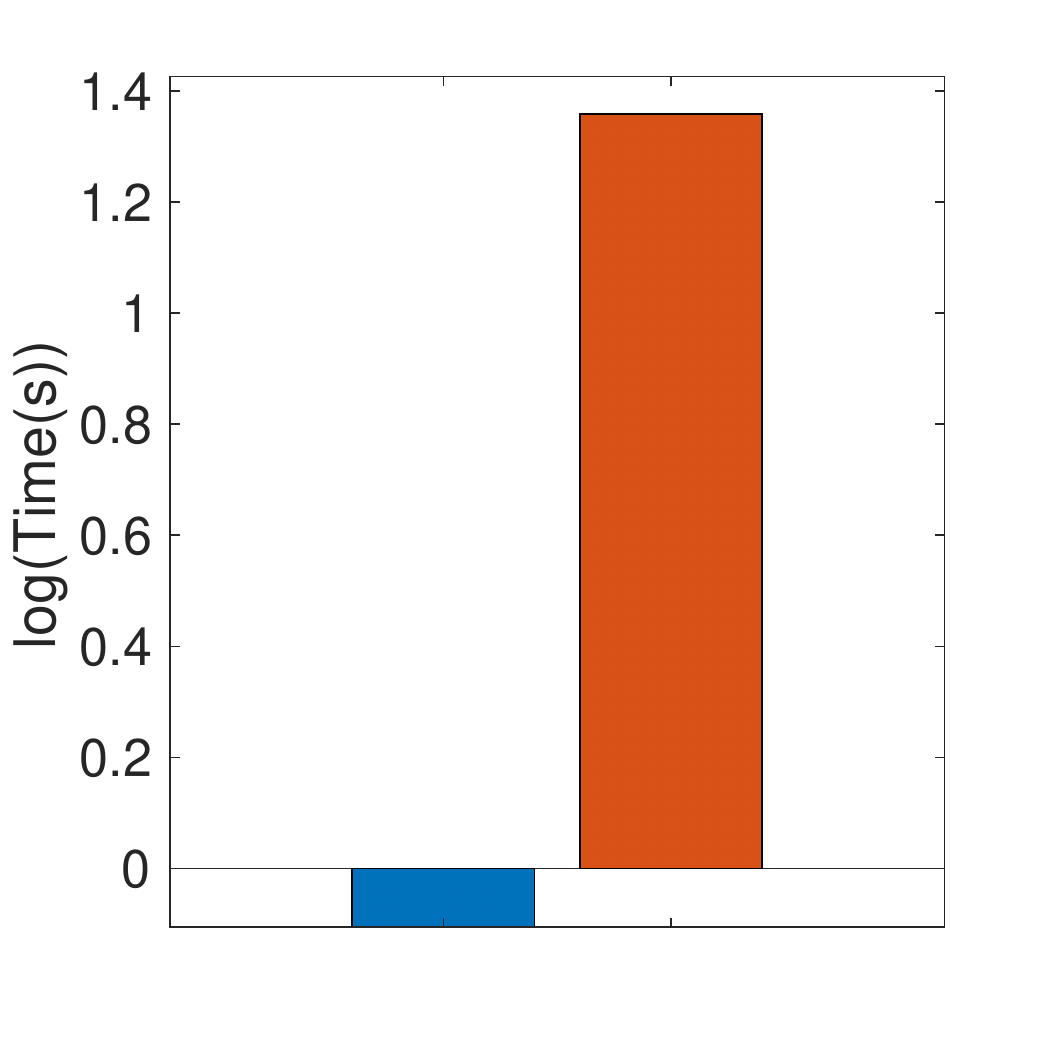}
      &\includegraphics[width=1.7cm]{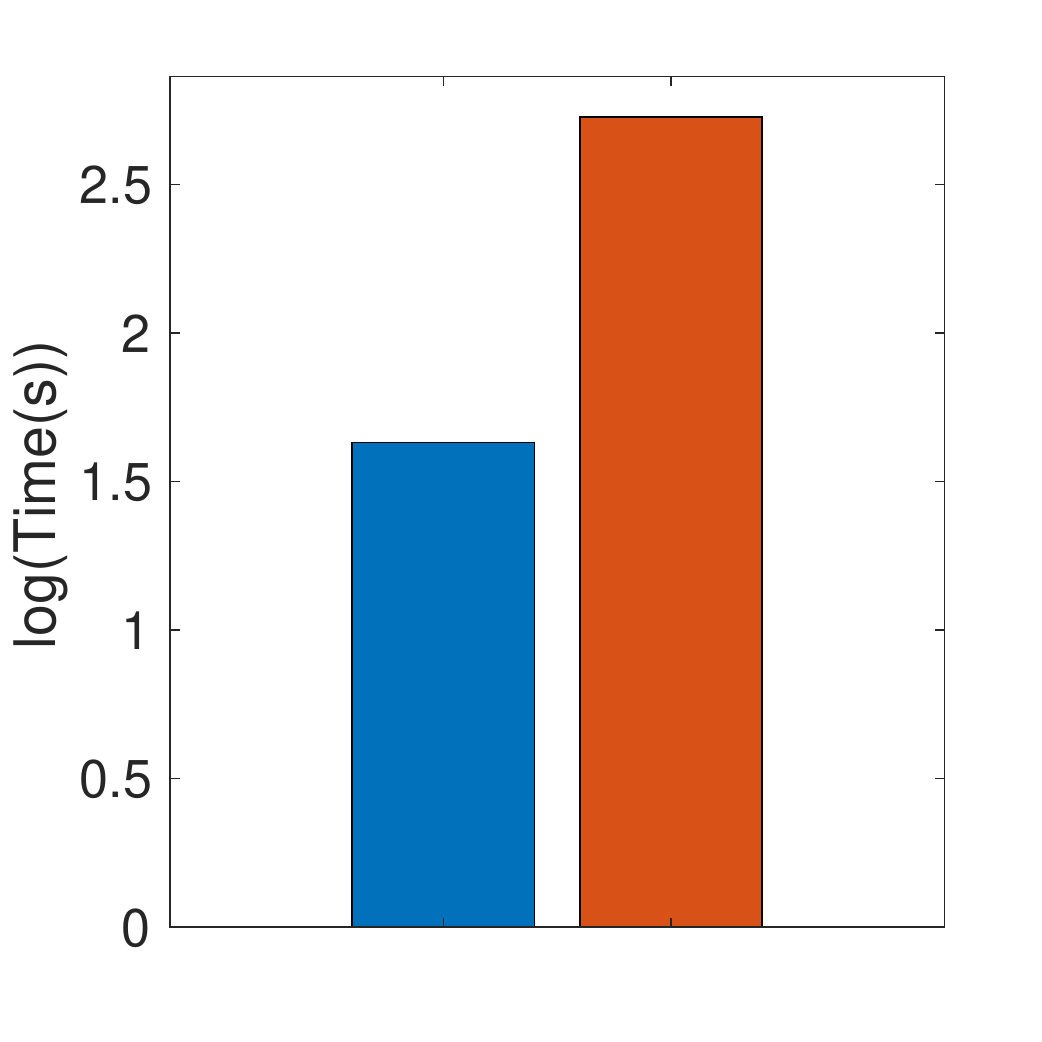}
      &\includegraphics[width=1.7cm]{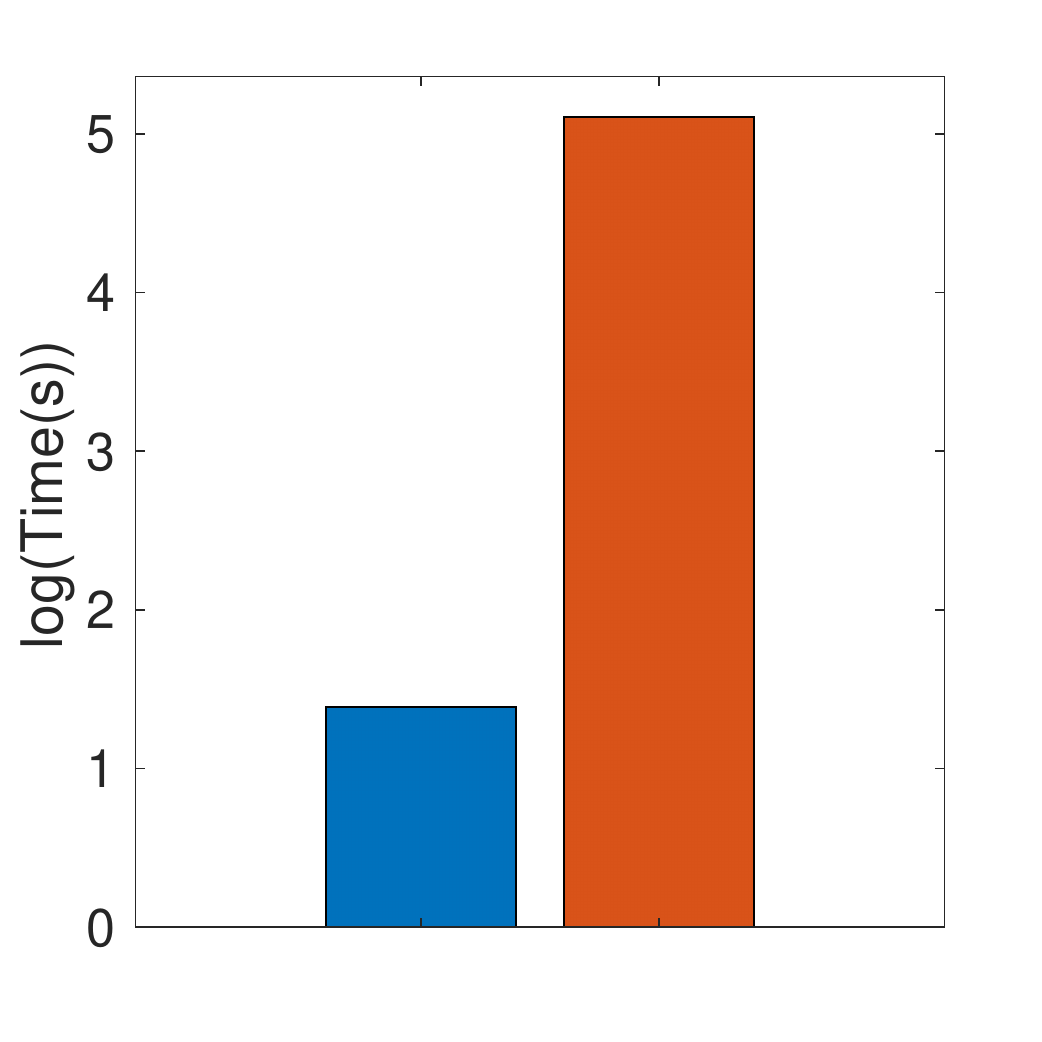}
      &\includegraphics[width=1.7cm]{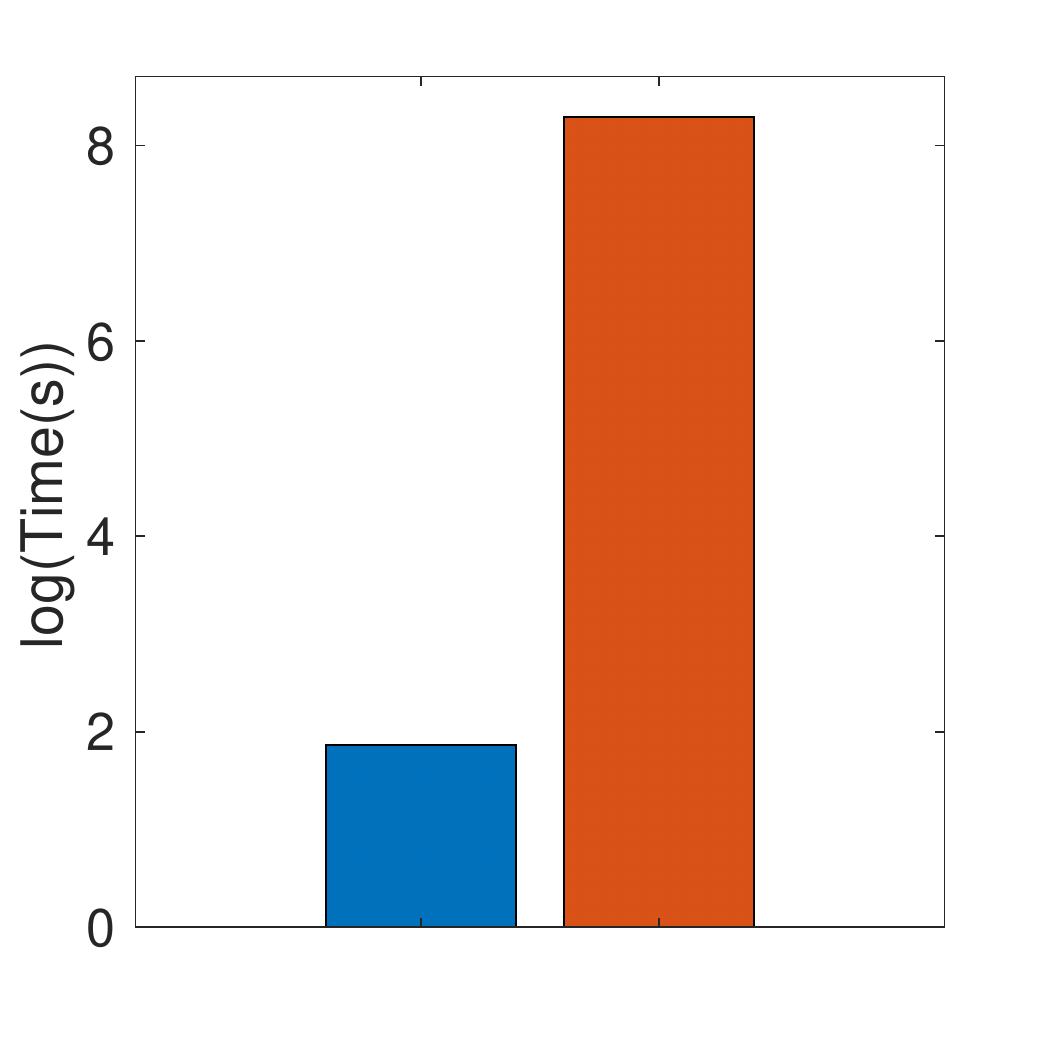}\\
      &\multicolumn{4}{c}{\includegraphics[width=7cm]{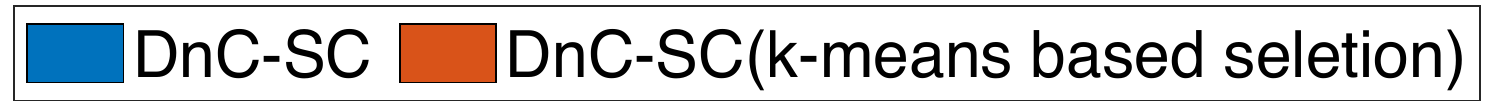}}\\
      \bottomrule
    \end{tabular}
  \end{threeparttable}
\end{table}

\begin{table}%[!t]
  \centering
  \caption{Clustering performance (ACC(\%), NMI(\%), and time costs(s)) for DnC-SC using approximate $K$-nearest landmarks and exact $K$-nearest landmarks.}
  \label{table:compare_approxKNN}
  \begin{threeparttable}
    \begin{tabular}{m{0.75cm}<{\centering}|m{1.45cm}<{\centering}m{1.45cm}<{\centering}m{1.45cm}<{\centering}m{1.45cm}<{\centering}}
      \toprule
      \emph{Data} & \emph{Letters} & \emph{MNIST} & \emph{TS-60K} & \emph{TM-1M} \\
      \midrule
      \multirow{1}{*}{ACC}
      &\includegraphics[width=1.7cm]{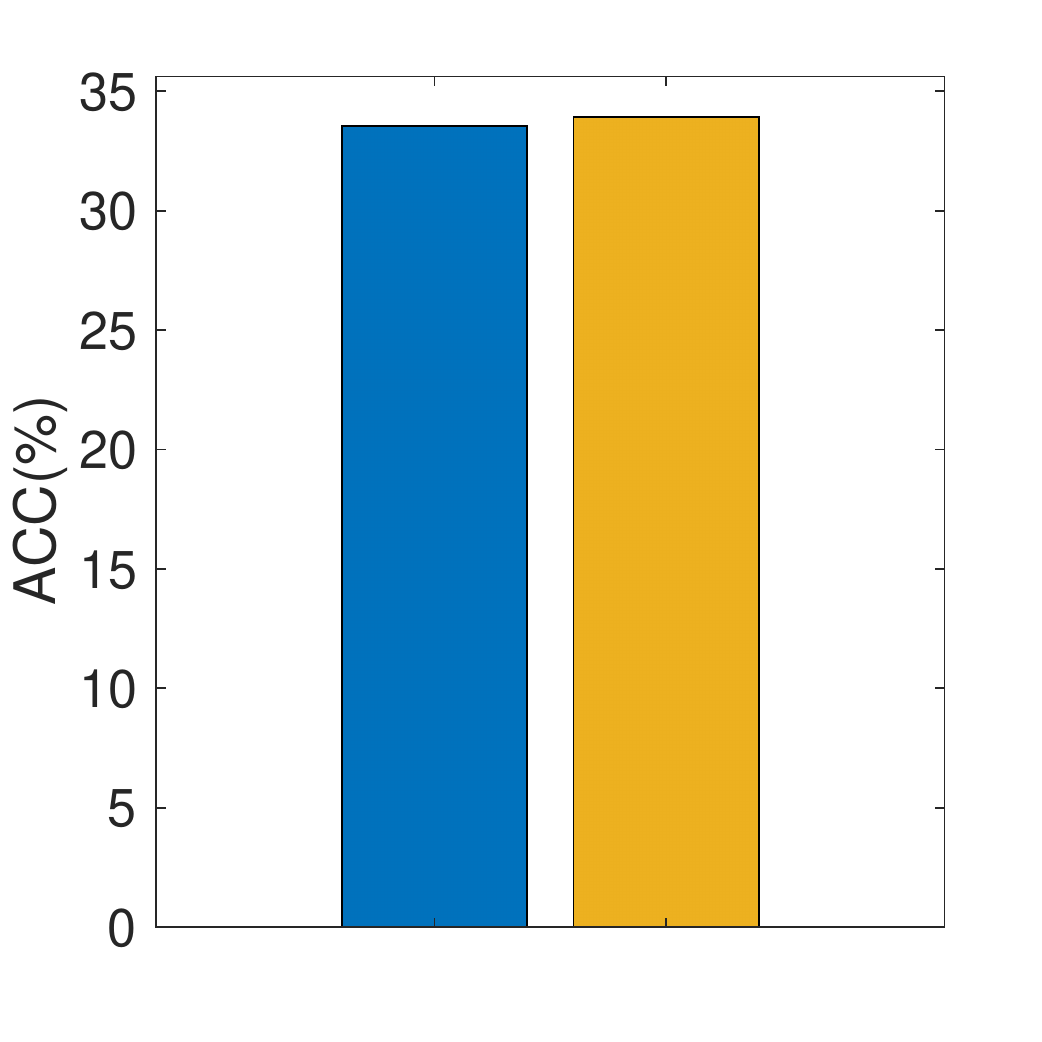}
      &\includegraphics[width=1.7cm]{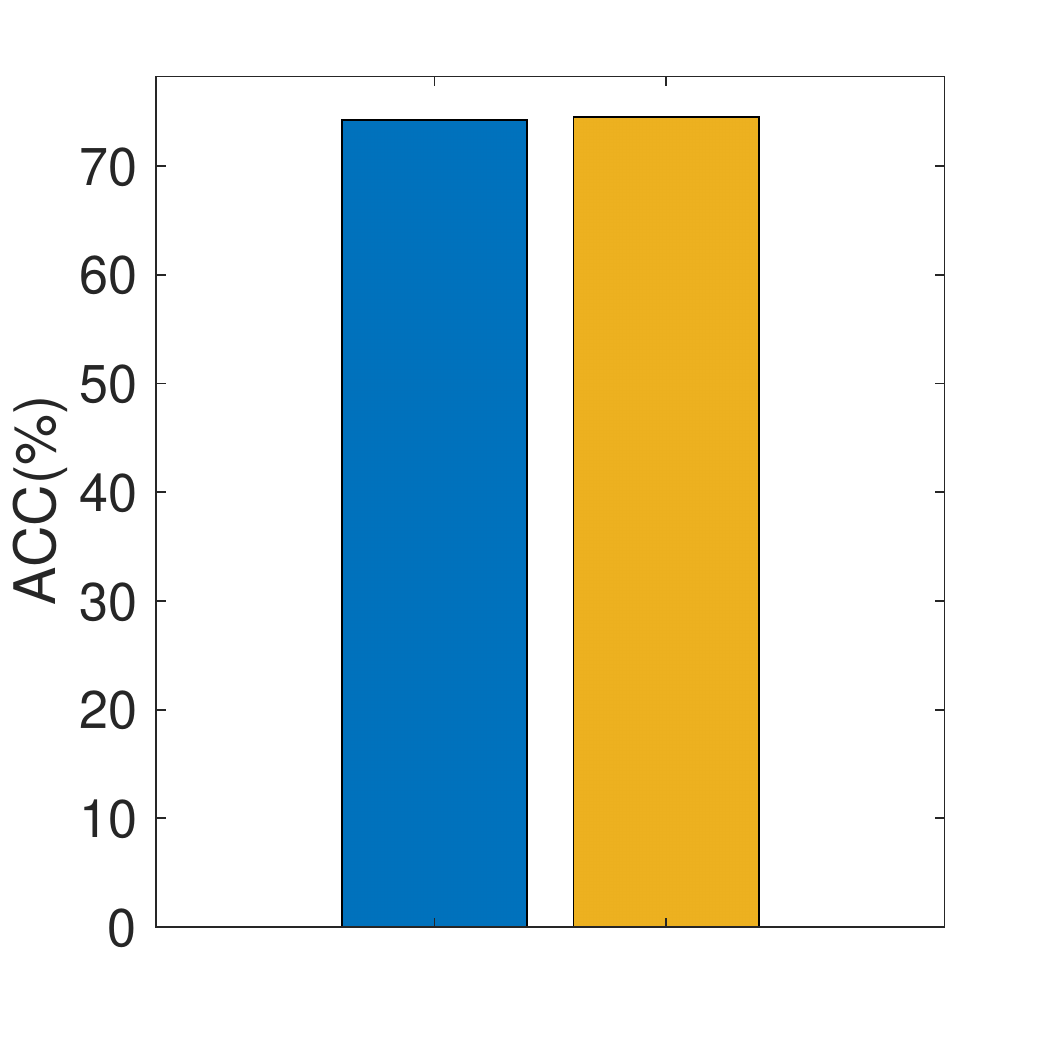}
      &\includegraphics[width=1.7cm]{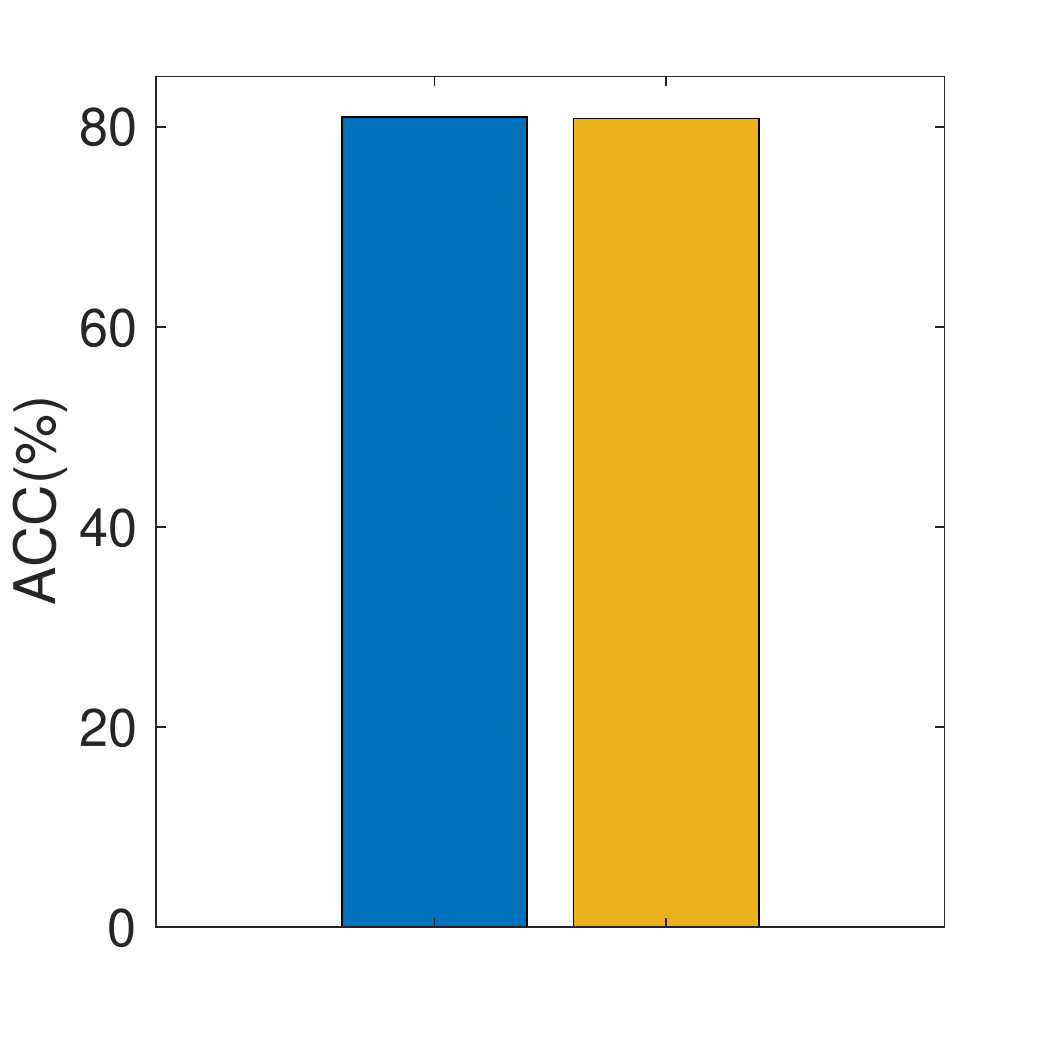}
      &\includegraphics[width=1.7cm]{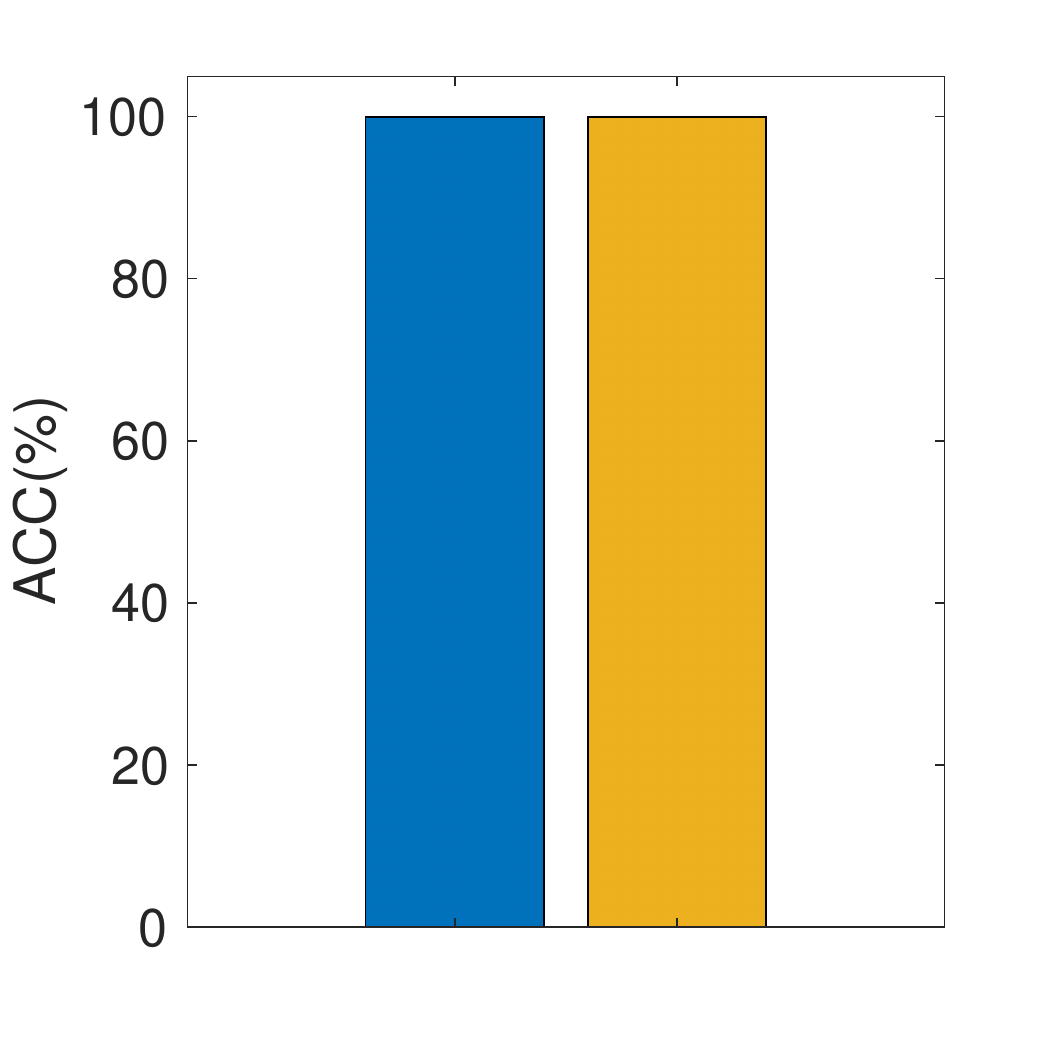}\\
      NMI
      &\includegraphics[width=1.7cm]{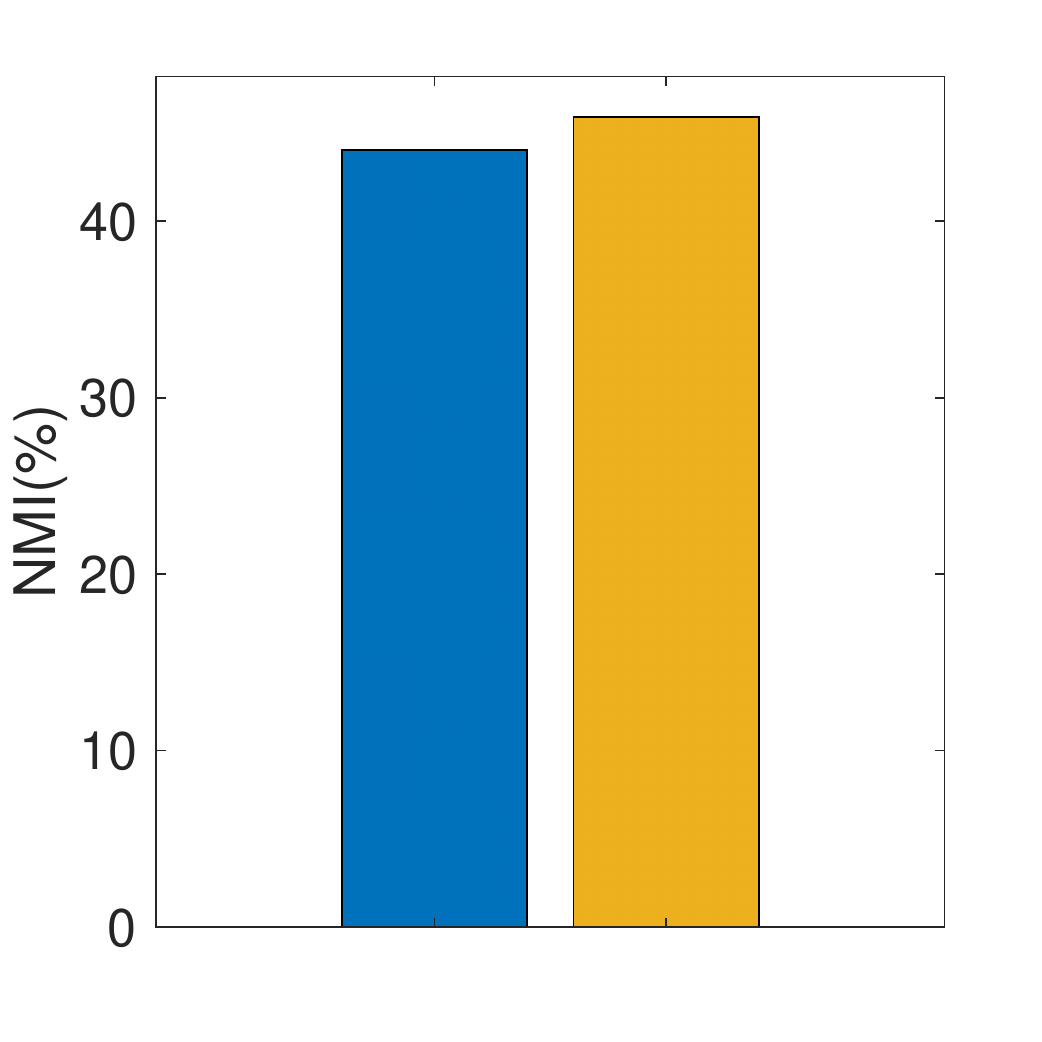}
      &\includegraphics[width=1.7cm]{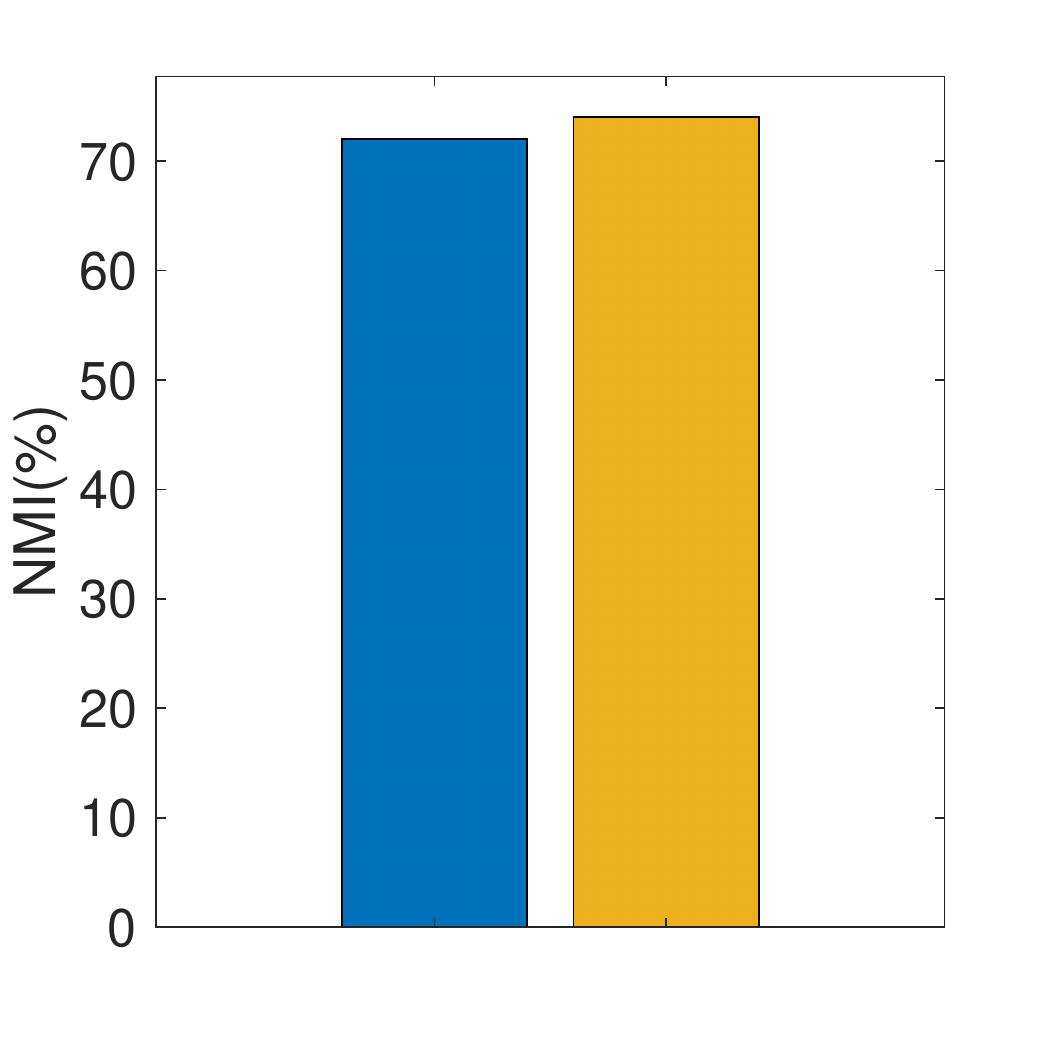}
      &\includegraphics[width=1.7cm]{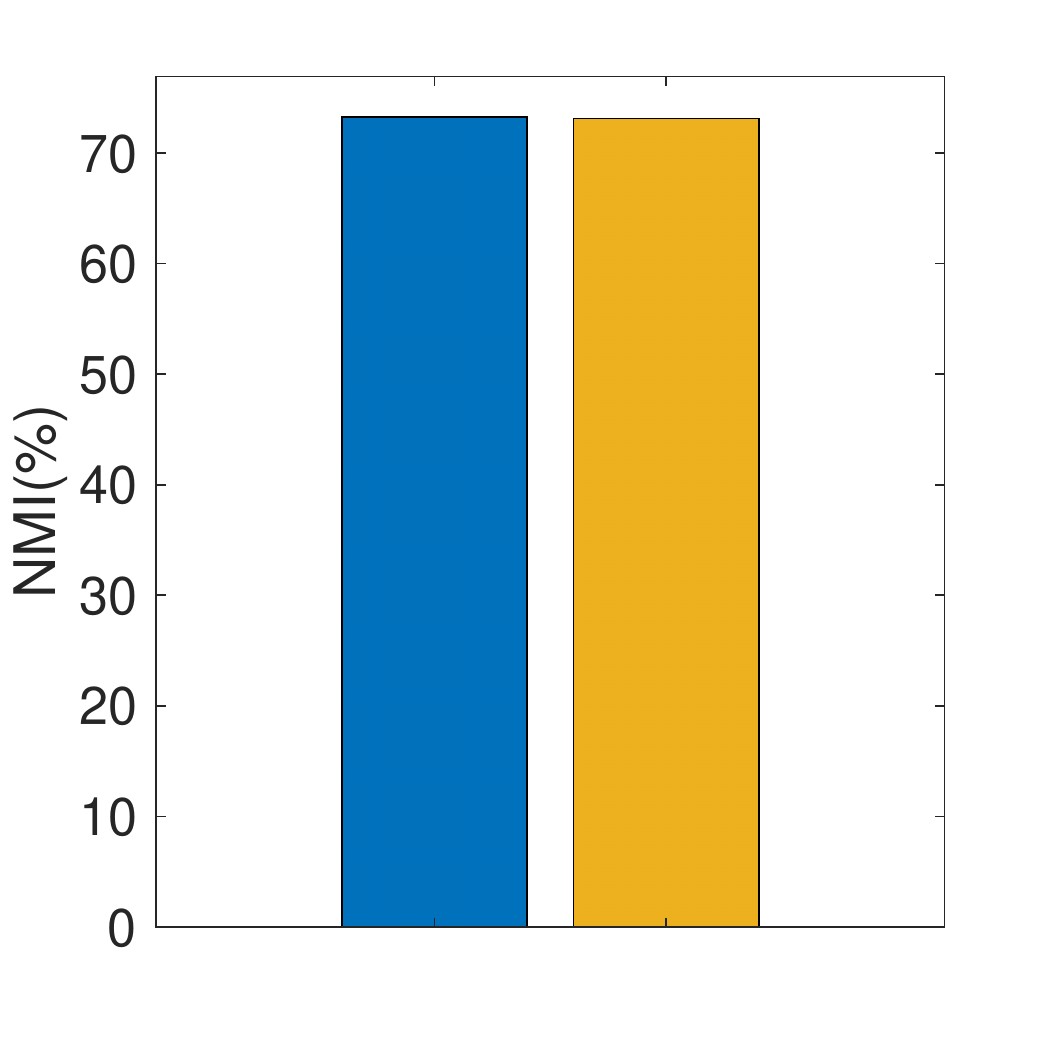}
      &\includegraphics[width=1.7cm]{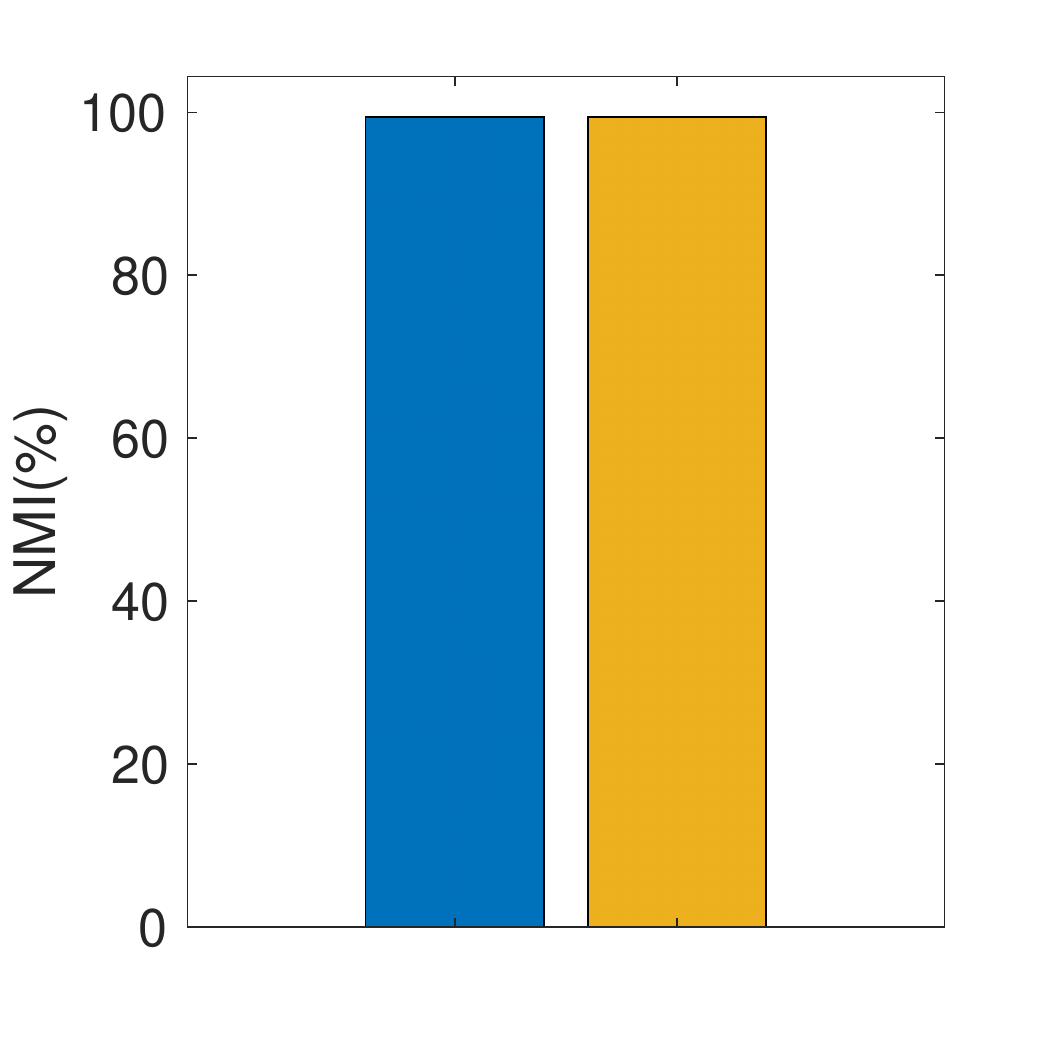}\\
      Time cost
      &\includegraphics[width=1.7cm]{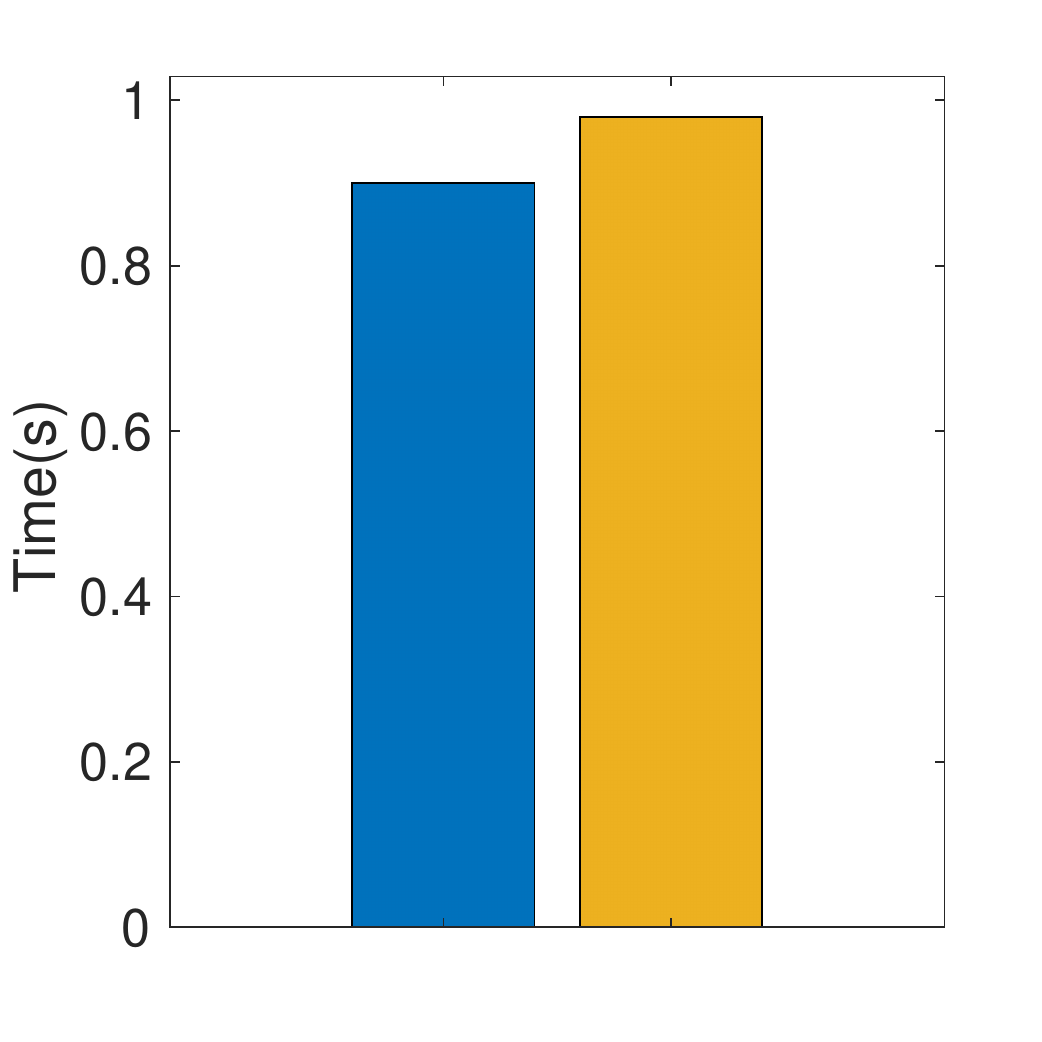}
      &\includegraphics[width=1.7cm]{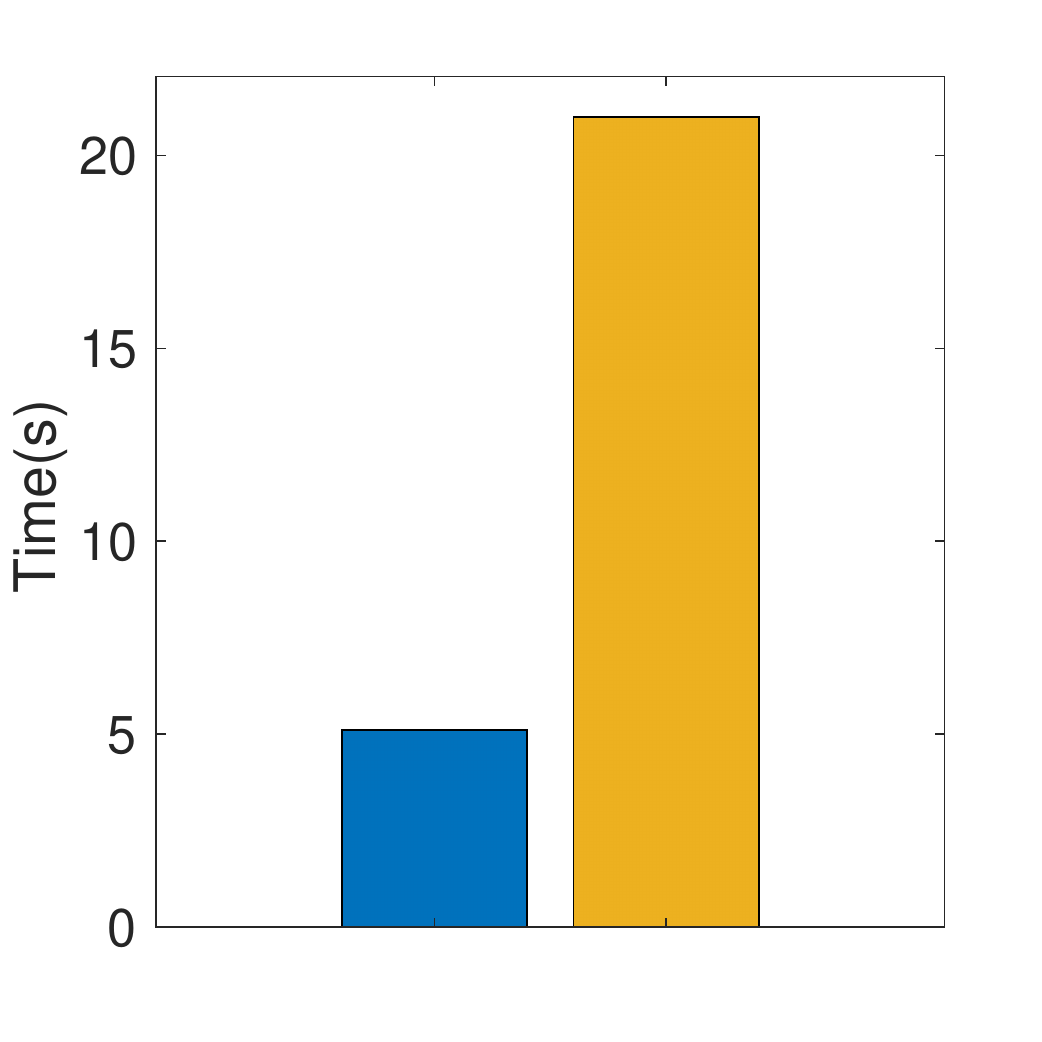}
      &\includegraphics[width=1.7cm]{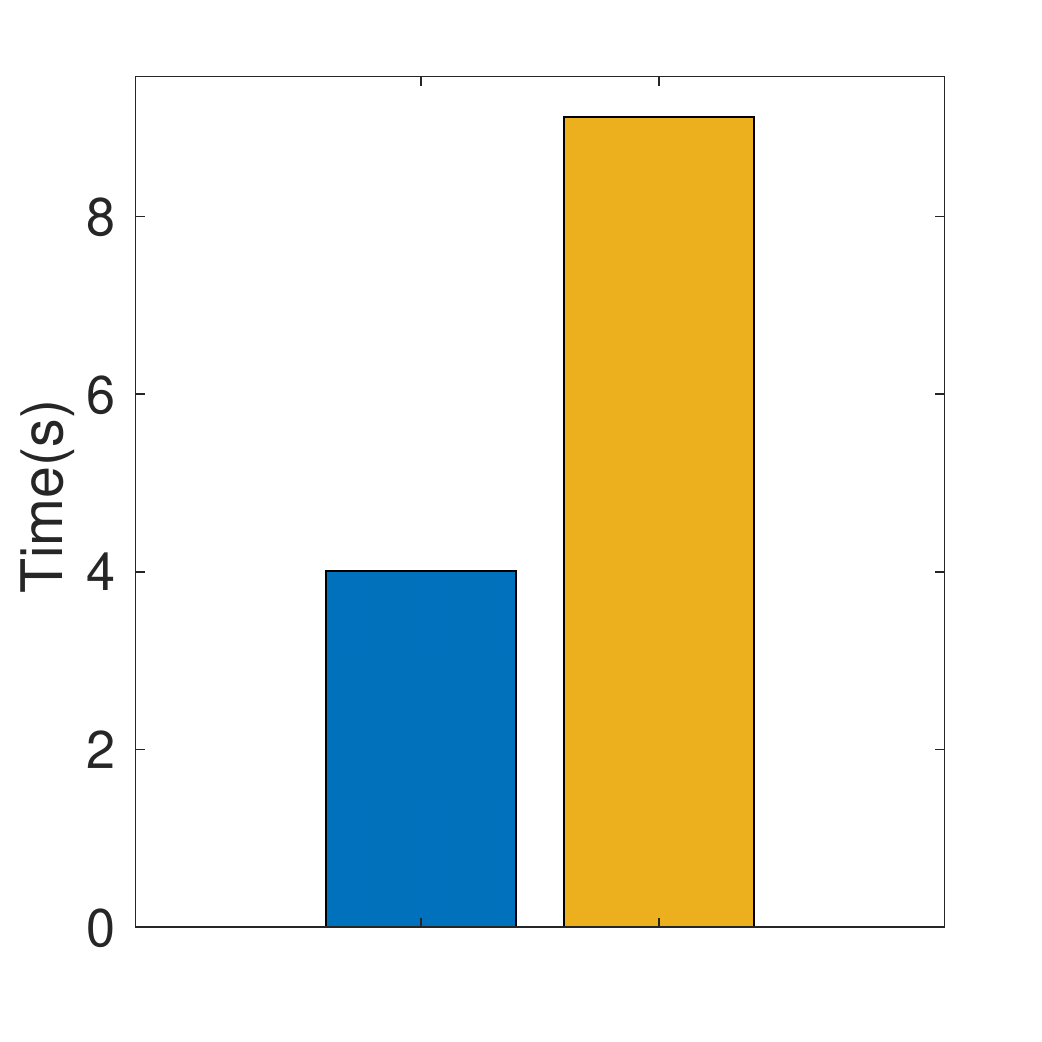}
      &\includegraphics[width=1.7cm]{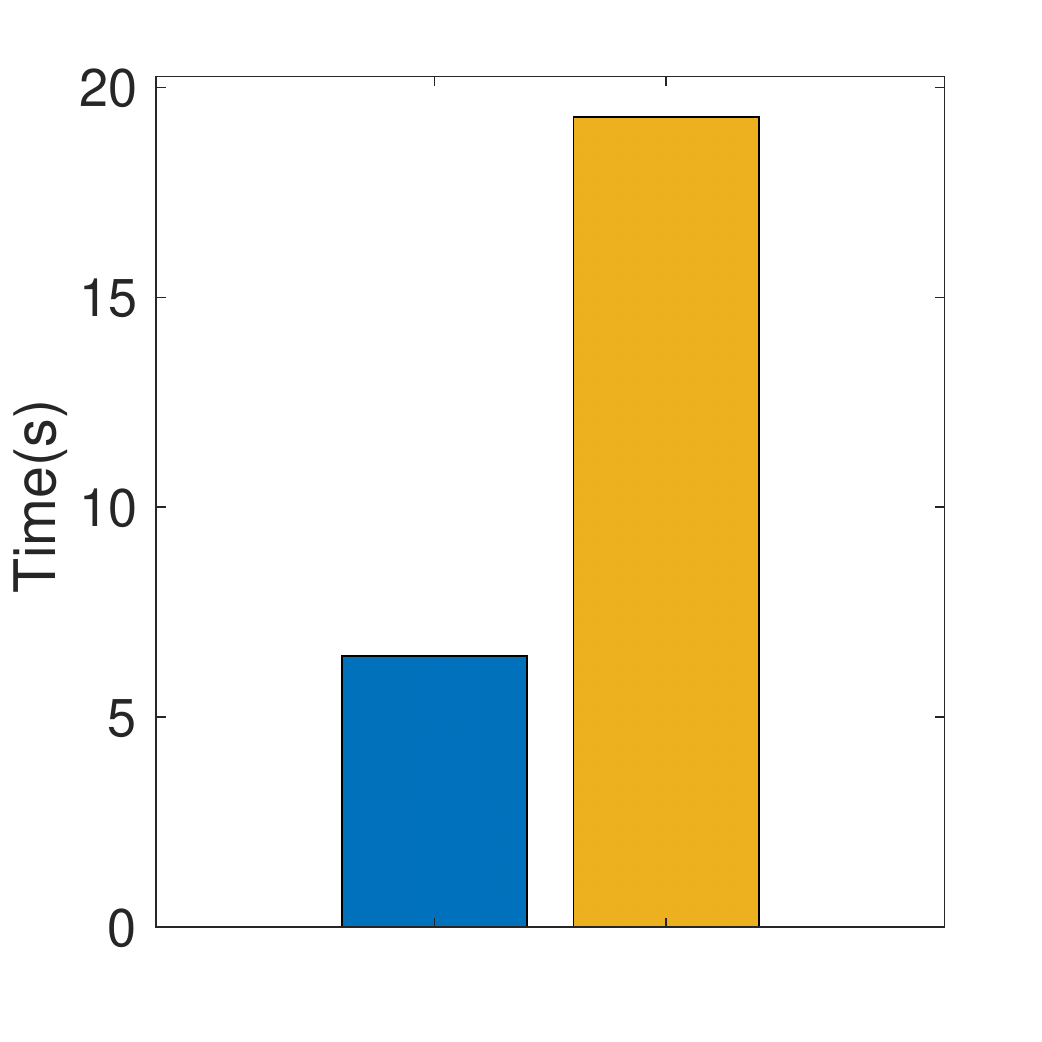}\\
      &\multicolumn{4}{c}{\includegraphics[width=5cm]{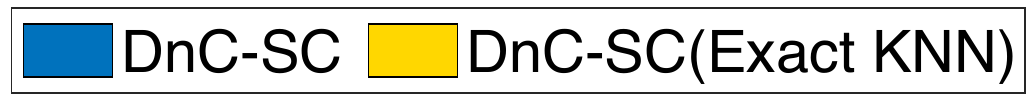}}\\
      \bottomrule
    \end{tabular}
  \end{threeparttable}
\end{table}

We conduct a series of parameters analysis experiments to demonstrate the performance of the proposed method varying different parameter settings.
We select four dataset (\emph{Letters}, \emph{MNIST}, \emph{TS-60K}, and \emph{TM-1M}) as benchmark datasets to conduct the following experiments.

\subsubsection{Number of Landmarks $p$}
\label{sec:para_p}

We first conduct parameter analysis to compare the large-scale spectral clustering methods by varying the number of landmarks $p$ (also called landmarks) and report the experimental results in Table~\ref{table:compare_para_p}.
In general, we can see that a larger value of $p$ brings a better performance of ACC and NMI but cost more time.
The proposed DnC-SC achieves the best ACC and NMI scores on all datasets except the \emph{MNIST}.
On \emph{MNIST} dataset, the proposed DnC-SC method shows the second-best ACC and NMI scores after the LSC-K method.
In terms of time cost, the proposed DnC-SC method shows the best efficiency on all datasets.
Overall, the proposed DnC-SC method shows significant effectiveness and efficiency in this comparison.

\subsubsection{Number of Nearest Landmarks $K$}
\label{sec:para_K}

We then conduct parameter analysis to compare the large-scale spectral clustering methods by varying the number of the nearest landmark $K$ and report the experimental results in Table~\ref{table:compare_para_Knn}.
Note that the Nystr{\"{o}}m method does not have the parameter $K$.
Therefore, we do not show the results of the Nystr{\"{o}}m method in this experiment.
According to Table ~\ref{table:compare_para_Knn}, the performance of most methods varies for different $K$ values.
The proposed method shows the best ACC and NMI of performance for the three of four datasets, and the second-best ACC and NMI on the \emph{MNIST} dataset.
Overall, the proposed DnC-SC shows superior effectiveness and the best efficiency on this comparison.

\subsubsection{Number of Nearest Landmarks $K$ and selection rate $\alpha$}
\label{sec:para_K_u}

To further demonstrate the proposed method, we evaluate the performances by varying parameters $K$ and $\alpha$ and report the experimental results in Table~\ref{table:compare_para_K_u}.
For proposed DnC-SC methods, the selection rate parameter $\alpha$ directly affects the computational complexity of landmark selection, while the number of nearest landmarks $K$ affects similarity construction, respectively.
As we can see, a larger $K$ or $\alpha$ generally leads more time cost while not necessarily achieves better performance.
Overall, the proposed method shows considerable robustness with various parameters on ACC and NMI.

\subsubsection{Efficiency analysis}

To explore the efficiency of the proposed method in each computational phase, we report the time costs of three different phrases: landmark selection, similarity construction, and graph partitioning.
We choose LSC-K, LSC-R, and U-SPEC algorithms that have similar mechanisms for comparison.
We list the strategies and methods used in each method in Table \ref{table:compare_three_phases}.
The experimental results are reported in Table \ref{table:compare_intime}.

For landmark selection, the LSC-K and LSC-R methods apply $k$-means and a random selection, respectively; the U-SPEC method uses a hybrid selection that conducts $k$-means on a small set of random candidates;
DnC-SC utilizes the divide-and-conquer selection.
Looking at the runtime of the landmark selection, we see that the random selection of LSC-R takes a little time, while the $k$-means selection takes much more time.
The divide-and-conquer selection of DnC-SC is the second-fastest method just behind the random selection.

For similarity construction, the LSC-K and LSC-R compute the exact similarity matrix without approximation, while U-SPEC and DnC-SC calculate the similarity by approximate schemes.
Compared with U-SPEC, DnC-SC uses the results of landmark selection to improve the approximate scheme.
For the runtime of similarity construction, we find that DnC-SC takes significantly less time than other methods, especially for the larger-scale dataset (\emph{TM-1M}).
Note that the approximate similarity matrix of U-SPEC takes more time than LSC-K or LSC-R in \emph{MNIST} dataset.
However, the similarity of U-SPEC takes less computational complexity than LSC-K or LSC-R.
This is because U-SPEC uses serial calculations in the approximation process.
In MATLAB, it will be much faster to perform the approximation in a batch processing manner (with optimized matrix computation) than in a serial processing manner.

For graph partitioning, LSC-K and LSC-R utilize SVD based method, while U-SPEC and DnC-SC apply transfer cuts.
Theoretically, both two graph partitioning methods can be considered as efficient solutions for bipartite graph partitioning \cite{li2012segmentation,cai2014large}.
But the transfer cuts take less computational complexity.
In Table \ref{table:compare_intime}, we can see that U-SPEC and DnC-SC take less time than LSC-K and LSC-R, which is consistent with the theoretical complexity.

Overall, DnC-SC shows the best efficiency in four methods, which is mainly due to the proposed landmark selection and approximate similarity construction.

\begin{table*}[]
  \centering
  \caption{Comparison for three phases for different methods.}
  \label{table:compare_three_phases}
  \begin{tabular}{@{}l|llll@{}}
    \toprule
    Phase                   & LSC-K     & LSC-R     & U-SPEC                          & DnC-SC                       \\ \midrule
    Landmark Selection      & $k$-means & Random    & Hybrid representative selection & Divide-and-conquer selection \\
    Similarity Construction & Exact     & Exact     & Approximate                     & Approximate                  \\
    Graph Partitioning      & SVD based & SVD based & Transfer cuts                   & Transfer cuts                \\ \bottomrule
  \end{tabular}
\end{table*}

\begin{table}%[!t]
  \centering
  \caption{Comparison of time costs in each phase for different methods.}
  \label{table:compare_intime}
  \begin{threeparttable}
    \begin{tabular}{m{0.8cm}<{\centering}|m{6cm}<{\centering}}
      \toprule
      \emph{Data} & Time costs                                                   \\

      \midrule
      \multirow{1}{*}{\emph{Letters}}
                  & \includegraphics[width=6cm]{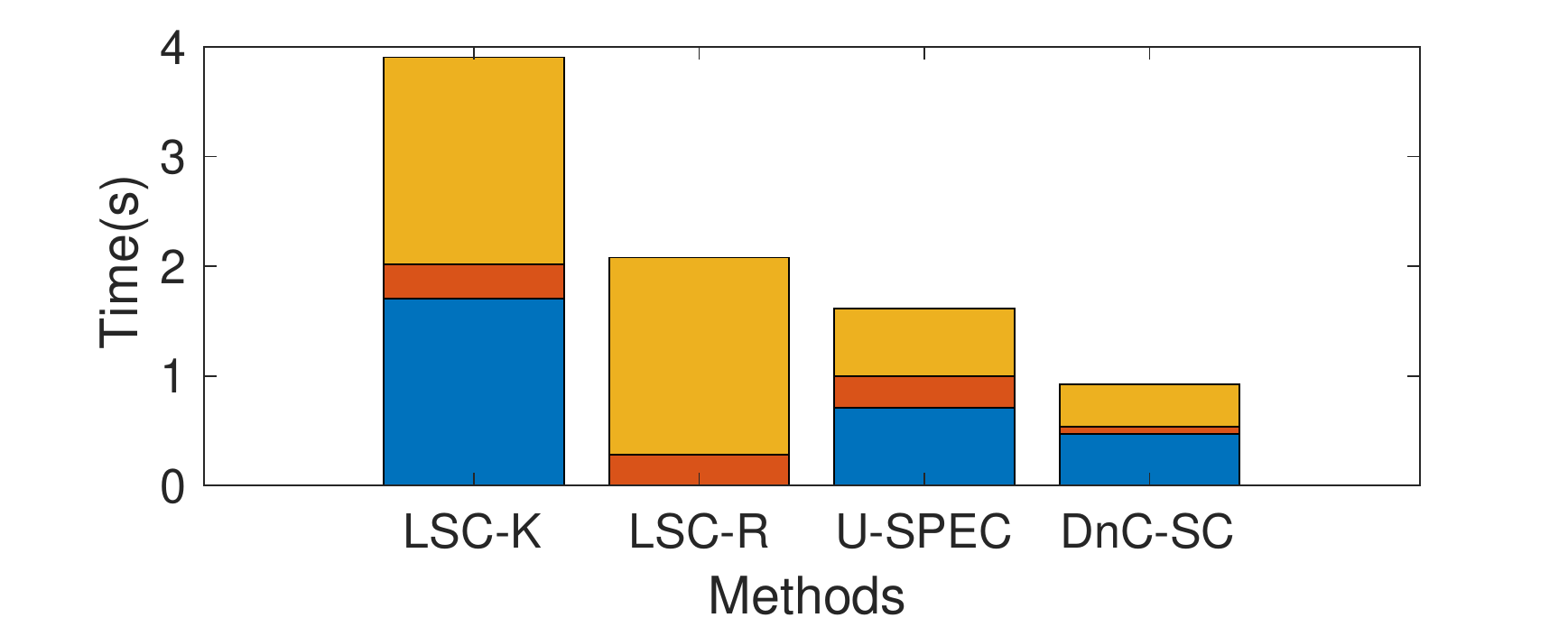} \\
      \emph{MNIST}
                  & \includegraphics[width=6cm]{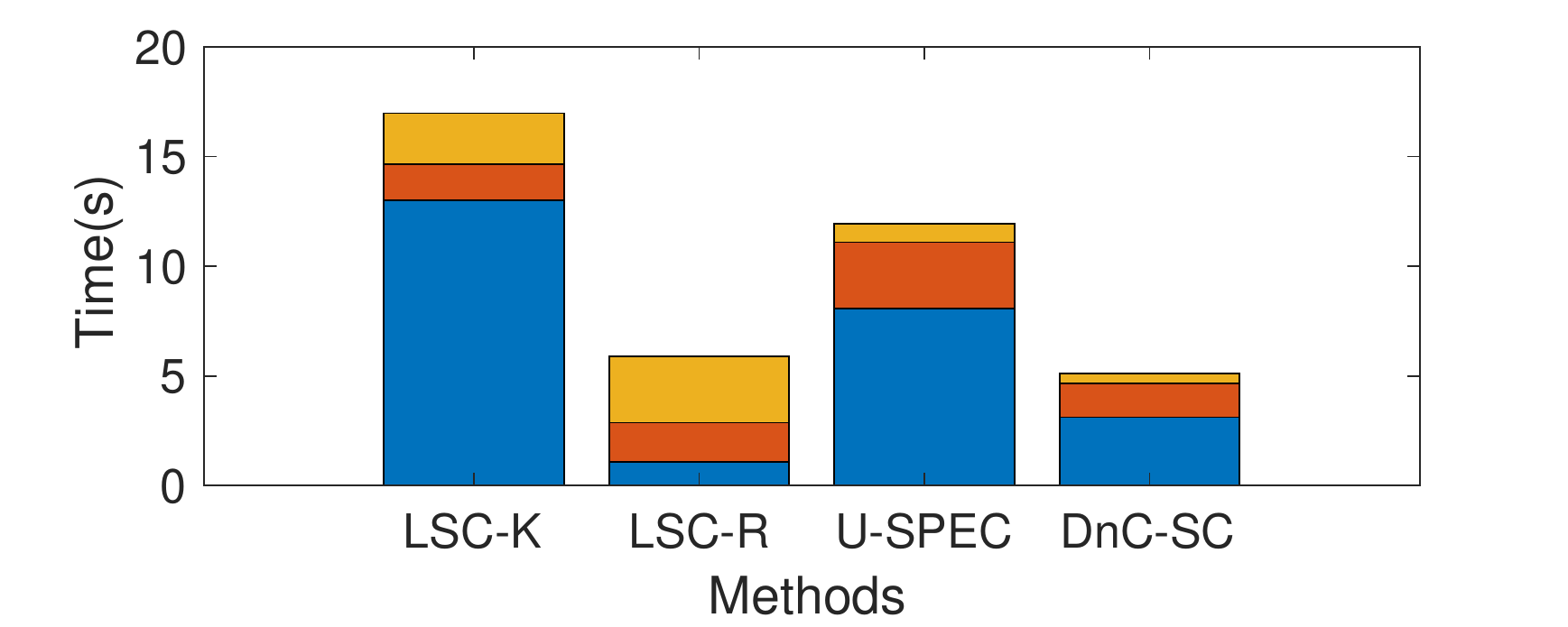}   \\
      \emph{TS-60K}
                  & \includegraphics[width=6cm]{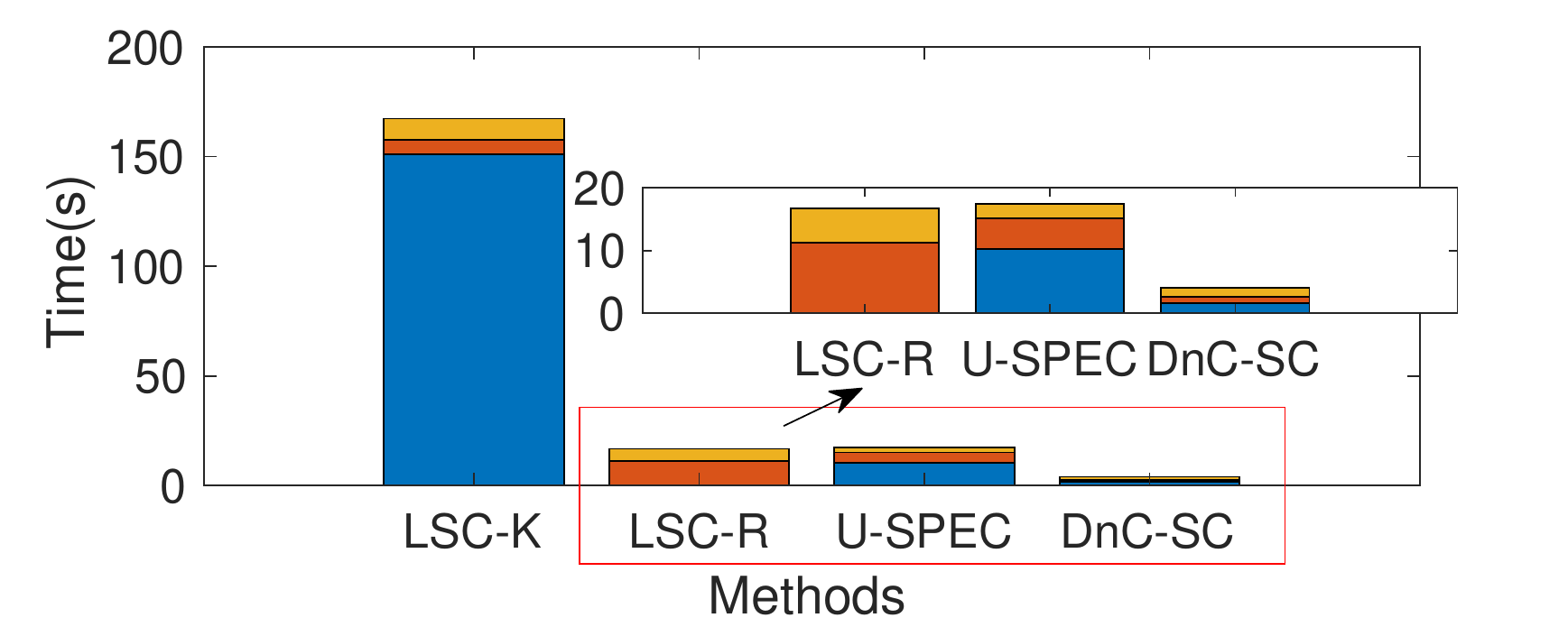}  \\
      \emph{TM-1M}
                  & \includegraphics[width=6cm]{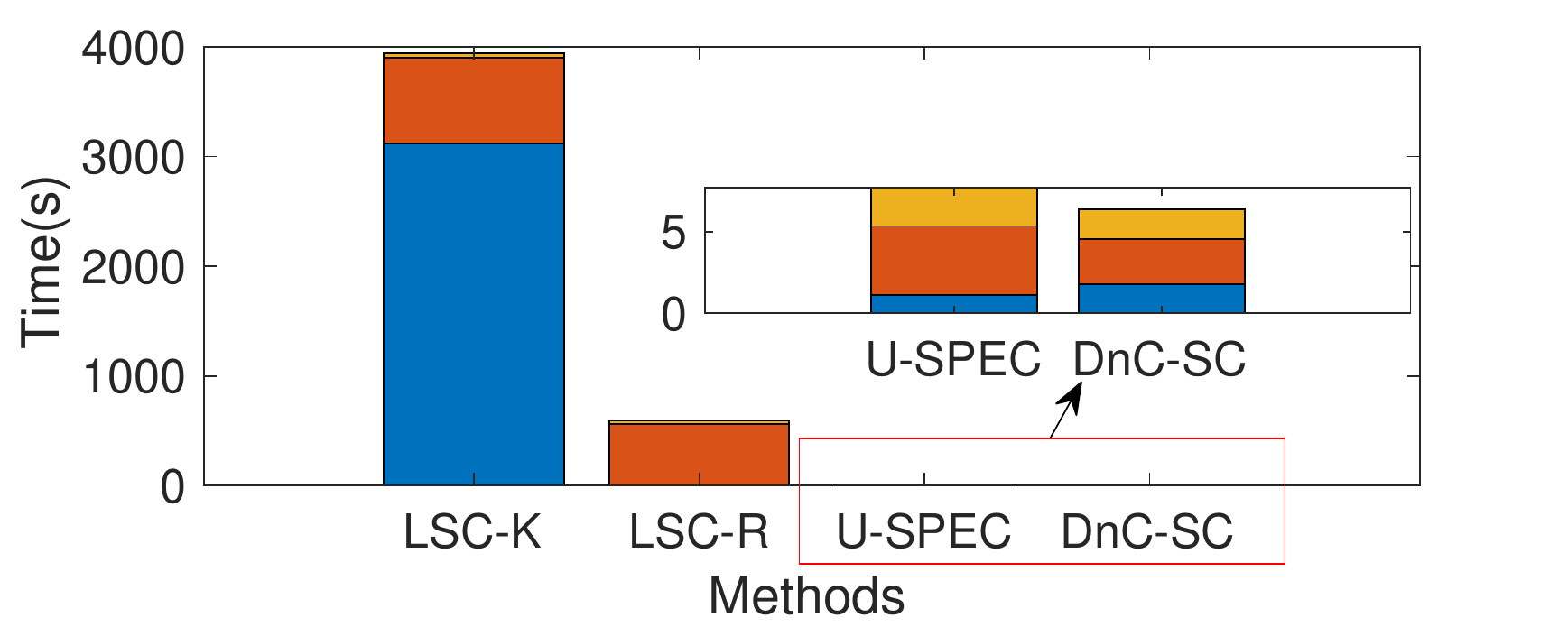}   \\
                  & {\includegraphics[width=6cm]{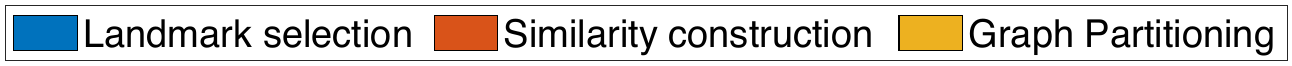}}        \\
      \bottomrule
    \end{tabular}
  \end{threeparttable}
\end{table}

\subsection{Influence of Landmark Selection Strategies}
\label{sec:cmpSelStrat}

Some existing works have shown that the performance of large-scale spectral clustering heavily relies on the proper strategy of landmark selection \cite{li2020hubness}.
In our proposed landmark selection, we propose a divide-and-conquer selection strategy and light-$k$-means to find a good balance between effectiveness and efficiency.
We test the purposed method with different landmark selection methods, i.e., $k$-means based landmark selection, divide-and-conquer selection without light-$k$-means, and divide-and-conquer selection with light-$k$-means.

In this section, we compare the performances between the divide-and-conquer based landmark selection and the $k$-means base landmark selection.
The experimental results are reported in Table~\ref{table:compare_sel_strategies}.
As we mentioned, the divide-and-conquer based landmark selection algorithm recursively solves the optimization problems \ref{eq: opt}, which $k$-means methods can also solve.
We have pointed out the lack of efficiency of directly applying $k$-means on large-scale datasets in Section \ref{sec:dnc-sc_complexity}.
Note that the number of maximum iterations of $k$-means in landmark selection is turned as 5, which is the same setting as LSC-K and U-SPEC implementation.
In Table~\ref{table:compare_sel_strategies}, $k$-means based landmark selection algorithm generally shows better ACC and NMI on most datasets except \emph{TM-1M} dataset, while the difference in performance is not significant.
Compared to $k$-means based selection, our divide-and-conquer based landmark selection algorithm strikes a balance between efficiency and effectiveness.
It achieves significantly better efficiency than the $k$-means based selection and yields competitive clustering quality compared to the $k$-means based selection.

\subsubsection{Performance comparison on simulation scenarios}
\begin{figure}%[!t]
  \begin{center}
    {\subfigure[\emph{500 Gaussian blobs} ]
      {\includegraphics[width=0.45\columnwidth]{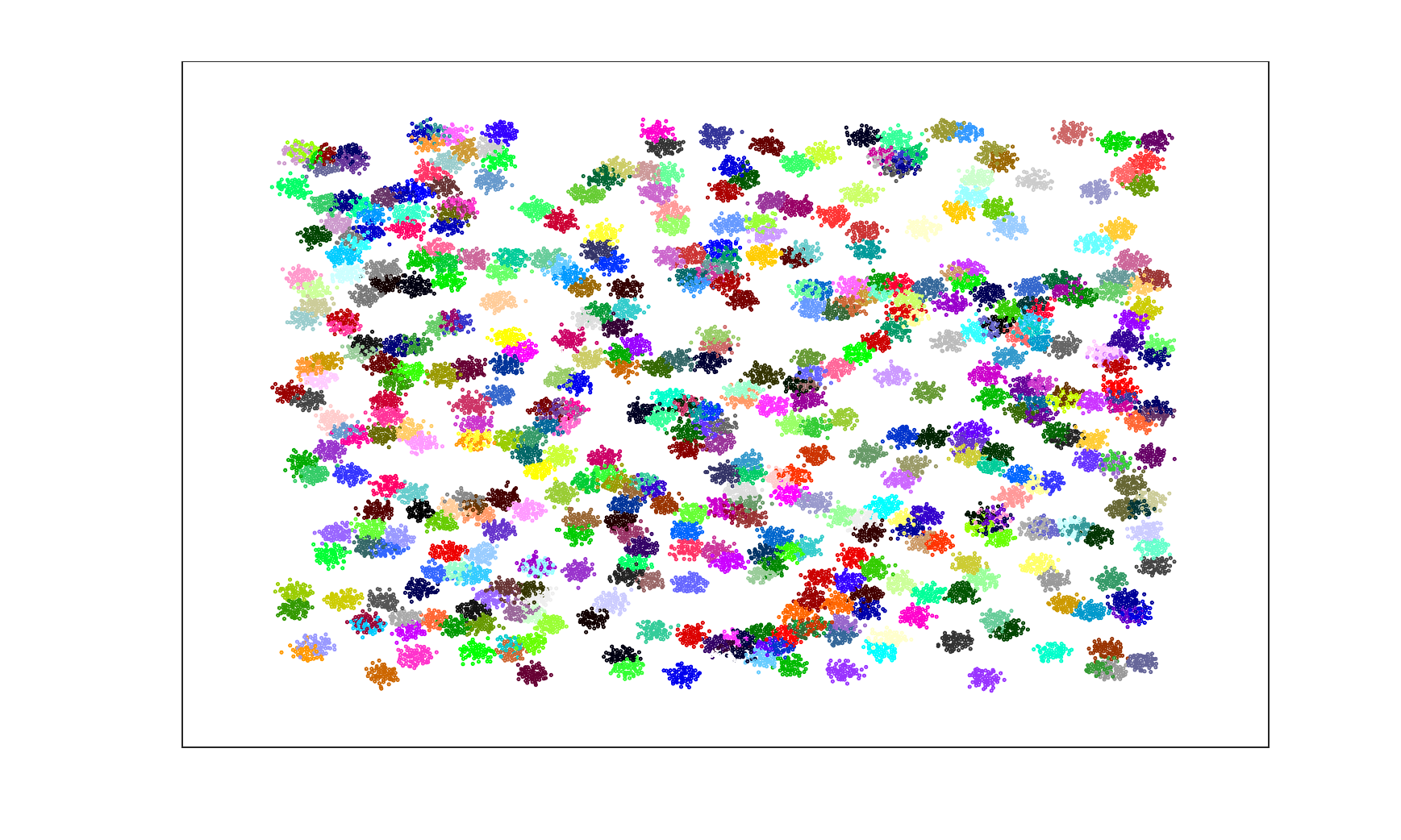}}}
    {\subfigure[\emph{1000 Gaussian blobs} ]
      {\includegraphics[width=0.45\columnwidth]{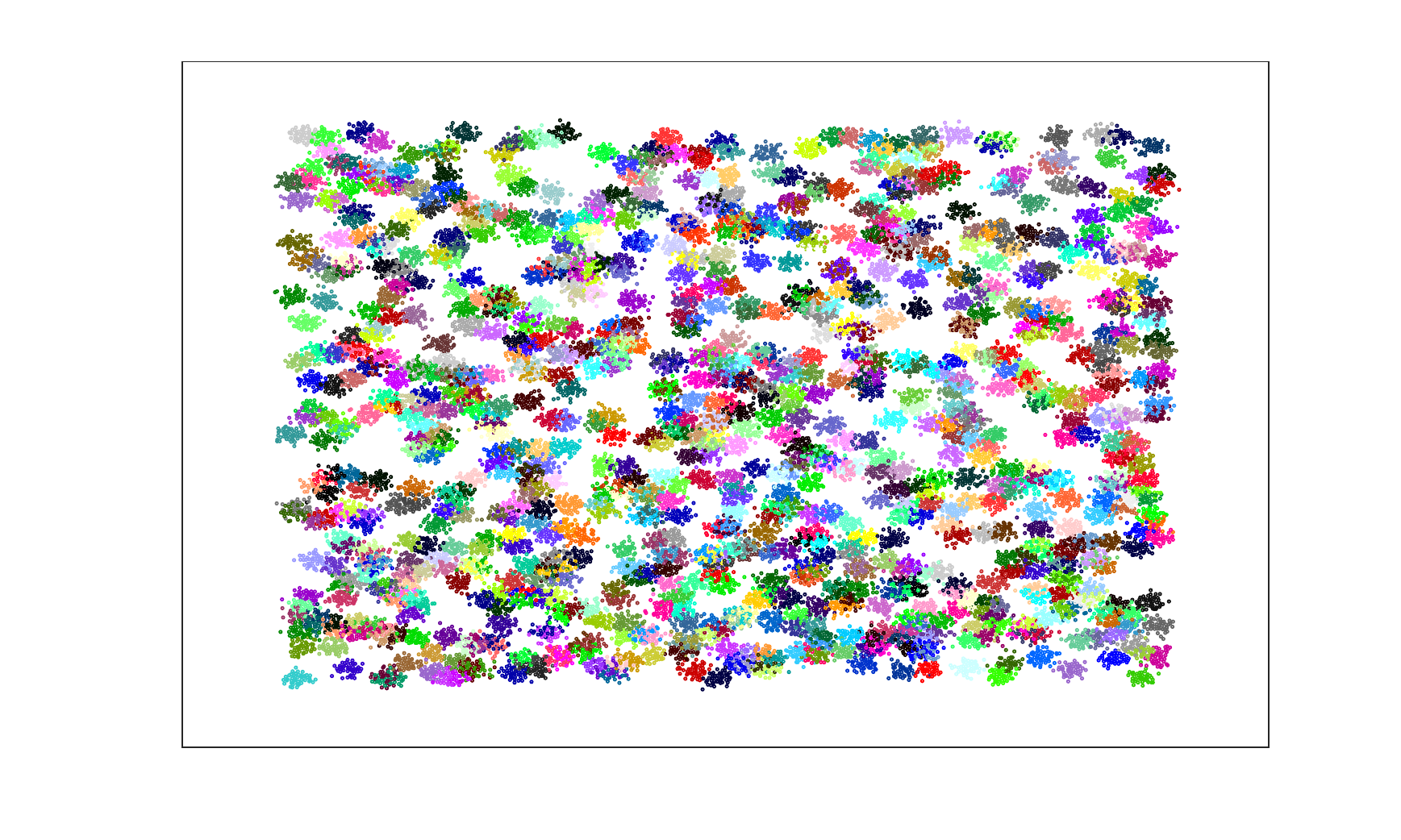}}}
    {\subfigure[\emph{1500 Gaussian blobs} ]
      {\includegraphics[width=0.45\columnwidth]{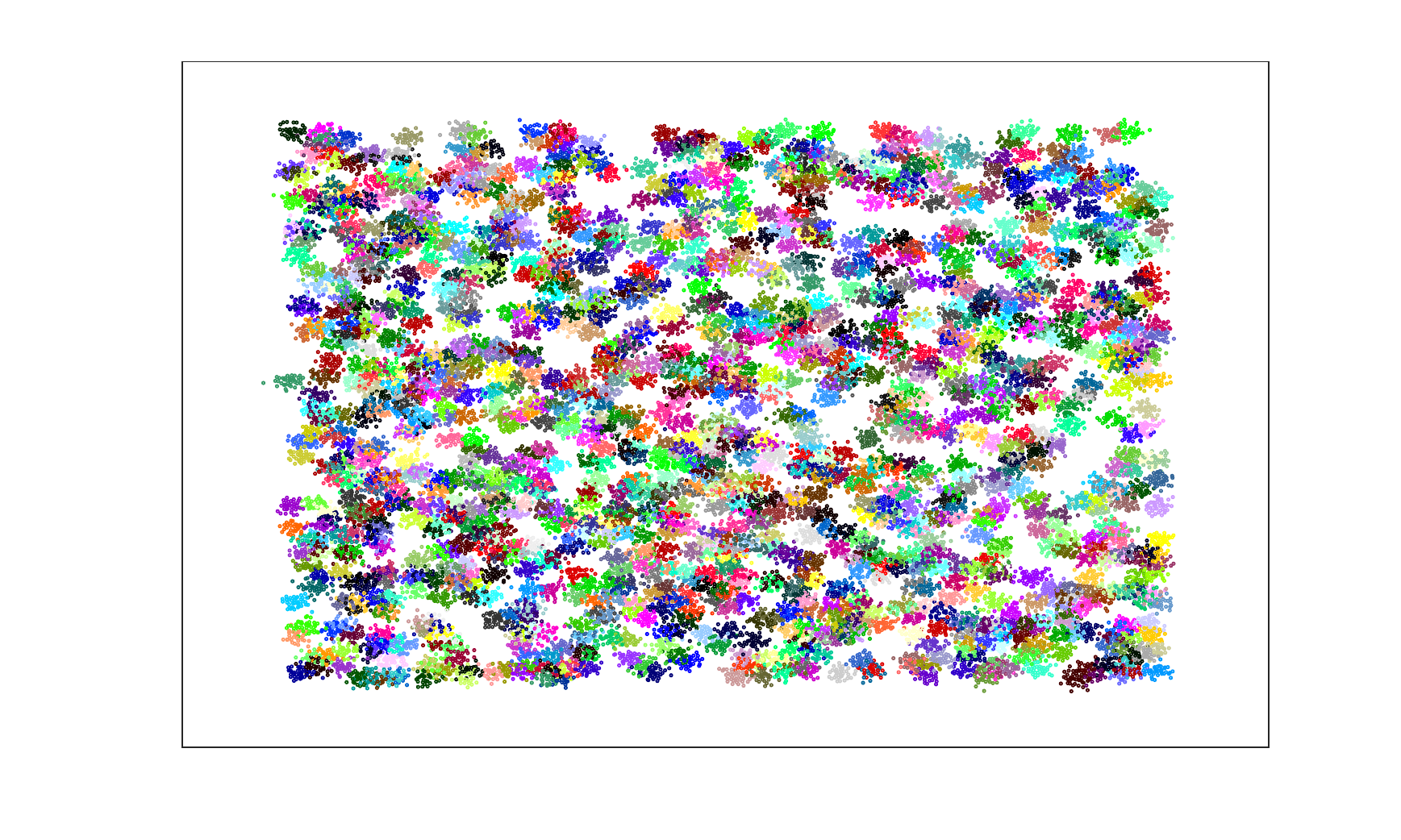}}}
    {\subfigure[\emph{2000 Gaussian blobs}]
      {\includegraphics[width=0.45\columnwidth]{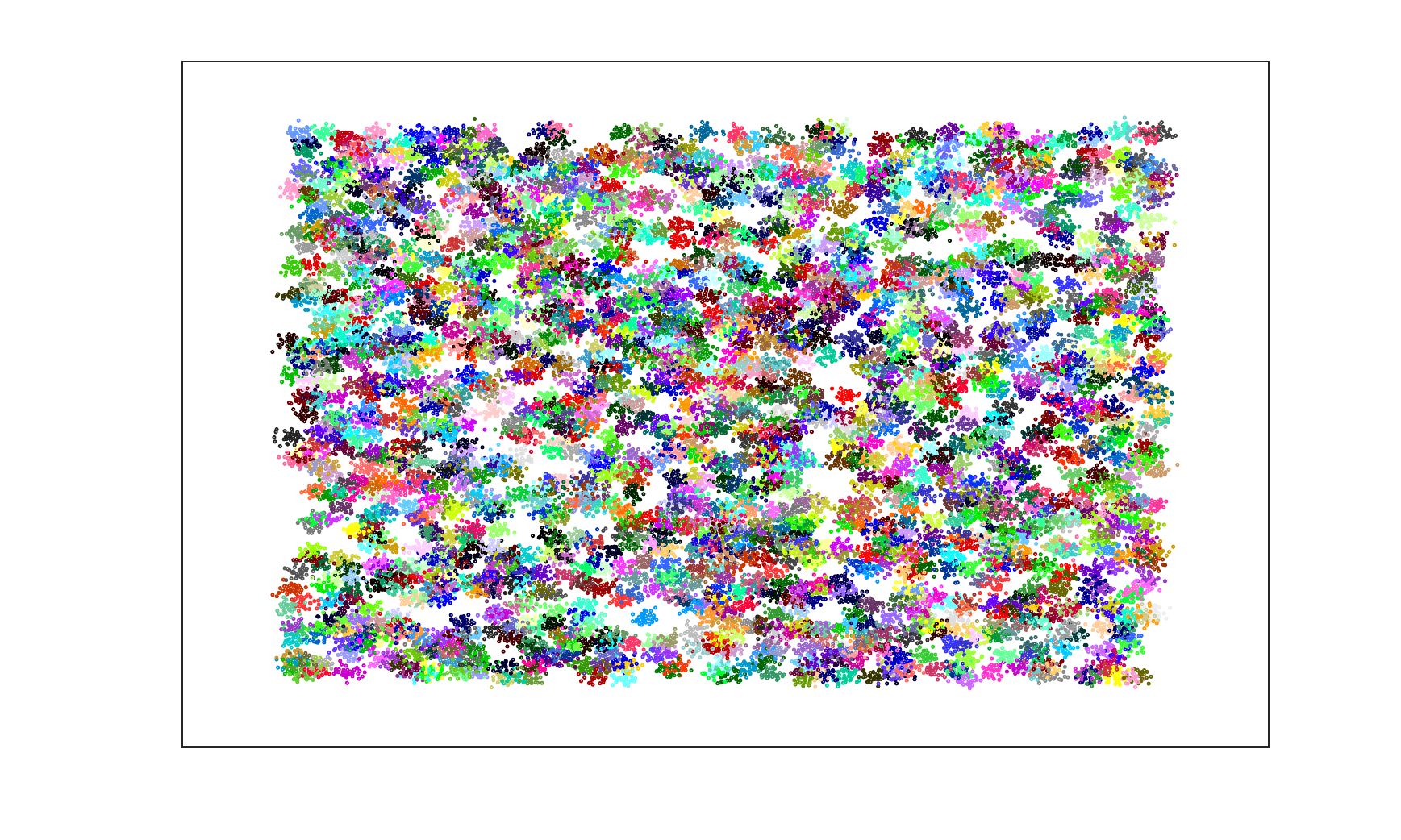}}}
    \caption{Illustration of four datasets with 500, 1000, 1500 and 2000 isotropic Gaussian blobs. The number of samples is 100,000 for each dataset.}
    \label{fig:simulation_landmarks}
  \end{center}
\end{figure}

To further investigate the performance of divide-and-conquer selection, we conduct a simulation experiment to simulate different scenarios for landmark selection.
For landmark selection, the number of landmarks is considered much larger than the desired number of clusters. 
If we view the landmark selection as a clustering task, then the landmark selection will be considered as a special clustering case with a large number of clusters.
Therefore, we generate four synthetic datasets with 500, 1000, 1500, 2000 clusters, respectively. 
The synthetic datasets are 2-dimensional isotropic Gaussian blobs, which are shown in Figure~\ref{fig:simulation_landmarks}.
We treat divide-and-conquer selection as a clustering algorithm to compare the clustering performance with $k$-means.
We report the clustering performance of NMI and time costs for all simulation scenarios in Table~\ref{tab:simulation_landmark_NMI} and Table~\ref{tab:simulation_landmark_time}. 

Though divide-and-conquer selection shows slightly lower NMI than $k$-means, its time cost is much less.
As landmark increases, the performance degradation associated with divide-and-conquer selection becomes progressively insignificant, while the improvement of efficiency becomes more significant.
The experimental results imply that the divide-and-conquer selection is suitable for a larger number of landmarks while $k$-means selection is suitable for a smaller number of landmarks.
Usually, more landmarks will lead to a better clustering result for large-scale spectral clustering \cite{cai2014large,huang2019ultra}.
Thus, the divide-and-conquer selection is more suitable than $k$-means selection for large-scale spectral clustering.

\begin{table}[]
    \centering
    \caption{The simulation performance of NMI(\%) varying different landmark selection scenarios.}
    \label{tab:simulation_landmark_NMI}
    \begin{tabular}{@{}lll@{}}
    \toprule
    Datasets & divide-and-conquer                 & $k$-means            \\ \midrule
    500 Gaussian blobs             & 90.76 & 92.01 \\
    1000 Gaussian blobs            & 87.67 & 88.51 \\
    1500 Gaussian blobs            & 85.90 & 86.40 \\
    2000 Gaussian blobs            & 84.59  & 84.82  \\ \bottomrule
    \end{tabular}
    \end{table}

    \begin{table}[]
    \centering
    \caption{The simulation time costs(s) varying different landmark selection scenarios.}
    \label{tab:simulation_landmark_time}
    \begin{tabular}{@{}lll@{}}
    \toprule
    Datasets & divide-and-conquer               & $k$-means            \\ \midrule
    500 Gaussian blobs             & 0.33 & 5.95  \\
    1000 Gaussian blobs            & 0.44 & 11.43 \\
    1500 Gaussian blobs            & 0.49 & 17.02 \\
    2000 Gaussian blobs            & 0.64 & 24.43 \\ \bottomrule
    \end{tabular}
    \end{table}

\subsection{Influence of Approximated $K$-nearest Landmarks}
\label{sec:cmpApproxKNN}

In this section, we compare the approximated $K$-nearest landmarks and exact $K$-nearest landmarks.
The experimental results are reported in Table \ref{table:compare_approxKNN}.
The approximated $K$-nearest landmarks approach first finds the possible candidates according to the center's nature of landmarks and then searches the $K$-nearest landmarks among them.
The exact $K$-nearest landmarks approach costs $O(Npd)$ computational time, while the proposed approximation can reduce the time cost to $O(NKd)$.
As the Tables~\ref{table:compare_approxKNN} shows, the exact $K$-nearest landmarks approach achieves slightly better ACC and NMI scores than the proposed approximation.
However, the performances of the two methods are not significantly different.
In terms of time cost, the proposed approximation approach shows highly efficient performance compared with the exact $K$-nearest landmarks.
Note that the exact $K$-nearest landmarks approach can not be conducted on datasets whose sizes are more than one million due to the high computational cost.
Overall, the proposed approximate $K$-nearest landmark approach shows the robustness and efficiency of this experiment.

\begin{table*}[]
  \centering
  \caption{Ablation Study on the proposed divide-and-conquer selection strategy, light-$k$-means, and approximate of K-nearest landmarks.}
  \label{tab:ablation}
  \begin{tabular}{@{}llllllll@{}}
    \toprule

    \multirow{3}{*}{Datasets} & \multirow{3}{*}{Landmark selection} & \multicolumn{6}{c}{$K$-nearest landmarks}                                  \\ \cmidrule(l){3-8}
    &                                     & \multicolumn{3}{l|}{Approximate}                 & \multicolumn{3}{l}{Exact}   \\ \cmidrule(l){3-8}
                                &                     & ACC(\%)        & NMI(\%)          & \multicolumn{1}{l|}{Time(s)}         & ACC(\%)          & NMI(\%)          & Time(s)   \\ \midrule
    \multirow{3}{*}{Letters}    & $k$-means           & {34.06}        & {\textbf{46.58}} & \multicolumn{1}{l|}{{3.89}}          & 34.41            & 45.56            & 4.05      \\
                                & DnC-$k$-means       & \textbf{34.71} & 45.19            & \multicolumn{1}{l|}{1.22}            & 33.76            & 45.17            & 1.34      \\
                                & DnC-light-$k$-means & {33.54}        & {45.37}          & \multicolumn{1}{l|}{{\textbf{0.90}}} & {33.93}          & {45.91}          & {1.05}    \\
    \midrule
    \multirow{3}{*}{MNIST}      & $k$-means           & {75.34}        & {73.07}          & \multicolumn{1}{l|}{{15.29}}         & \textbf{79.28}   & \textbf{74.74}   & 29.02     \\
                                & DnC-$k$-means       & 74.46          & 73.11            & \multicolumn{1}{l|}{9.50}            & 74.12            & 74.12            & 24.81     \\
                                & DnC-light-$k$-means & {74.24}        & {72.00}          & \multicolumn{1}{l|}{{\textbf{5.11}}} & {74.04}          & {74.04}          & {21.00}   \\
    \midrule
    \multirow{3}{*}{TS-60K}     & $k$-means           & {83.27}        & {\textbf{77.18}} & \multicolumn{1}{l|}{{165.21}}        & \textbf{86.41}   & 76.51            & 172.72    \\
                                & DnC-$k$-means       & 81.06          & 73.92            & \multicolumn{1}{l|}{8.14}            & 84.30            & 70.15            & 12.75     \\
                                & DnC-light-$k$-means & {81.00}        & {73.84}          & \multicolumn{1}{l|}{{\textbf{4.01}}} & {80.82}          & {73.12}          & {9.12}    \\
    \midrule
    \multirow{3}{*}{TM-1M}      & $k$-means           & {99.23}        & {99.50}          & \multicolumn{1}{l|}{{3997.12}}       & 99.95            & 99.59            & 4023.12   \\
                                & DnC-$k$-means       & 99.95          & 99.48            & \multicolumn{1}{l|}{12.78}           & \textbf{99.97}   & \textbf{99.57}   & 25.65     \\
                                & DnC-light-$k$-means & {99.96}        & {99.52}          & \multicolumn{1}{l|}{{\textbf{6.46}}} & {99.95}          & {99.45}          & {19.30}   \\
    \midrule
    \midrule
    \multirow{3}{*}{Avg. score} & $k$-means           & {72.98}        & {{74.08}}        & \multicolumn{1}{l|}{{1045.38}}       & {\textbf{75.01}} & {\textbf{74.10}} & {1057.23} \\
                                & DnC-$k$-means       & {72.55}        & {72.93}          & \multicolumn{1}{l|}{{7.91}}          & {73.04}          & {72.25}          & {16.14}   \\
                                & DnC-light-$k$-means & {72.19}        & {72.68}          & \multicolumn{1}{l|}{{\textbf{4.12}}} & {72.19}          & {73.13}          & {{12.62}} \\
    \bottomrule
  \end{tabular}%
\end{table*}

\subsection{Ablation Study}

To strike a good balance between efficiency and effectiveness, the proposed method applies three strategies: (a) divide-and-conquer selection, (b) light-$k$-means, and (c) approximate of $K$-nearest landmarks.
An ablation study about the influence of the combination of each part is conducted to show the contribution of each strategy.
The experimental results are reported in Table \ref{tab:ablation}.
Modules (a) and (b) are used in landmark selection.
In Table \ref{tab:ablation}, DnC-$k$-means indicates a modified divide-and-conquer selection that utilizes $k$-means algorithm for the dividing process, and DnC-light-$k$-mean indicates the original divide-and-conquer selection that utilizes the light-$k$-means algorithm for the dividing process.
To show the effects of (a) and (b), we choose $k$-means selection as the baseline.
For $K$-nearest landmarks, we provide the exact $K$-nearest landmark option for each landmark selection method.
There are three landmark selections and two $K$-nearest landmark methods provided in this ablation study.
Thus we have six combinations for comparison. 

Table \ref{tab:ablation} shows the performance on six combinations according to different landmark selection and $K$-nearest landmarks methods.
The bold texts represent the best ACC, NMI, and Time for each dataset.
We first compare the different landmark selection methods:
the $k$-means selection archives the best ACC on two datasets and the best NMI on three datasets, but takes much more runtime on all datasets;
DnC-$k$-means selection archives the best ACC on two datasets and best NMI on one dataset with much less runtime than $k$-means selection;
our DnC-light-$k$-means takes the least time on all datasets and shows a competitive performance of ACC and NMI.
For $K$-nearest landmark, the extra approach archives the best ACC on three datasets and the best NMI on two datasets, which slightly outperforms the approximate approach.
We also report the average score for each combination.
The combination of $k$-means and extra $K$-nearest landmarks show the best average scores of ACC and NMI, but the most time-consuming.
Our proposed method that is the combination of DnC-light-$k$-means and approximate $K$-nearest landmarks shows the fastest speed and competitive performance of ACC and NMI.

Overall, the proposed method significantly improves the efficiency of large-scale spectral clustering while keeping the clustering quality acceptable.
In detail, we can see that the use of strategy (a) provides the most important contribution to the computational efficiency, while modules (b) and (c) further reduce the computational cost.

\section{Conclusion}
\label{sec:conclusion}
In this paper, we propose a large-scale clustering method, termed divide-and-conquer based spectral clustering (DnC-SC).
In DnC-SC, a divide-and-conquer based landmark selection algorithm is designed to obtain the landmarks effectively.
A new approximate similarity matrix construction approach is proposed to utilize the center's nature of the landmarks to fast construct the similarity matrix between data points and $K$-nearest landmarks.
Finally, the bipartite graph partition is conducted to obtain the final clustering results. 
The experimental results on synthetic and real-world datasets show that the proposed method outperforms other state-of-the-art large-scale spectral clustering methods.

\section*{Acknowledgment}
This study was supported by in part by the New Energy and Industrial Technology Development Organization (NEDO) Grant (ID:18065620) and JST COI-NEXT.

\bibliographystyle{cas-model2-names}

\bibliography{main.bib}

\newpage

\bio{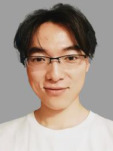}
Hongmin Li is currently working toward a Ph.D. degree at the Department of Computer Science, University of Tsukuba, Japan. He received his MS degree in computer science from the University of Tsukuba, Japan. His current research interests include clustering, machine learning, and its application fields.
\endbio

\bio{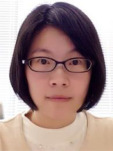}
Xiucai Ye received the Ph.D. degree in computer science from the University of Tsukuba, Tsukuba, Japan, in 2014. She is currently an Assistant Professor with the Department of Computer Science, and Center for Artificial Intelligence Research (C-AIR), University of Tsukuba. Her current research interests include feature selection, clustering, bioinformatics, machine learning and its application fields. She is a member of IEEE.
\endbio

\bio{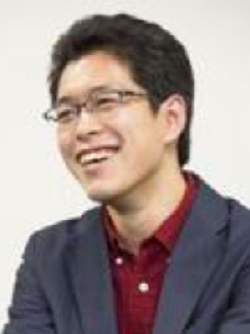}
Akira Imakura is an Associate Professor at Faculty of Engineering, Information and Systems, University of Tsukuba, Japan. He received Ph.D. (2011) from Nagoya University, Japan. He was appointed as Japan Society for the Promotion of Science Research Fellowship for Doctor Course Student (DC2) from 2010 to 2011, as a Research Fellow at Center for Computational Sciences, University of Tsukuba, Japan from 2011 to 2013, and also as a JST ACTI researcher from 2016 to 2019. His current research interests include developments and analysis of highly parallel algorithms for large matrix computations. Recently, he also investigates matrix factorization-based machine learning algorithms. He is a member of JSIAM, IPSJ and SIAM.
\endbio

\bio{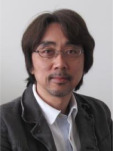}
Tetsuya Sakurai is a Professor of Department of Computer Science, and the Director of Center for Artificial Intelligence Research (C-AIR) at the University of Tsukuba. He is also a visiting professor at the Open University of Japan, and a visiting researcher of Advanced Institute of Computational Science at RIKEN. He received a Ph.D. in Computer Engineering from Nagoya University in 1992. His research interests include high performance algorithms for large-scale simulations, data and image analysis, and deep neural network computations. He is a member of the Japan Society for Industrial and Applied Mathematics (JSIAM), the Mathematical Society of Japan (MSJ), Information Processing Society of Japan (IPSJ), Society for Industrial and Applied Mathematics (SIAM).
\endbio
\end{document}